\documentclass[10pt,journal,compsoc]{IEEEtran}

\usepackage{pgfplots}
\pgfplotsset{compat=newest}
\usepackage{tikz}

\ifCLASSOPTIONcompsoc
  \usepackage[nocompress]{cite}
\else
  \usepackage{cite}
\fi

\usepackage{amsmath}
\usepackage{amssymb}
\usepackage{amsthm}

\ifCLASSOPTIONcompsoc
 \usepackage[caption=false,font=footnotesize,labelfont=sf,textfont=sf]{subfig}
\else
 \usepackage[caption=false,font=footnotesize]{subfig}
\fi

\usepackage{times}
\usepackage{epsfig}
\usepackage{graphicx}

\usepackage[utf8x]{inputenc}
\usepackage{color}
\usepackage{xcolor}
\usepackage{booktabs,siunitx} 
\usepackage{makecell}
\usepackage{enumitem}
\usepackage{rotating}
\usepackage[misc]{ifsym}

\usepackage[pagebackref=true,breaklinks=true,colorlinks,bookmarks=false]{hyperref}

\if@mathematic
   
\else
   
\fi

\definecolor{road}                {RGB}{128, 64,128}
\definecolor{sidewalk}            {RGB}{244, 35,232}
\definecolor{building}            {RGB}{ 70, 70, 70}
\definecolor{wall}                {RGB}{102,102,156}
\definecolor{fence}               {RGB}{190,153,153}
\definecolor{pole}                {RGB}{153,153,153}
\definecolor{traffic light}       {RGB}{250,170, 30}
\definecolor{traffic sign}        {RGB}{220,220,  0}
\definecolor{vegetation}          {RGB}{107,142, 35}
\definecolor{terrain}             {RGB}{152,251,152}
\definecolor{sky}                 {RGB}{ 70,130,180}
\definecolor{person}              {RGB}{220, 20, 60}
\definecolor{rider}               {RGB}{255,  0,  0}
\definecolor{car}                 {RGB}{  0,  0,142}
\definecolor{truck}               {RGB}{  0,  0, 70}
\definecolor{bus}                 {RGB}{  0, 60,100}
\definecolor{train}               {RGB}{  0, 80,100}
\definecolor{motorcycle}          {RGB}{  0,  0,230}
\definecolor{bicycle}             {RGB}{119, 11, 32}
\definecolor{void}                {RGB}{  0,  0,  0}

\newcommand{\vecnorm}[1]{\left\|#1\right\|}
\newtheorem{thm}{Theorem}

\newcommand\ver[1]{\rotatebox[origin=c]{90}{#1}}

\newcommand{\yes}{\checkmark}
\newcommand{\no}{$\times$}

\newcommand{\best}[1]{\textbf{#1}}

\newcommand{\PAR}[1]{\vskip4pt \noindent{\bf #1~}}

\begin{document}

\title{Map-Guided Curriculum Domain Adaptation and Uncertainty-Aware Evaluation for Semantic Nighttime Image Segmentation}

\author{Christos~Sakaridis,
        Dengxin~Dai,
        and~Luc~Van~Gool
\IEEEcompsocitemizethanks{\IEEEcompsocthanksitem C.~Sakaridis is with the Department of Information Technology and Electrical Engineering, ETH Z\"urich, Switzerland.\protect\\
E-mail: csakarid@vision.ee.ethz.ch
\IEEEcompsocthanksitem D.~Dai is with the Department of Information Technology and Electrical Engineering, ETH Z\"urich, Switzerland.
\IEEEcompsocthanksitem L.~Van Gool is with the Department of Information Technology and Electrical Engineering, ETH Z\"urich, Switzerland, and with the Department of Electrical Engineering, KU Leuven, Belgium.}
}

\markboth{IEEE Transactions on Pattern Analysis and Machine Intelligence,~Vol.~xx, No.~xx, December~2020}%
{Sakaridis \MakeLowercase{\textit{et al.}}: Map-Guided Curriculum Domain Adaptation and Uncertainty-Aware Evaluation for Semantic Nighttime Image Segmentation}

\IEEEtitleabstractindextext{%
\begin{abstract}
We address the problem of semantic nighttime image segmentation and improve the state-of-the-art, by adapting daytime models to nighttime without using nighttime annotations. Moreover, we design a new evaluation framework to address the substantial \emph{uncertainty} of semantics in nighttime images. Our central contributions are: 1) a curriculum framework to gradually adapt semantic segmentation models from day to night through progressively darker times of day, exploiting cross-time-of-day correspondences between daytime images from a reference map and dark images to guide the label inference in the dark domains; 2) a novel uncertainty-aware annotation and evaluation framework and metric for semantic segmentation, including image regions \emph{beyond human recognition capability} in the evaluation in a principled fashion; 3) the \emph{Dark Zurich} dataset, comprising 2416 unlabeled nighttime and 2920 unlabeled twilight images with correspondences to their daytime counterparts plus a set of 201 nighttime images with fine pixel-level annotations created with our protocol, which serves as a first benchmark for our novel evaluation. Experiments show that our map-guided curriculum adaptation significantly outperforms state-of-the-art methods on nighttime sets both for standard metrics and our uncertainty-aware metric. Furthermore, our uncertainty-aware evaluation reveals that selective invalidation of predictions can improve results on data with ambiguous content such as our benchmark and profit safety-oriented applications involving invalid inputs.
\end{abstract}

}

\maketitle

\IEEEdisplaynontitleabstractindextext

\IEEEpeerreviewmaketitle

\IEEEraisesectionheading{\section{Introduction}\label{sec:intro}}

\IEEEPARstart{T}{he} state of the art in semantic segmentation is rapidly improving in recent years. Despite the advance, most methods are designed to operate at daytime, under favorable illumination conditions. However, many outdoor applications require robust vision systems that perform well at all times of day, under challenging lighting conditions, and in bad weather~\cite{vision:atmosphere,dehazing:LAPNet}. Currently, the popular approach to solving perceptual tasks such as semantic segmentation is to train deep neural networks~\cite{pspnet,refinenet,dilated:convolution} using large-scale human annotations~\cite{pascal:2011,Cityscapes,Mapillary}. This supervised scheme has achieved great success for daytime images, but it scales badly to adverse conditions like nighttime. The adversity of nighttime poses further challenges for perceptual tasks compared to daytime. The extracted features become corrupted due to visual hazards~\cite{cv:hazop} such as underexposure, noise, and motion blur. In this work, we focus on semantic segmentation at nighttime, both at the method level and the evaluation level.

At the method level, this work adapts semantic segmentation models from daytime to nighttime, without annotations in the latter domain.
To this aim, we propose a new method called Map-Guided Curriculum Domain Adaptation (MGCDA). The underpinnings of MGCDA are threefold: continuity of time, prior knowledge of place, and power of data. \textbf{Time}: environmental illumination changes continuously from daytime to nighttime. This enables adding intermediate domains between the two to smoothly transfer semantic knowledge. This idea is found to be effective in~\cite{SynRealDataFogECCV18,daytime:2:nighttime}; we extend it by adding two more modules. \textbf{Place}: images taken over different time but with the same 6D camera pose share a large portion of content. The shared content can be used to guide the knowledge transfer process from a favorable condition (daytime) to an adverse condition (nighttime). 
We formalize this observation and propose a method for large-scale applications. The method stores the daytime images and the distilled semantic knowledge into a digital map and enhances the semantic nighttime image segmentation by this geo-referenced map in an adaptive fusion framework. This supplement is especially important for nighttime perception as observing partial information and uncertain data is a frequent situation at nighttime.
\textbf{Data}: MGCDA takes advantage of the powerful image translation techniques to stylize real annotated daytime datasets to darker target domains in order to perform standard supervised learning.

At the evaluation level, this work proposes an uncertainty-aware annotation and evaluation framework for semantic segmentation.
The degradation of regions of nighttime images affected by visual hazards is often so intense that they are rendered \emph{indiscernible}, i.e.\ determining their semantic content is impossible even for humans. We term such regions as \emph{invalid} for the task of semantic segmentation. A robust model should predict with high \emph{uncertainty} on invalid regions while still being confident on valid (discernible) regions, and a sound evaluation framework should reward such behavior. The above requirement is particularly significant for safety-oriented applications such as autonomous cars, since having the vision system declare a prediction as invalid can help the downstream driving system avoid the fatal consequences of this prediction being false, e.g.\ when a pedestrian is missed.

To this end, we design a generic uncertainty-aware annotation and evaluation framework for semantic segmentation in adverse conditions which explicitly distinguishes invalid from valid regions of input images, and apply it to nighttime. On the annotation side, our novel protocol leverages privileged information in the form of daytime counterparts of the annotated nighttime scenes, which reveal a large portion of the content of invalid regions. This allows to reliably label invalid regions and to indeed \emph{include} invalid regions in the evaluation, contrary to existing semantic segmentation benchmarks~\cite{Cityscapes} which completely exclude them from evaluation. Moreover, apart from the standard class-level semantic annotation, each image is annotated with a mask which designates its invalid regions. On the evaluation side, we allow the \emph{invalid} label in predictions and adopt from~\cite{wilddash} the principle that for invalid pixels with legitimate semantic labels, both these labels and the \emph{invalid} label are considered correct predictions. However, this principle does not cover the case of valid regions. We address this by introducing the concept of false invalid predictions. This enables calculation of \emph{uncertainty-aware intersection-over-union (UIoU)}, a joint performance metric for valid and invalid regions which generalizes standard IoU, reducing to the latter when no invalid prediction exists. UIoU rewards predictions which exhibit confidence that is \emph{consistent} to human annotators, i.e.\ which have higher confidence on valid regions than invalid ones, meeting the aforementioned requirement.

Finally, we present \emph{Dark Zurich}, a dataset of 8779 real images which contains corresponding images of the same driving scenes at daytime, twilight and nighttime. We use this dataset to feed real data to MGCDA and to create a benchmark with 201 nighttime images for our uncertainty-aware evaluation. Our dataset is publicly available\footnote{\scriptsize{\url{https://www.trace.ethz.ch/publications/2019/GCMA_UIoU}}} and is used for hosting a CVPR 2020 challenge on nighttime segmentation\footnote{\scriptsize{\url{https://competitions.codalab.org/competitions/23553}}}.
Our code is publicly available\footnote{\scriptsize{\url{https://github.com/sakaridis/MGCDA}}}.

An earlier version of this work has appeared in the International Conference on Computer Vision~\cite{GCMA_UIoU:v1}. Compared to the conference version, this paper makes the following additional contributions:
\begin{enumerate}
    \item An improved version of our domain adaptation method which involves a \emph{geometry-aware} formulation for refining semantic predictions via cross-time-of-day correspondences and leads to improved performance over the conference version.
    \item An extension of the annotated nighttime part of our \emph{Dark Zurich} dataset with 50 additional images, leading to a total of 201 annotated nighttime images.
    \item Substantially more extensive experiments, including i.a.\ detailed comparisons with more recent state-of-the-art domain adaptation methods for semantic segmentation, evaluation on additional nighttime sets, thorough ablation studies for the components of our method, and application of our approach at test time.
    \item Other enhanced parts, including related work and dataset statistics.
\end{enumerate}

\section{Related Work} 
\label{sec:related} 

\PAR{Vision at Nighttime.}
Nighttime has attracted a lot of attention in the literature due to its ubiquitous nature. Several works pertain to human detection at nighttime, using FIR cameras~\cite{night:vision:pedestrian:05,pedestrian:detection:tracking:night:09}, visible light cameras~\cite{cnn:human:detection:nighttime:17}, or a combination of both~\cite{nighttime:pedestrian:detection:08,kaist:day:night:data:18}. In driving scenarios, a few methods have been proposed to detect cars~\cite{nighttime:object:proposal:18} and vehicles' rear lights~\cite{night:rear:lights:16}. Contrary to these domain-specific methods, previous work also includes both methods designed for robustness to illumination changes, by employing domain-invariant representations~\cite{road:detection:illumination:invariant,outdoor:transformation:labeling:iv15} or fusing information from complementary modalities and spectra~\cite{AdapNet:adverse:17}, and datasets with adverse illumination~\cite{Oxford,localization:benchmarking:adverse,astar:3d:dataset:adverse:icra20}.
A recent work~\cite{daytime:2:nighttime} on semantic nighttime segmentation shows that images captured at twilight are helpful for supervision transfer from daytime to nighttime. Our work is partially inspired by~\cite{daytime:2:nighttime} and extends it by proposing a map-guided curriculum adaptation framework which learns jointly from stylized images and unlabeled real images of increasing darkness and exploits the prior knowledge from a map. There is a rich literature on low-light image enhancement~\cite{MBLLEN,learning:see:dark:cvpr18,Wang_2019_CVPR,ZeroDCE}, which is also relevant to our work. However, its focus is on the low-level goal of visual quality improvement rather than the high-level goal of accurate semantic scene understanding.

\PAR{Domain Adaptation.}
Performance of semantic segmentation on daytime scenes has increased rapidly in recent years. As a consequence, attention is now turning to adaptation to adverse conditions~\cite{AdapNet:adverse:17,benchmark:sensor:adverse:weather:18,wulfmeier2017addressing,continuous:manifold:adaptation}. A case in point are recent efforts to adapt clear-weather models to fog~\cite{SFSU_synthetic,SynRealDataFogECCV18,CMAda:IJCV2020,FoggySynscapes}, by using both labeled synthetic images and unlabeled real images of increasing fog density. This work instead focuses on the nighttime domain, which poses very different and---as we would claim---greater challenges than the foggy domain (e.g.\ artificial light sources casting very different illumination patterns at night). A major class of adaptation approaches, including~\cite{cyCADA,learning:synthetic:data:cvpr18,chen2018road,adapt:structured:output:cvpr18,incremental:adversarial:DA:18,conditional:GAN:adaptation,FCNs:adaptation,conservative:loss:adaptation,DCAN:adaptation,bidirectional:learning:adaptation,advent:adaptation,madan:adaptation,ccm:adaptation,plca:adaptation}, involves adversarial confusion or feature alignment between domains.
The general concept of curriculum learning has been successfully applied to domain adaptation by ordering tasks~\cite{curriculum:domain:adaptation:17}, target-domain pixels~\cite{self:training:adaptation}, or domains~\cite{SynRealDataFogECCV18,daytime:2:nighttime,CMAda:IJCV2020,compound:domain:adaptation}. Our method belongs to the last group. Cross-domain correspondences as guidance have only been used very recently in~\cite{cross:season:correspondence}, which requires pixel-level matches to be given, while we require more generic image-level correspondences.

\begin{figure*}
    \centering
    \includegraphics[clip,width=0.8\textwidth,trim=405mm 390mm 310mm 150mm]{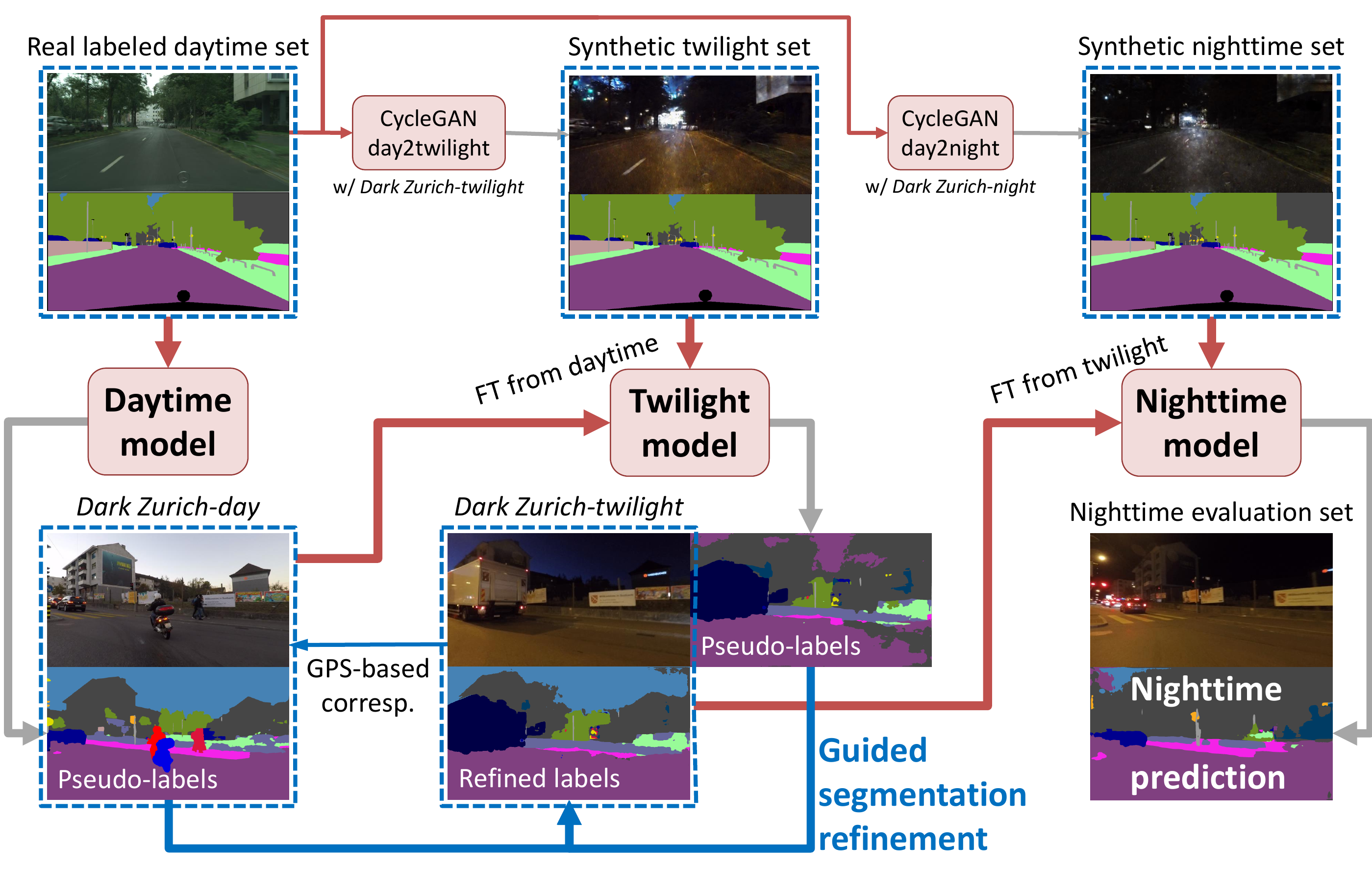}
    \caption{A general overview of our MGCDA method for adaptation to nighttime. FT stands for fine-tuning. Red arrows denote training of a model, while gray arrows denote generation of predictions. Inference with the final nighttime model (bottom right) does not require guided refinement.}
    \label{fig:mgcda}
\end{figure*}

\PAR{Semantic Segmentation Evaluation.}
Semantic segmentation evaluation is commonly performed with the IoU metric~\cite{pascal:2011}. Cityscapes~\cite{Cityscapes} introduced an instance-level IoU (iIoU) to remove the large-instance bias, as well as mean average precision for the task of instance segmentation. The two tasks have recently been unified into panoptic segmentation~\cite{panoptic:segmentation}, with a respective panoptic quality metric. The most closely related work to ours in this regard is WildDash~\cite{wilddash}, which uses standard IoU together with a fine-grained evaluation to measure the impact of visual hazards on performance. In contrast, we introduce UIoU, a new semantic segmentation metric that handles images with regions of uncertain semantic content and is suited for adverse conditions. Our uncertainty-aware evaluation is \emph{complementary} to uncertainty-aware methods such as~\cite{uncertainty:bayesian} and \cite{simultaneous:segmentation:outliers} that explicitly incorporate uncertainty in their model formulation and aims to promote the development of such methods, as UIoU rewards models that accurately capture heteroscedastic aleatoric uncertainty~\cite{uncertainty:bayesian} in the input images through the different treatment of invalid and valid regions.  

\PAR{Map-Guided Vision Applications.} 
One of the major application domains of maps is robot localization, which is a large research field on its own and has a rich literature~\cite{localization:probabilistic:maps,video:localization:iv14}. Maps have also been enriched to be leveraged for other vision tasks beyond localization such as road surface detection~\cite{map:drivable:space:12,map:road:recognition:13}, navigation~\cite{navigation:openstreetmap:10,drive:surroundview:route:planner}, object detection~\cite{map:aided:driving:scene:understanding:15,HDNET:2018}, tracking~\cite{argoverse:cvpr19} and forecasting~\cite{action:prediction:map:20}. This work uses a new form of map-based prior knowledge, i.e.\ daytime images and their distilled semantics, to supplement the task of semantic image segmentation. This supplement is especially important when the online segmentation system operates in challenging weather or lighting conditions, e.g.\ at nighttime. Our learning method uses geo-referenced maps as an additional source of information in an adaptive fusion scheme.

\section{Map-Guided Curriculum Domain Adaptation} 
\label{sec:mgcda}

\subsection{Problem Formulation}
\label{sec:mgcda:general}
MGCDA involves a source domain $\mathcal{S}$, an ultimate target domain $\mathcal{T}$, and an intermediate target domain $\dot{\mathcal{T}}$. In this work, $\mathcal{S}$ is daytime, $\mathcal{T}$ is nighttime, and $\dot{\mathcal{T}}$ is twilight time with an intermediate level of darkness between $\mathcal{S}$ and $\mathcal{T}$. These domains are formally defined according to the solar elevation angle~\cite{daytime:2:nighttime}, which can be derived from geographical location and time of capture. MGCDA adapts semantic segmentation models through this sequence of domains $(\mathcal{S}, \dot{\mathcal{T}}, \mathcal{T})$, which is sorted in ascending order with respect to level of darkness. The approach proceeds progressively and adapts the model from one domain in the sequence to the next. The knowledge is transferred through the domain sequence via this gradual adaptation process. The transfer is performed using two coupled branches: 1) learning from labeled synthetic stylized images and 2) learning from real data without annotations, to jointly leverage the assets of both. Stylized images inherit the human annotations of their original counterparts but contain unrealistic artifacts, whereas real images have less reliable pseudo-labels but are characterized by artifact-free textures. An overview of MGCDA is presented in Fig.~\ref{fig:mgcda}.

Let us use $z \in \{1,2,3\}$ as the index in $(\mathcal{S}, \dot{\mathcal{T}}, \mathcal{T})$. 
Once the model for the current domain $z$ is trained, its knowledge can be distilled on unlabeled real data from $z$, and then used, along with a new version of synthetic data from the next domain $z+1$ to adapt the current model to $z+1$.

\begin{figure}
    \centering
    \subfloat[$\mathcal{D}^1_{lr}$: Cityscapes]{\includegraphics[height=0.12\textwidth]{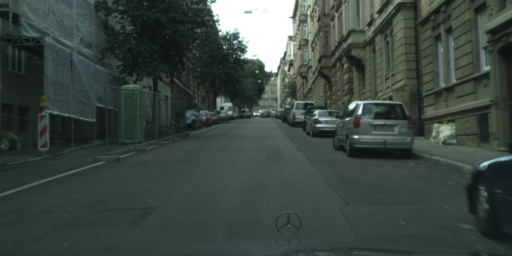}}
    \hfil
    \subfloat[$\mathcal{D}^1_{ur}$: \emph{Dark Zurich-day}]{\includegraphics[height=0.12\textwidth]{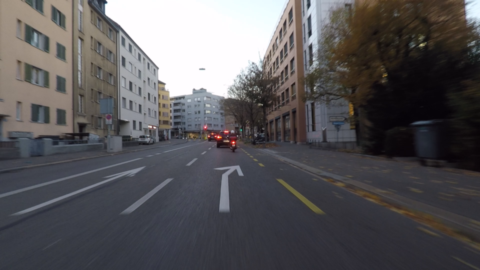}}
    \\
    \vspace{-0.3cm}
    \subfloat[$\mathcal{D}^2_{ls}$: Cityscapes-twilight style]{\includegraphics[height=0.12\textwidth]{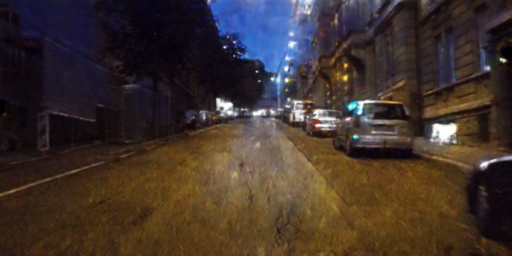}}
    \hfil
    \subfloat[$\mathcal{D}^2_{ur}$: \emph{Dark Zurich-twilight}]{\includegraphics[height=0.12\textwidth]{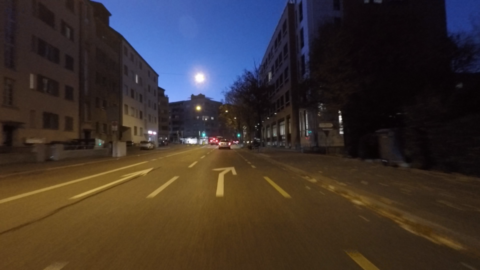}}
    \\
    \vspace{-0.3cm}
    \subfloat[$\mathcal{D}^3_{ls}$: Cityscapes-nighttime style]{\includegraphics[height=0.12\textwidth]{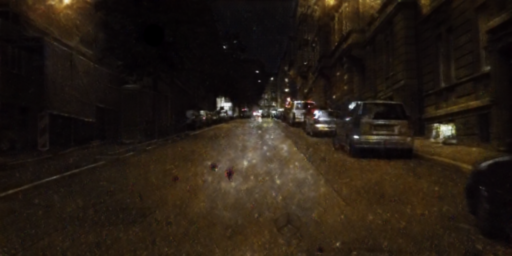}}
    \hfil
    \subfloat[$\mathcal{D}^3_{ur}$: \emph{Dark Zurich-night}]{\includegraphics[height=0.12\textwidth]{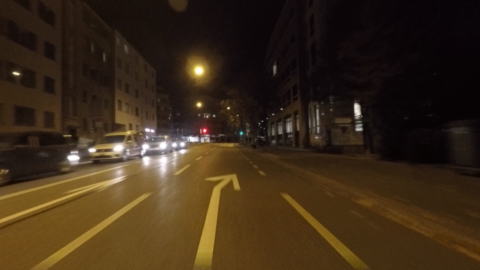}}
    \caption{Sample images from the training sets used in MGCDA.}
    \label{fig:training:datasets}
\end{figure}

Before diving into the details, we first define all datasets used. 
The inputs for MGCDA consist of: 1) a labeled daytime set with $M$ real images $\mathcal{D}^1_{lr}=\{(I_m^1,Y^1_m)\}_{m=1}^M$, e.g.\ Cityscapes~\cite{Cityscapes}, where $Y_m^{1}(i,j) \in \mathcal{C} = \{1, ..., C\}$ is the ground-truth label of pixel $(i,j)$ of $I_m^{1}$; 2) an unlabeled daytime set of $N_1$ images $\mathcal{D}^1_{ur}=\{I_n^1\}_{n=1}^{N_1}$; 3) an unlabeled twilight set of $N_2$ images $\mathcal{D}^2_{ur}=\{I_n^2\}_{n=1}^{N_2}$; and 4) an unlabeled nighttime set of $N_3$ images $\mathcal{D}_{ur}^3=\{I_n^3\}_{n=1}^{N_3}$. In order to perform knowledge transfer with annotated data, $\mathcal{D}_{lr}^1$ is rendered in the style of $\mathcal{D}^2_{ur}$ and $\mathcal{D}_{ur}^3$. We use CycleGAN~\cite{cycleGAN} to perform this style transfer, leading to two more sets: $\mathcal{D}^2_{ls}=\{(\bar{I}_m^2,Y^1_m)\}_{m=1}^M$ and $\mathcal{D}^3_{ls}=\{(\bar{I}_m^3,Y^1_m)\}_{m=1}^M$, where $\bar{I}_m^2$ and $\bar{I}_m^3$ are the stylized twilight and nighttime version of $I_m^1$ respectively, and labels are copied. 
For $z=1$, the semantic segmentation model $\phi^1$ is trained directly on $\mathcal{D}^1_{lr}$.
In order to perform knowledge transfer with unlabeled data, pseudo-labels for all three unlabeled real datasets need to be generated. The pseudo-labels for $\mathcal{D}_{ur}^1$ are generated using the model $\phi^1$ via $\hat{Y}_n^1=\phi^1(I_n^1)$. 
For $z>1$, training $\phi^z$ and generating $\hat{Y}_m^z$ is performed progressively as MGCDA proceeds, as is detailed in Sec.~\ref{sec:mgcda:learning}. All six datasets are summarized in Table~\ref{tab:GCMA:notations}. In Fig.~\ref{fig:training:datasets}, we show visual examples from the six training sets. Cityscapes~\cite{Cityscapes} is used to instantiate the labeled sets, while our \emph{Dark Zurich} dataset, which we detail in Sec.~\ref{sec:dark:zurich}, is used to instantiate the unlabeled sets.

\setlength{\tabcolsep}{1pt}
\begin{table}
\caption{The training sets used in MGCDA. $I$ indicates an image and $Y$ its label map; $\bar{I}$ is a synthetic image and $\hat{Y}$ a pseudo-label map. See the text for details.}
    \label{tab:GCMA:notations}
    \centering
    \footnotesize
    \begin{tabular}{lccc}
 \toprule
 &  \multicolumn{2}{c}{Labeled}  & Unlabeled   \\
 & Real & Synthetic & Real  \\
 1. Daytime & $\{(I_m^1,Y^1_m)\}_{m=1}^M$ & &  $\{(I_n^1,\hat{Y}^1_n)\}_{n=1}^{N_1}$  \\
 2. Twilight time & & $\{(\bar{I}_m^2,Y^1_m)\}_{m=1}^M$ &   $\{(I_n^2,\hat{Y}^2_n)\}_{n=1}^{N_2}$  \\
 3. Nighttime & & $\{(\bar{I}_m^3,Y^1_m)\}_{m=1}^M$ &  $\{(I_n^3,\hat{Y}^3_n)\}_{n=1}^{N_3}$ \\ 
 \bottomrule
\end{tabular}
\end{table}

\begin{figure*}
    \centering
    \subfloat[Establishment of image-level correspondences via map guidance]{\includegraphics[width=0.48\textwidth]{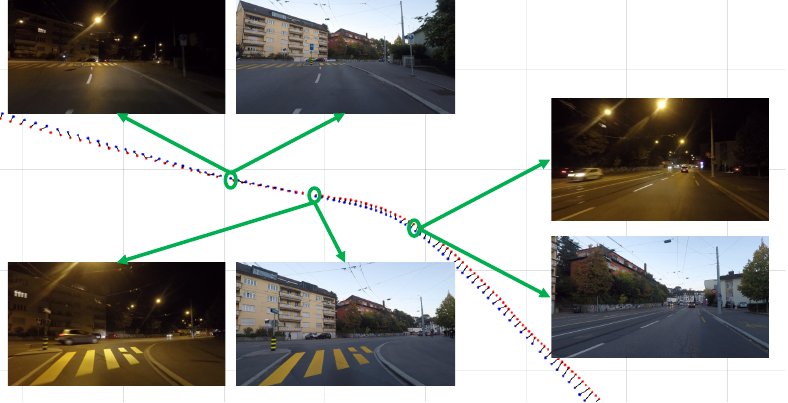}\label{fig:map:matching}} 
    \hfil     
    \subfloat[Establishment of dense pixel-level correspondences]{\includegraphics[width=0.48\textwidth]{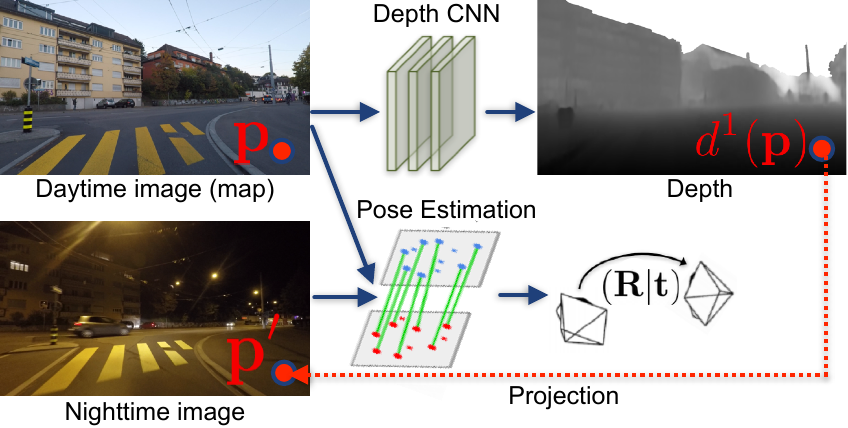}\label{fig:image:warping}}
    \hfil
    \caption{Illustration of two components of MGCDA involving the identification of cross-time-of-day correspondences at image level and pixel level.}
    \label{fig:map:matching:image:warping}
\end{figure*}

\subsubsection{Map-Guided Curriculum Domain Adaptation} 
\label{sec:mgcda:learning}
Since the method proceeds in an iterative manner, we present the algorithmic details only for a single adaptation step from $z-1$ to $z$. The presented algorithm is straightforward to generalize to multiple intermediate target domains.
In order to adapt the semantic segmentation model $\phi^{z-1}$ from the previous domain $z-1$ to the current domain $z$, we generate synthetic stylized data in domain $z$: $\mathcal{D}_{ls}^z$.   

For real unlabeled images, since no human annotations are available, we rely on a strategy of self-learning or curriculum learning. Our motivating assumption is that objects are generally easier to recognize in lighter conditions, so the tasks are solved in ascending order with respect to the level of darkness and the easier, solved tasks are used to re-train the model to solve the harder tasks. This is in line with the concept of curriculum learning~\cite{curriculum:learning}.  
In particular, the model $\phi^{z-1}$ for domain $z-1$ can be applied to the unlabeled real images of domain $z-1$ to generate supervisory labels for training $\phi^{z}$. Specifically, the dataset of real images with pseudo-labels for adaptation to domain $z$ is $\mathcal{D}_{ur}^{z-1}=\{(I_n^{z-1}, \hat{Y}_n^{z-1})\}_{n=1}^{N_{z-1}}$, where $\hat{Y}_n^{z-1}$ denotes the predicted labels of image $I_n^{z-1}$. A simple way to get these labels is by directly feeding $I_n^{z-1}$ to $\phi^{z-1}$, similar to the approach of~\cite{SynRealDataFogECCV18,CMAda:IJCV2020} for the case of fog. This choice, however, suffers from accumulation of substantial errors in the prediction of $\phi^{z-1}$ into the subsequent training step if domain $z-1$ is not the daytime domain. We instead propose a method to refine these errors by using \emph{weak supervision} from the semantics of a daytime image $I_n^{1}$ that \emph{corresponds} to $I_n^{z-1}$, i.e.\ depicts roughly the same scene as $I_n^{z-1}$ (the difference in the camera pose is small):
\begin{equation} \label{eq:guided:prediction} 
\hat{Y}_n^{z-1} = G\left(\phi^{z-1}(I_n^{z-1}), I_n^{z-1}, \phi^{1}(I_{A_{z-1 \rightarrow 1}(n)}^{1}), I_{A_{z-1 \rightarrow 1}(n)}^{1}\right),
\end{equation}
where $G$ is a guidance function which will be defined in Sec.~\ref{sec:mgcda:guidance} and $z-1 > 1$. $A_{z-1 \rightarrow 1}(n)$ is the correspondence function giving the index of the daytime image that corresponds to $I_n^{z-1}$.

Once we have the two training sets $\mathcal{D}^{z-1}_{ur}$ (with labels inferred through \eqref{eq:guided:prediction}) and $\mathcal{D}^z_{ls}$, learning $\phi^z$ is performed by optimizing a loss function that involves both datasets:
\begin{equation} \label{eq:ssl}
\min_{\phi^z}  \bigg(\sum_{\substack{(I, Y)  \\ \in \mathcal{D}_{ls}^z}} L(\phi^z(I),Y) + \mu \sum_{\substack{(I, \hat{Y}) \\ \in \mathcal{D}_{ur}^{z-1}}} L(\phi^z(I), \hat{Y}) \bigg),
\end{equation}
where $L(.,.)$ is the cross entropy loss and $\mu$ is a hyper-parameter balancing the contribution of the two datasets.

In order to leverage the \emph{map} prior at large scale to improve predictions through the guided label refinement defined in \eqref{eq:guided:prediction}, specific aligned training datasets need to be compiled. With this aim, we collected the \emph{Dark Zurich} dataset by driving several laps in disjoint areas of Zurich; each lap was driven multiple times during the same day, starting from daytime through twilight to nighttime. The recording overhead is moderate compared to the unsupervised setting where only a single nighttime recording would be performed. The recordings include GPS readings and are split into three sets: daytime, twilight and nighttime (cf.\ Sec.~\ref{sec:dark:zurich}). Since different drives of the same lap correspond to the same route, the camera orientation at a certain point of the lap is similar across all drives. We implement the correspondence function $A_{z\rightarrow1}$ that assigns to each image in domain $z$ its daytime counterpart using a GPS-based nearest neighbor assignment, as shown in Fig.~\ref{fig:map:matching:image:warping}\subref{fig:map:matching}. Thus, the requirements for the training dataset compared to the unsupervised adaptation case do not significantly restrict the applicability of MGCDA. The method presented in Sec.~\ref{sec:mgcda:guidance} carefully handles the effects of misalignment and dynamic objects in paired images. 

The geo-referenced daytime images, along with their semantic pseudo-labels, are used as a new form of map knowledge. We acknowledge that our method uses a very simple map-matching method that may not be sufficient for other tasks. However, we show that our learning method is able to benefit from the corresponding map data already. Developing and using more sophisticated map-matching algorithms is orthogonal to our learning algorithm.

\subsection{Geometrically Guided Segmentation Refinement}
\label{sec:mgcda:guidance}
In the following presentation of our guided segmentation refinement for dark images using corresponding daytime images, we drop for brevity the subscript which was used to indicate this correspondence. In the conference version of this paper, the specific formulation of the guidance function $G$ for our refinement approach which was introduced in a general form in \eqref{eq:guided:prediction} was
\begin{equation} \label{eq:guidance:explicit}
G\left(\phi^z(I^z), I^z, \phi^1(I^1)\right) = R\left(\phi^z(I^z),\,B(\phi^1(I^1), I^z)\right),
\end{equation}
i.e.\ the composition of a cross bilateral filter $B$ on the daytime predictions, which aligns them to the dark image, with a fusion function $R$, which adaptively combines the aligned daytime predictions with the initial dark image predictions to refine the latter. In this extended version, we propose an improved, geometry-aware formulation for the alignment of the daytime predictions to the dark image, which explicitly incorporates the respective two-view geometry and performs the alignment by \emph{warping} the daytime predictions to the viewpoint of the dark image. The specification of the guidance function $G$ in the newly proposed, geometrically guided refinement is
\begin{equation} \label{eq:guidance:geom:explicit}
G\left(\phi^z(I^z), I^z, \phi^1(I^1), I^1\right) = R\left(\phi^z(I^z),\,Q(\phi^1(I^1), I^1, I^z)\right),
\end{equation}
where the fusion function $R$ is the same as in \eqref{eq:guidance:explicit} and the cross bilateral filter $B$ has been replaced by a warping function $Q$ which maps the daytime predictions to the dark view $I^z$. This warping function can be further analyzed as
\begin{equation}\label{eq:warp:general}
Q(\phi^1(I^1), I^1, I^z) = W(\phi^1(I^1),\,d(I^1),\,\delta{}T(I^1, I^z)),
\end{equation}
where $d(I^1) = d^1$ is the estimated depth map for the daytime image and $\delta{}T(I^1, I^z)$ is the estimated camera motion between the daytime view and the dark view. In the remainder of this section, we present the details of the individual modules of our guided segmentation refinement corresponding to functions $B$, $Q$ and $R$.

\subsubsection{Cross Bilateral Filter for Prediction Alignment}
\label{sec:mgcda:guidance:bilateral}

The correspondences between real images that are used in MGCDA are not perfect, in the sense that they are not aligned at a pixel-accurate level. Therefore, to leverage the prediction for the daytime image $I^{1}$ as \emph{guidance} for refining the respective prediction for the dark image $I^{z}$, it is necessary to first align the former prediction to $I^{z}$. To this end, we operate on \emph{soft} predictions and define a cross bilateral filter on the initial soft prediction map $\mathbf{S}^1 = \phi^{1}(I^1)$ which uses the color of the dark image $I^{z}$ as reference:
\begin{align}
&\tilde{\mathbf{S}}^{1}(\mathbf{p}) \nonumber\\
&{=}\;\frac{\displaystyle\sum_{\mathbf{q} \in \mathcal{N}(\mathbf{p})} G_{\sigma_s}(\vecnorm{\mathbf{q} - \mathbf{p}}) G_{\sigma_r}(\vecnorm{I^{z}(\mathbf{q}) - I^{z}(\mathbf{p})}) \mathbf{S}^{1}(\mathbf{q}) }{ \displaystyle\sum_{\mathbf{q} \in \mathcal{N}(\mathbf{p})} G_{\sigma_s}(\vecnorm{\mathbf{q} - \mathbf{p}}) G_{\sigma_r}(\vecnorm{I^{z}(\mathbf{q}) - I^{z}(\mathbf{p})}) }. \label{eq:cross:bilateral}
\end{align}
In \eqref{eq:cross:bilateral}, $\mathbf{p}$ and $\mathbf{q}$ denote pixel positions, $\mathcal{N}(\mathbf{p})$ is the neighborhood of $\mathbf{p}$, $G_{\sigma_s}$ is the spatial-domain Gaussian kernel and $G_{\sigma_r}$ is the color-domain kernel. The definition of the filter implies that only pixels $\mathbf{q}$ with similar color to the examined pixel $\mathbf{p}$ in the dark image $I^{z}$ contribute to the output $\tilde{\mathbf{S}}^{1}(\mathbf{p})$, which shifts salient edges in the initial daytime prediction to their correct position in the dark image. For the color-domain kernel, we use the CIELAB version of $I^{z}$, as it is more appropriate for measuring color similarity~\cite{bilateral:grid}. We set the spatial parameter $\sigma_s$ to $80$ to account for large misalignment, and $\sigma_r$ to $10$ following~\cite{bilateral:grid,SynRealDataFogECCV18}. In different settings from ours, $\sigma_s$ needs to be scaled proportionally to the image dimensions.

\subsubsection{Depth-Based Warping for Prediction Alignment}
\label{sec:mgcda:guidance:warping}

\begin{figure*}
    \centering
    \subfloat[Dark image $I^z$]{\includegraphics[width=0.24\textwidth]{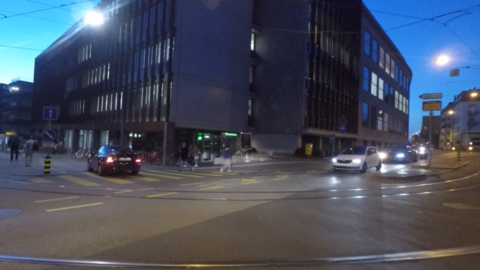}\label{fig:prediction:alignment:twilight}}
    \hfil
    \subfloat[Initial prediction $\mathbf{S}^z$ for $I^z$]{\includegraphics[width=0.24\textwidth]{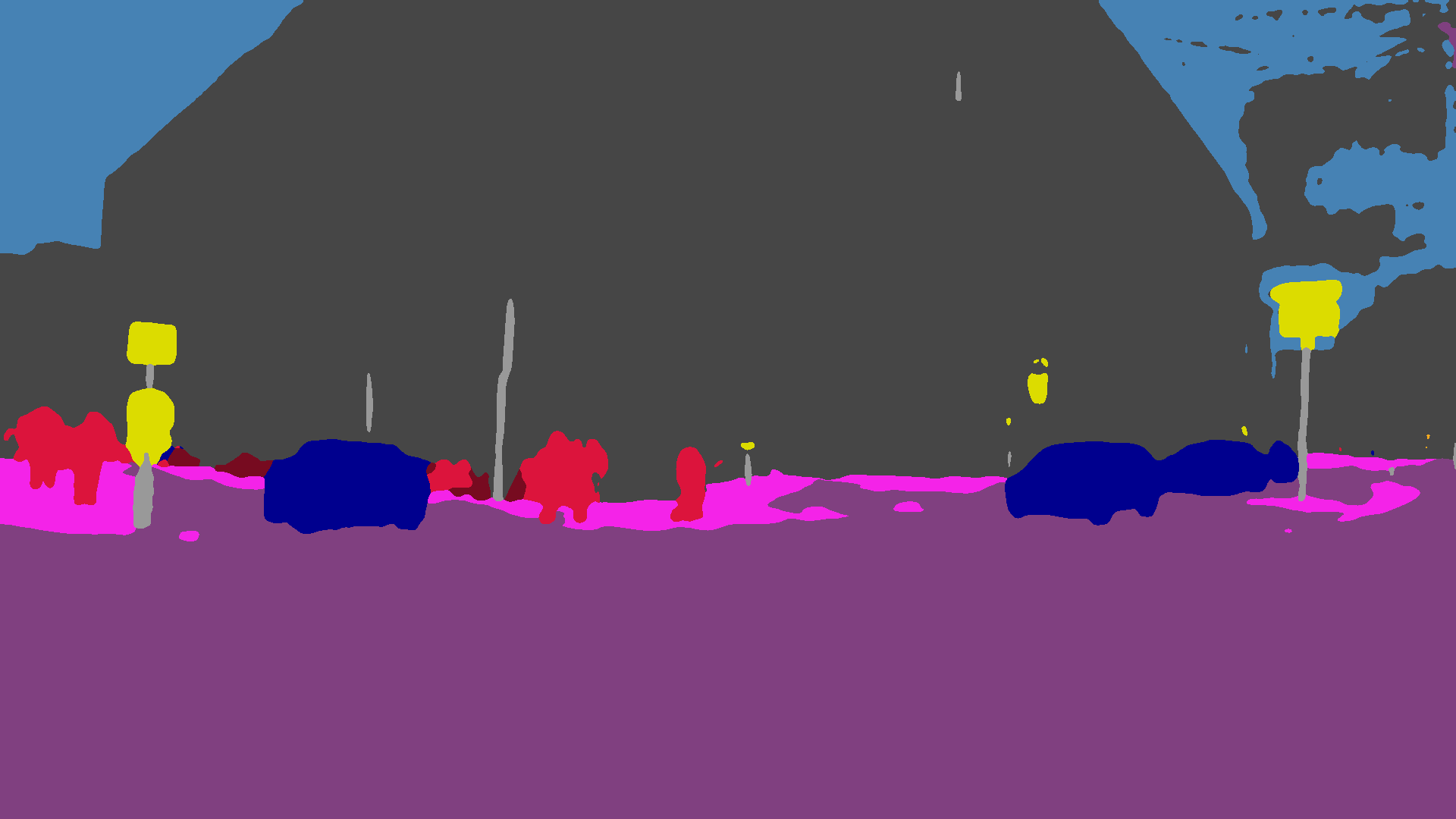}\label{fig:prediction:alignment:twilight:init}}
    \hfil
    \subfloat[Daytime image $I^{1}$]{\includegraphics[width=0.24\textwidth]{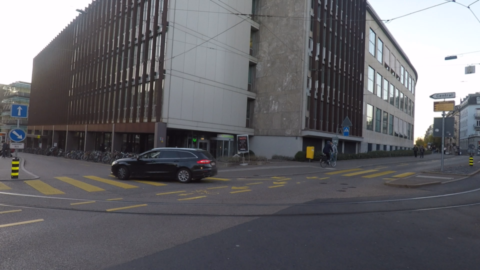}\label{fig:prediction:alignment:day}}
    \hfil 
    \vspace{-3mm}
    \subfloat[Initial prediction $\mathbf{S}^1$ for $I^1$]{\includegraphics[width=0.24\textwidth]{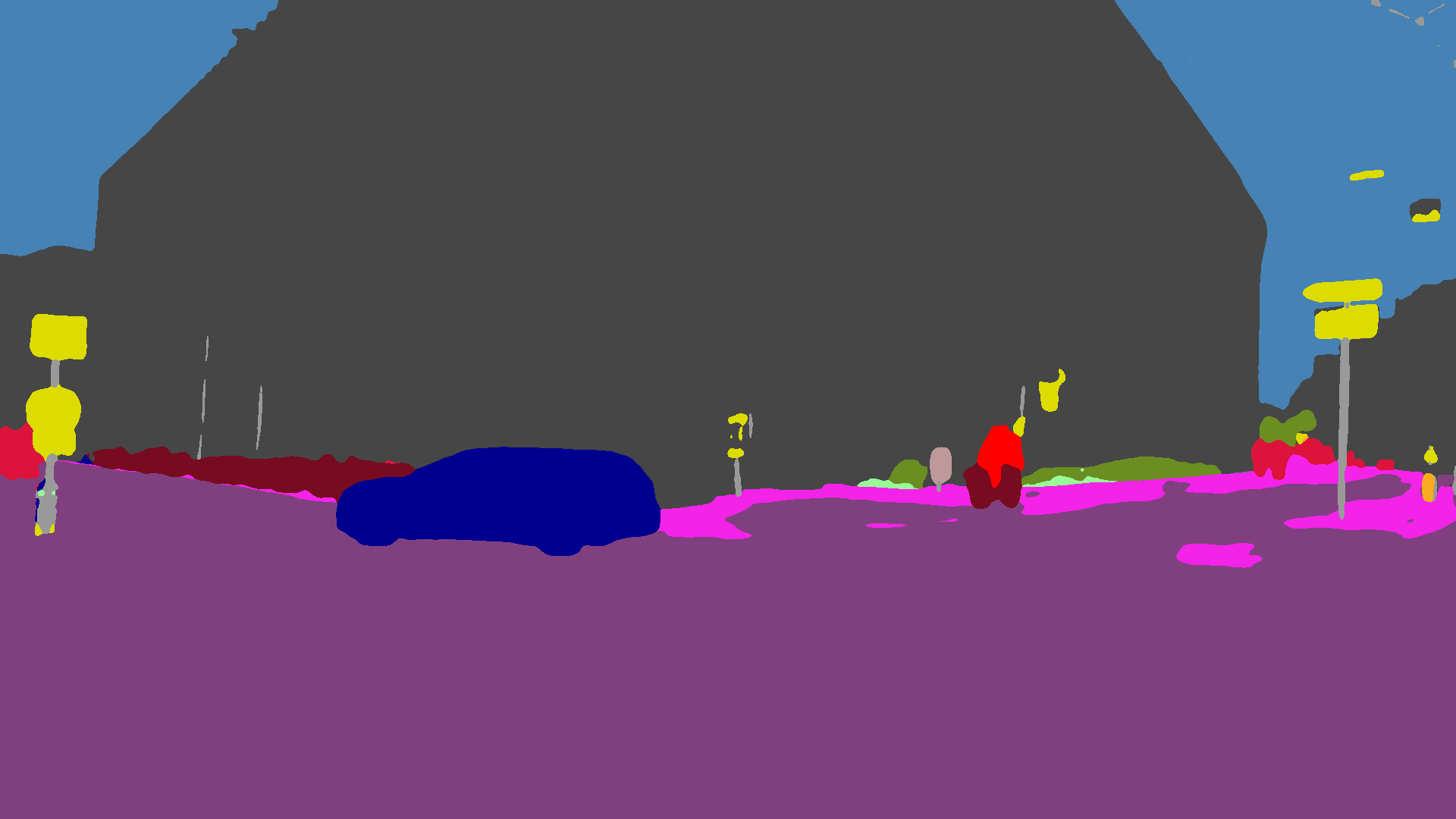}\label{fig:prediction:alignment:day:init}}
    \hfil
    \subfloat[Aligned prediction $\tilde{\mathbf{S}}^{1}$ for $I^1$ with cross bilateral filter]{\includegraphics[width=0.24\textwidth]{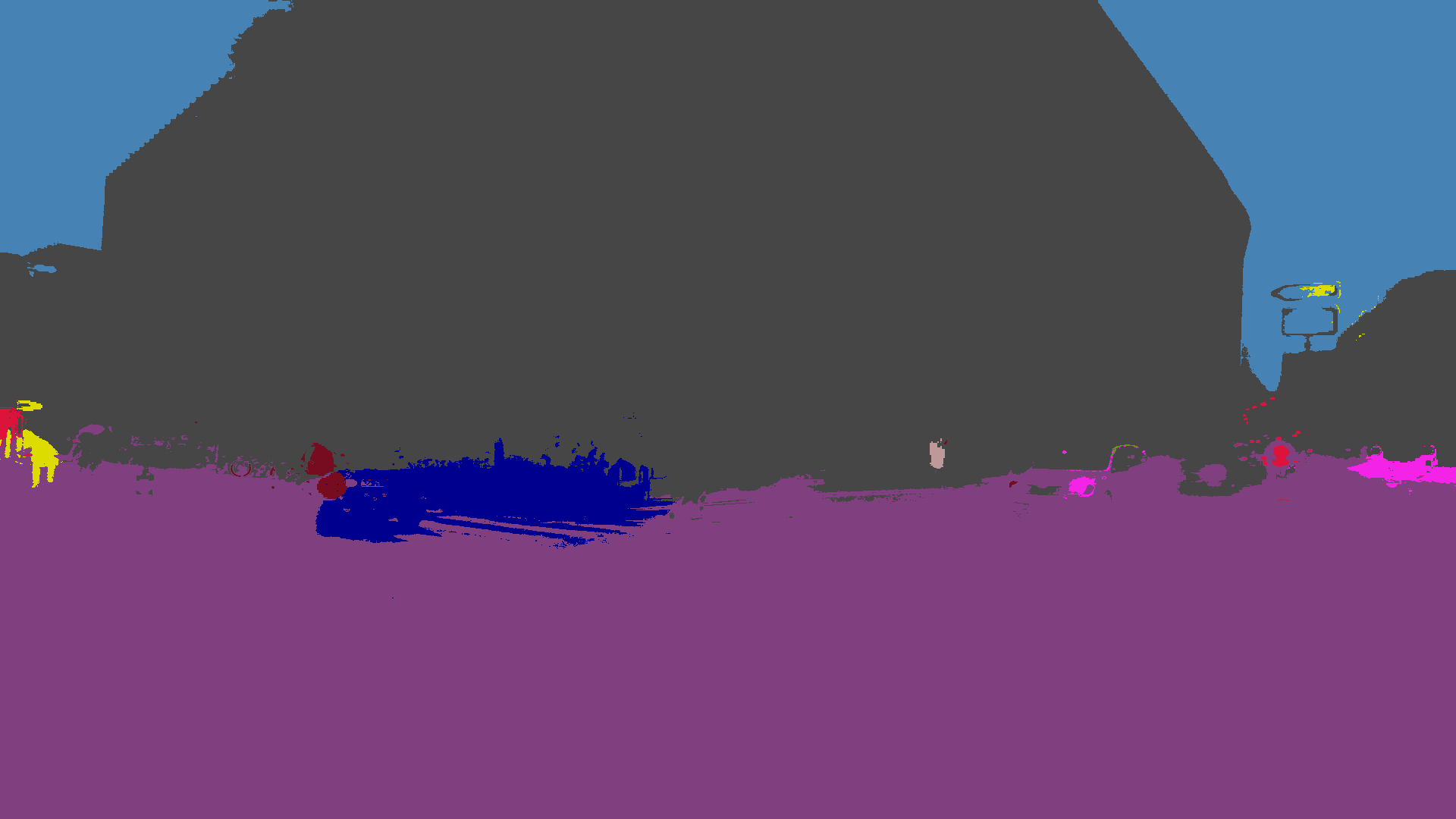}\label{fig:prediction:alignment:day:bilateral}}
    \hfil
    \subfloat[Aligned prediction $\tilde{\mathbf{S}}^{1}$ for $I^1$ with depth-based warping]{\includegraphics[width=0.24\textwidth]{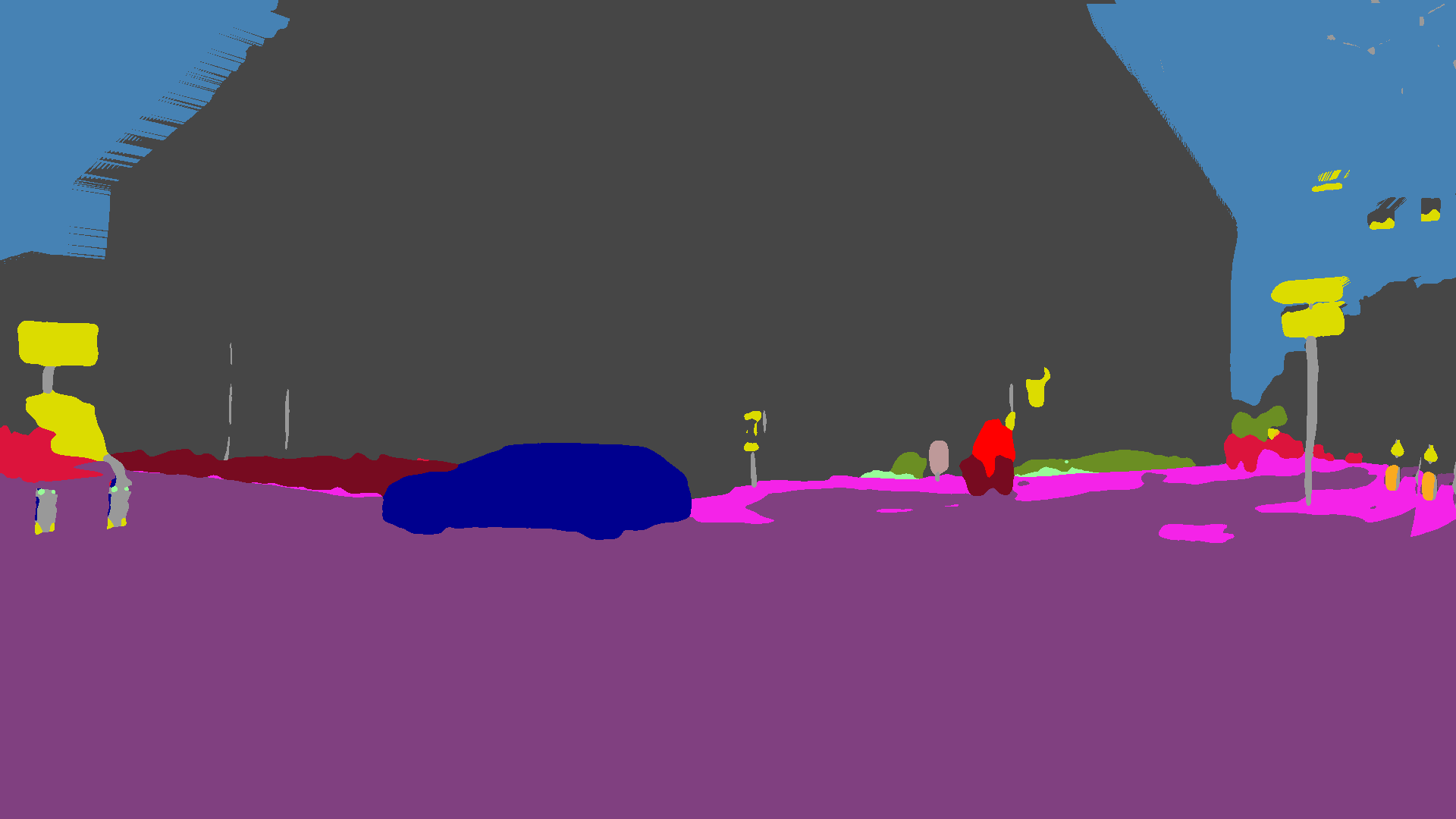}\label{fig:prediction:alignment:day:warping}}
    \hfil
    \subfloat[Refined prediction $\hat{\mathbf{S}}^{z}$ for $I^z$ with cross bilateral filter]{\includegraphics[width=0.24\textwidth]{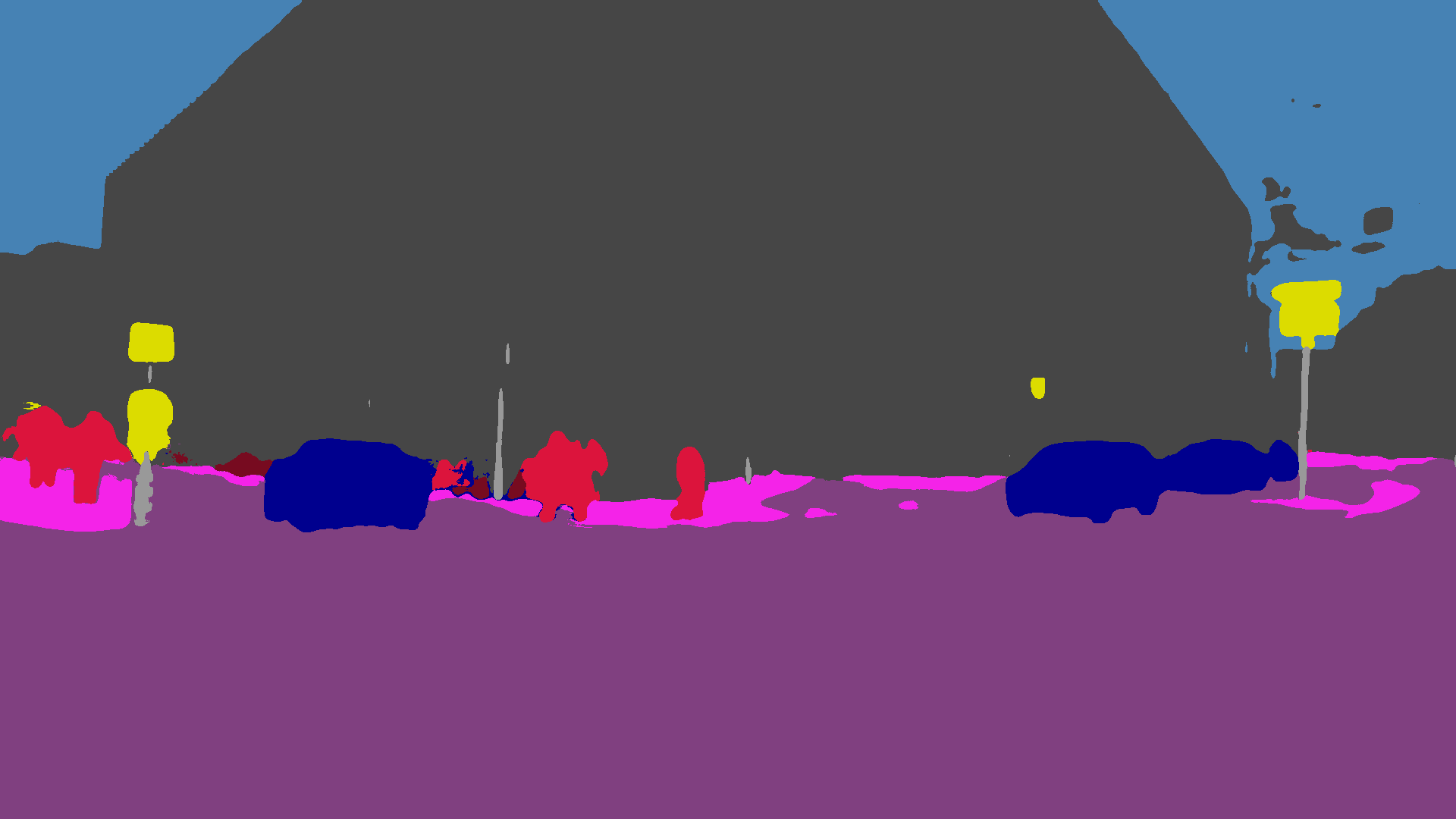}\label{fig:prediction:alignment:twilight:bilateral}}
    \hfil
    \subfloat[Refined prediction $\hat{\mathbf{S}}^{z}$ for $I^z$ with depth-based warping]{\includegraphics[width=0.24\textwidth]{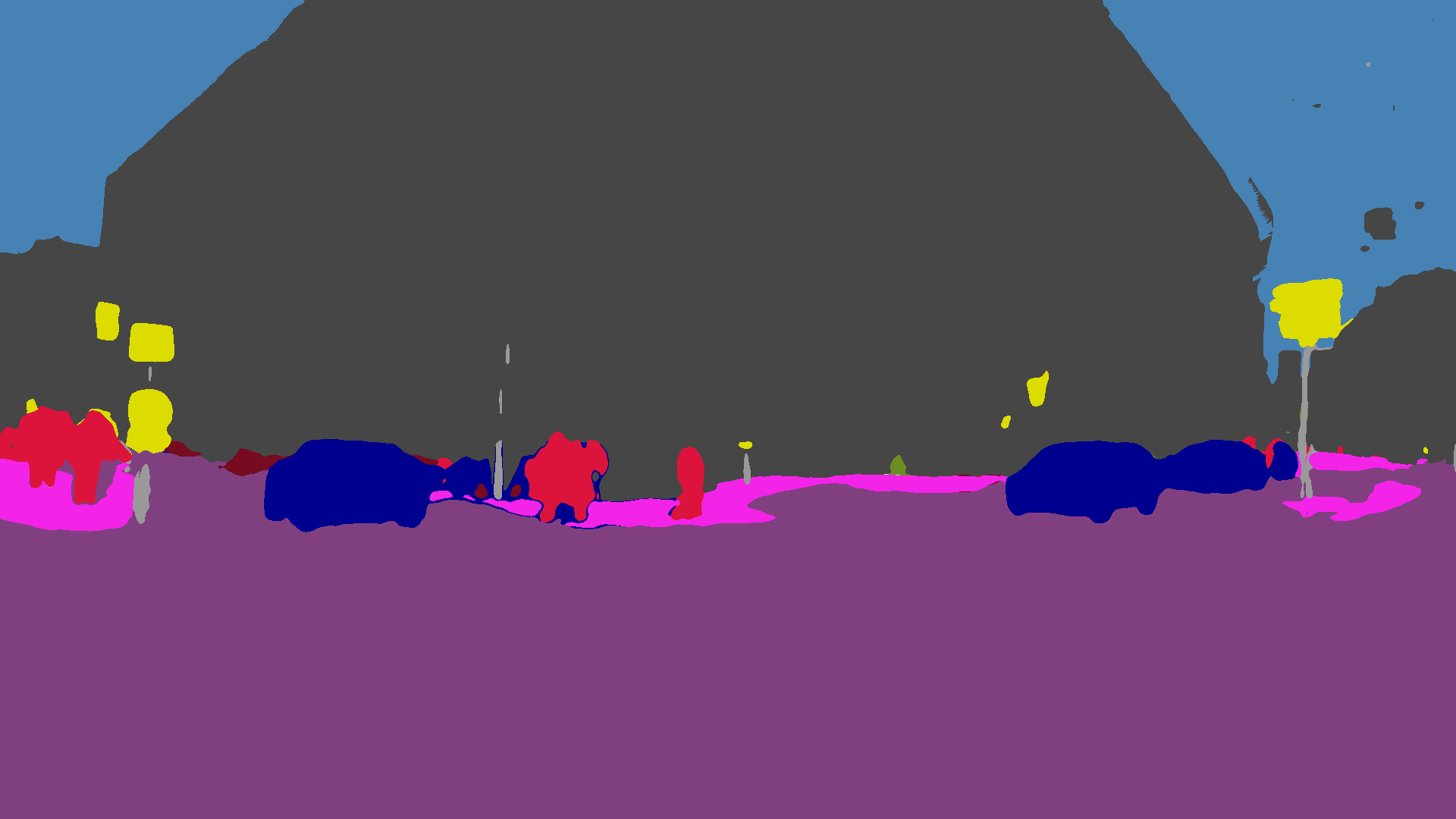}\label{fig:prediction:alignment:twilight:warping}}
    \caption{Comparison of guided segmentation refinement using prediction alignment with cross bilateral filter versus depth-based warping on an example pair of corresponding images from \emph{Dark Zurich}. Best viewed on a screen with zoom.}
    \label{fig:prediction:alignment}
\end{figure*}

An important drawback of the above prediction alignment approach with a cross bilateral filter is its \emph{uniform} operation on all image regions, despite the fact that the magnitude of misalignment between corresponding points in the two views varies across the image, depending on the depth of the examined points as well as the particular camera motion between the two views. Furthermore, in case the magnitude of misalignment is larger than the diameter of the respective ground-truth semantic segment, there is no common support between the regions this segment occupies in $\mathbf{S}^1$ and $I^z$, inevitably leading to erroneous outputs of the filter on small objects.

In order to address this issue, we explicitly model the \emph{two-view geometry} which pertains to the daytime image and the dark image at hand, and use the estimated camera motion together with the depth map for the daytime view to apply a dense pixel-level warping of the daytime predictions to the target viewpoint that corresponds to the dark image. In this way, we are able to capture the diverse magnitudes and directions of pixel flow within the image, aligning the daytime predictions accurately in a dense pixel-level fashion. We illustrate this process in Fig.~\ref{fig:map:matching:image:warping}\subref{fig:image:warping}.

More formally, we first establish dense correspondences from the pixel grid of the daytime image to that of the dark image. Consider a pixel $\mathbf{p} = (x_p,\,y_p)^T$ in the daytime image. We denote the depth value at this pixel by $d^1(\mathbf{p})$. Moreover, we denote the calibration matrices for the two views by $\mathbf{K}^1$ and $\mathbf{K}^z$, and the transformation from the coordinate system of the daytime view to that of the dark view, which models camera motion $\delta{}T(I^1, I^z)$, by $(\mathbf{R}|\mathbf{t})$. The point $\mathbf{p}^{\prime}$ in the dark image that corresponds to $\mathbf{p}$ is identified by first back-projecting $\mathbf{p}$ into 3D space and then reprojecting this 3D point to the dark view, which can be expressed as
\begin{equation}\label{eq:corresp:day:dark}
\left(
\begin{array}{c}
\mathbf{p}^{\prime}\\
1
\end{array}
\right) \sim \mathbf{K}^z (\mathbf{R}|\mathbf{t})
\left(
\begin{array}{c}
d^1(\mathbf{p}) {\left(\mathbf{K}^1\right)}^{-1}
\left(
\begin{array}{c}
\mathbf{p}\\
1
\end{array}
\right)\\
1
\end{array}
\right),
\end{equation}
where $\sim$ denotes equality up to scale. These dense 2D-2D correspondences enable us to warp the soft prediction map $\mathbf{S}^1$ for the daytime image $I^1$ to the viewpoint of the dark image $I^z$. Note that although inverse warping, which would use the depth map of the target dark view, is known to perform better in the literature~\cite{monocular:neural:rendering} and to avoid discretization artifacts, we choose instead to apply forward warping, which uses the depth map of the source daytime view, as shown in \eqref{eq:corresp:day:dark}. The reasoning behind this choice is that a ground-truth depth map is not available for either of the views and the consequent monocular depth estimation outputs much more reliable results on the easier, daytime domain of the source view $I^1$ than on the adverse, dark domain of $I^z$.

In particular, we apply this forward warping by defining a quadrilateral mesh on the pixel grid of $I^1$ using 4-connectivity to form the quads. This mesh is deformed through \eqref{eq:corresp:day:dark}, which results in fractional coordinates for the quad vertices in general. For each pixel $\mathbf{p}$ in the dark view, we assign it the quad $q$ that contains it and calculate the warped soft prediction $\tilde{\mathbf{S}}^{1}(\mathbf{p})$ by performing bilinear interpolation of the soft predictions on the four original vertices of $q$ in the daytime view. The bilinear weights are defined by the position of $\mathbf{p}$ inside the deformed $q$.

One of the challenges that occur in the aforementioned setting are potential fold-backs of the mesh, which correspond to \emph{occlusions} of visible parts of $I^1$ in $I^z$. In this case, for an affected pixel there are more than one candidate quads that contain it, corresponding to the occluder and one or more occludees. We solve this ambiguity by assigning to the pixel the quad with the smallest depth value $d^1$ (computed as the average over all quad vertices), which corresponds to the occluder that is actually visible in the target dark view. Another challenge is related to \emph{disocclusion}, in which case the assigned quads of disoccluded pixels are elongated across the direction perpendicular to the associated depth edge. We handle this case as follows: we first detect such irregular quads in the deformed mesh by calculating the ratio of lengths of the second longest side of each quad to its third longest side; if this ratio is larger than $5$, the quad is deemed as irregular. These quads accept a clear binary cut which corresponds to removing their two longest sides and reflects the depth discontinuity within them. Since the content of disoccluded pixels is generally similar to the region in the source view which is on the \emph{distant} side of the associated depth edge, we modify the interpolation weights so that only those quad vertices which are on the side of the aforementioned cut with the \emph{larger} values for $d^1$ are considered for the interpolation. Finally, for pixels in $I^z$ that are not assigned any quad, which implies that they are invisible in $I^1$, the soft prediction is directly copied from the pixel with the same position in $\mathbf{S}^1$, based on the prior that the mean of the distribution of camera motion between the two views is zero.

We now elaborate further on the implementation details for computing the dense correspondences between $I^1$ and $I^z$ established via \eqref{eq:corresp:day:dark}. As far as the depth map $d^1$ is concerned, we use a pre-trained Monodepth2~\cite{monodepth2} model (trained with stereo supervision at $1024\times{}320$ resolution) to obtain an absolute estimate for $d^1$. We process this estimate further by setting all pixels for which $\phi^1$ predicts \emph{sky} to the maximum possible value for Monodepth2, which is $d_{\text{max}} = 540\text{ m}$. In order to compute the camera motion between the two views, we first apply SURF~\cite{SURF} to extract keypoints and respective descriptors in both images. The descriptors for $I^z$ are matched with their nearest neighbors in $I^1$, using three criteria for rejecting matches. In particular, (1) we only accept mutual nearest neighbors, (2) we apply a threshold $\theta_{\text{sec}} = 0.7$ to the ratio of squared Euclidean distances of each nearest neighbor to the respective second nearest neighbor, and (3) we apply a threshold $\theta_{\text{rel}} = 20$ to the ratio of squared Euclidean distances of the current match to the match with globally minimum distance. After identifying the putative matches, we run the 7-point RANSAC algorithm to compute the final inlier set of matches and obtain the fundamental matrix $\mathbf{F}$, using $1000$ iterations and an inlier threshold of $t = 2$ pixels for RANSAC. We then compute the essential matrix $\mathbf{E} = {\left(\mathbf{K}^z\right)}^T \mathbf{F} \mathbf{K}^1$; this step assumes that the camera is calibrated for both images. $\mathbf{E}$ is decomposed into the rotational component $\mathbf{R}$ and the translational component $\mathbf{t}$ of the camera motion we are after, only that $\mathbf{t}$ is determined at this point only up to scale. We recover the scale of $\mathbf{t}$ by triangulating the matched points and estimating the median scaling factor that needs to be applied to their $Z$-coordinates so that these match the respective values of the depth map $d^1$.

We compare the two approaches for prediction alignment on a pair of a twilight image and a corresponding daytime image from \emph{Dark Zurich} in Fig.~\ref{fig:prediction:alignment}, where we depict hard predictions for easier visualization. As can be seen from Fig.~\ref{fig:prediction:alignment}\subref{fig:prediction:alignment:day:bilateral} and \ref{fig:prediction:alignment}\subref{fig:prediction:alignment:day:warping}, the depth-based warping preserves small-scale objects such as traffic signs and poles well in the aligned prediction $\tilde{\mathbf{S}}^{1}$, in contrast to the cross bilateral filter which completely extinguishes such objects due to its inability to handle misalignments that are very large relative to the object's scale.

\subsubsection{Confidence-Adaptive Prediction Fusion}
\label{sec:mgcda:guidance:fusion}

The final step in our refinement approach, which is applied after either of the preceding prediction alignment approaches, is to fuse the aligned prediction $\tilde{\mathbf{S}}^{1}$ for $I^{1}$ with the initial prediction $\mathbf{S}^{z} = \phi^z(I^{z})$ for $I^{z}$ in order to obtain the refined prediction $\hat{\mathbf{S}}^{z}$, the hard version of which is subsequently used in training. We propose an adaptive fusion scheme, which uses the \emph{confidence} associated with the two predictions at each pixel to weigh their contribution in the output and addresses disagreements due to dynamic content by properly adjusting the fusion weights. Let us denote the confidence of the aligned prediction $\tilde{\mathbf{S}}^{1}$ for $I^{1}$ at pixel $\mathbf{p}$ by $F^{1}(\mathbf{p}) = \max_{c \in \mathcal{C}} \tilde{S}_c^1(\mathbf{p})$ and respectively the confidence of the initial prediction $\mathbf{S}^{z}$ for $I^{z}$ by $F^{z}(\mathbf{p})$. Our confidence-adaptive fusion is then defined as
\begin{equation} \label{eq:fusion}
\hat{\mathbf{S}}^{z} = \frac{F^{z}}{F^{z} + \alpha F^{1}} \mathbf{S}^{z} + \frac{\alpha F^{1}}{F^{z} + \alpha F^{1}} \tilde{\mathbf{S}}^{1},
\end{equation}
where $0 < \alpha = \alpha(\mathbf{p}) \leq 1$ may vary and we have completely dropped the pixel argument $\mathbf{p}$ for brevity. In this way, we allow the daytime image prediction to have a greater effect on the output at regions of the dark image which were not easy for model $\phi^z$ to classify, while preserving the initial prediction $\mathbf{S}^{z}$ at lighter regions of the dark image where $\mathbf{S}^{z}$ is more reliable.

Our fusion distinguishes between dynamic and static scene content by regulating $\alpha$. In particular, $\alpha$ downweights $\tilde{\mathbf{S}}^{1}$ to induce a preference towards $\mathbf{S}^{z}$ when both predictions have high confidence. However, apart from imperfect alignment, the two scenes also differ due to dynamic content. Intuitively, the prediction of a dynamic object in the daytime image should be assigned an even lower weight in case the corresponding prediction in the dark image does not agree, since this object might only be present in the former scene. More formally, we denote the subset of $\mathcal{C}$ that includes dynamic classes by $\mathcal{C}_d$ and define
\begin{align}
&\alpha(\mathbf{p}) \nonumber\\
&{=}\;\left\{
\begin{array}{cl}
    \alpha_l, & \text{if } c_1 = \arg\displaystyle\max_{c \in \mathcal{C}} \tilde{S}_c^1(\mathbf{p}) \in \mathcal{C}_d \text{ and } S_{c_1}^z(\mathbf{p}) \leq \eta \\
     & \text{ or } c_2 = \arg\displaystyle\max_{c \in \mathcal{C}} S_c^z(\mathbf{p}) \in \mathcal{C}_d \text{ and } \tilde{S}_{c_2}^1(\mathbf{p}) \leq \eta, \\
    \alpha_h & \text{otherwise.}
\end{array}
\right.
\label{eq:alpha}
\end{align}
The majority of pixels falls in the second case of \eqref{eq:alpha} and the choice of $\alpha_h$ is important, as it involves a trade-off between exploiting the daytime prediction and avoiding overly relying on it. In our experiments, we \emph{manually} tune $\alpha_l = 0.3$, $\alpha_h = 0.6$ and $\eta = 0.2$ on a couple of training images (no grid search), which implies that our method is not sensitive to the exact value of these parameters. An illustration of the effect of $\alpha_h$ on the prediction fusion results is included in Appendix~\ref{supp:sec:parameters}. Comparative results of our complete guided segmentation refinement for the two prediction alignment approaches are shown in Fig.~\ref{fig:prediction:alignment}\subref{fig:prediction:alignment:twilight:bilateral} and \ref{fig:prediction:alignment}\subref{fig:prediction:alignment:twilight:warping}. Note the improved correction of the \emph{sky} region on the top right part of the image as well as the better preservation of fine objects such as distant traffic signs achieved with depth-based warping.

\section{Uncertainty-Aware Evaluation}
\label{sec:evaluation} 

Images taken under adverse conditions such as nighttime contain invalid regions, i.e.\ regions with indiscernible semantic content. Invalid regions are closely related to the concept of negative test cases which was considered in~\cite{wilddash}. However, invalid regions constitute intra-image entities and can co-exist with valid regions in the same image, whereas a negative test case refers to an entire image that should be treated as invalid. We build upon the evaluation of~\cite{wilddash} for negative test cases and generalize it to be applied uniformly to all images in the evaluation set, whether they contain invalid regions or not. Our annotation and evaluation framework includes invalid regions in the set of evaluated pixels, but treats them differently from valid regions to account for the high uncertainty of their content. In the following, we elaborate on the generation of ground-truth annotations using privileged information through the day-night correspondences of our dataset and present our UIoU metric.

\subsection{Annotation with Privileged Information}
\label{sec:evaluation:annotation}

\begin{figure*}
    \centering
    \subfloat[Input image $I$]{\includegraphics[width=0.24\textwidth]{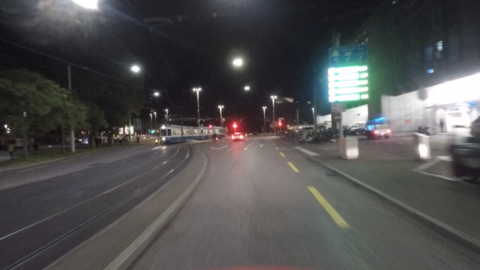}\label{fig:annotation:input}}
    \hfil
    \subfloat[Auxiliary image $I^{\prime}$]{\includegraphics[width=0.24\textwidth]{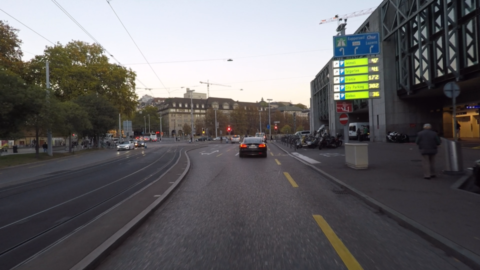}\label{fig:annotation:auxiliary}}
    \hfil
    \subfloat[GT invalid mask $J$]{\includegraphics[width=0.24\textwidth]{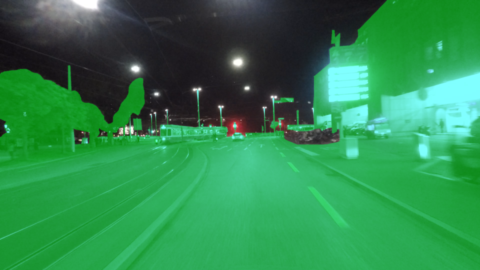}\label{fig:annotation:invalid}}
    \hfil
    \subfloat[GT semantic labeling $H$]{\includegraphics[width=0.24\textwidth]{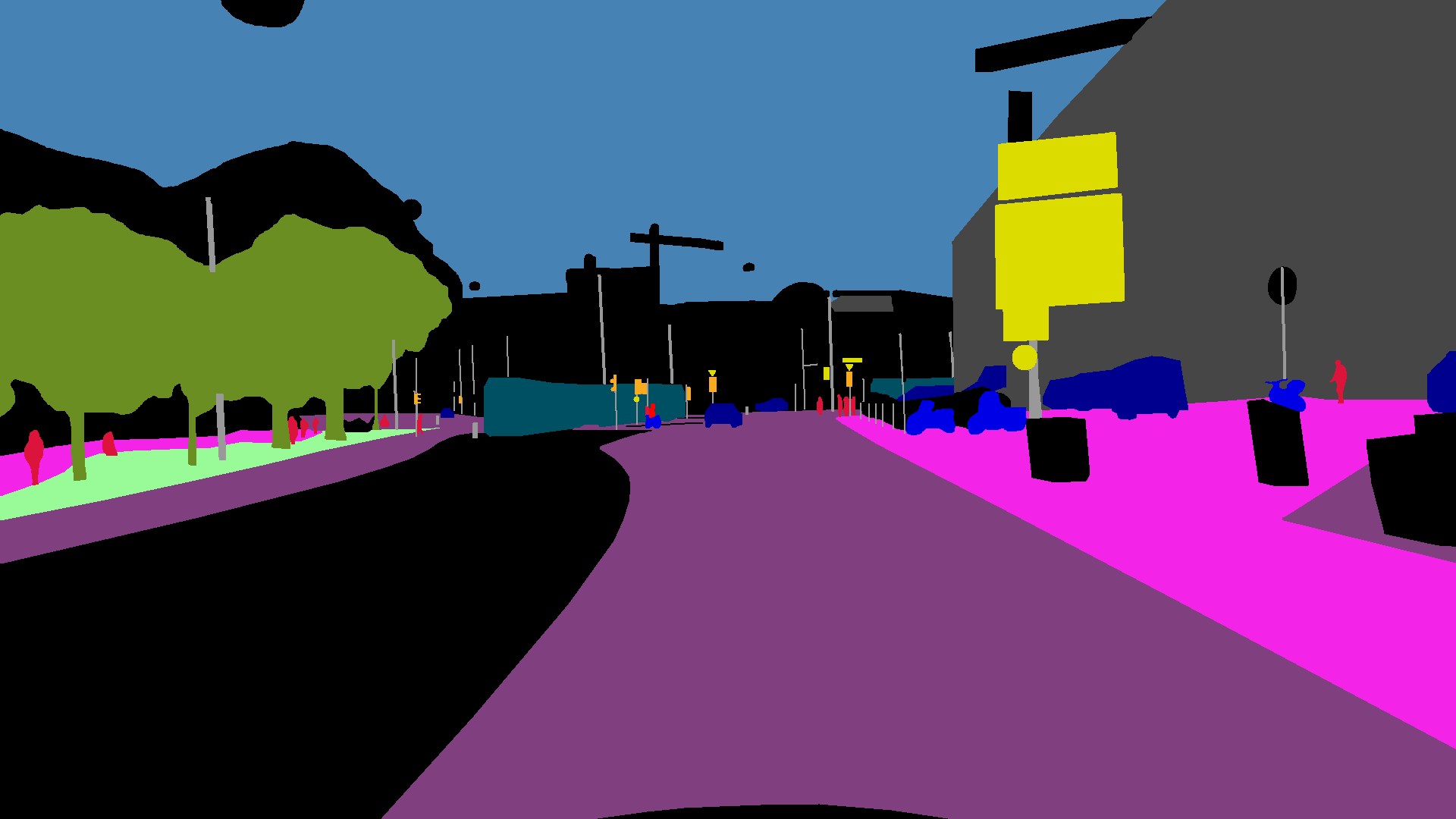}\label{fig:annotation:gt}}
    \caption{Example input images from \emph{Dark Zurich-test} and output annotations with our protocol. Valid pixels in $J$ are marked green.}
    \label{fig:annotation}
\end{figure*}

For each image $I$, the annotation process involves two steps: 1) creation of the ground-truth invalid mask $J$, and 2) creation of the ground-truth semantic labeling $H$.

For the semantic labels, we consider a predefined set $\mathcal{C}$ of $C$ classes, which is equal to the set of Cityscapes~\cite{Cityscapes} evaluation classes ($C = 19$). The annotator is first presented only with $I$ and is asked to mark the valid regions in it as the regions which she can unquestionably assign to one of the $C$ classes or declare as not belonging to any of them. The result of this step is the invalid mask $J$, which is set to 0 at valid pixels and 1 at invalid pixels.

Secondly, the annotator is asked to mark the semantic labels of $I$, only that this time she also has access to an \emph{auxiliary} image $I^{\prime}$. This latter image has been captured with roughly the same 6D camera pose as $I$ but under more favorable conditions. In our dataset, $I^{\prime}$ is captured at daytime whereas $I$ is captured at nighttime. The large overlap of static scene content between the two images allows the annotator to label certain regions in $H$ with a legitimate semantic label from $\mathcal{C}$, even though the same regions have been annotated as invalid (and are kept as such) in $J$. This allows joint evaluation on valid and invalid regions, as it creates regions which can accept both the \emph{invalid} label and the ground-truth label from $\mathcal{C}$ as correct predictions. Due to the imperfect match of the camera poses for $I$ and $I^{\prime}$, the labeling of invalid regions in $H$ is done conservatively, marking a coarse boundary which may leave unlabeled zones around the true semantic boundaries in $I$, so that no pixel is assigned a wrong label. The parts of $I$ which remain indiscernible even after inspection of $I^{\prime}$ are left unlabeled in $H$. These parts as well as instances of classes outside $\mathcal{C}$ are not considered during evaluation. We illustrate a visual example of our annotation inputs and outputs in Fig.~\ref{fig:annotation}.

\subsection{Uncertainty-Aware Predictions}

The semantic segmentation prediction that is fed to our evaluation is expected to include pixels labeled as \emph{invalid}. Instead of defining a separate, explicit \emph{invalid} class, which would potentially require the creation of new training data to incorporate this class, we allow a more flexible approach for soft predictions with the original set of semantic classes by using a \emph{confidence threshold}, which affords an evaluation curve for our UIoU metric by varying this threshold.

In particular, we assume that the evaluated method outputs an intermediate soft prediction $\mathbf{S}(\mathbf{p})$ at each pixel $\mathbf{p}$ as a probability distribution among the $C$ classes, which is subsequently converted to a hard assignment by outputting the class $\tilde{H}(\mathbf{p}) = \arg\max_{c \in \mathcal{C}}\{S_c(\mathbf{p})\}$ with the highest probability. In this case, $S_{\tilde{H}(\mathbf{p})}(\mathbf{p}) \in [1/C,\,1]$ is the effective confidence associated with the prediction. This assumption is not very restrictive, as most recent semantic segmentation methods are based on CNNs with a softmax layer that outputs such soft predictions.

The final evaluated output $\hat{H}$ is computed based on a free parameter $\theta \in [1/C,\,1]$ which acts as a confidence threshold by invalidating those pixels where the confidence of the prediction is lower than $\theta$, i.e.\ $\hat{H}(\mathbf{p}) = 
\tilde{H}(\mathbf{p})$ if $S_{\tilde{H}(\mathbf{p})}(\mathbf{p}) \geq \theta$ and \emph{invalid} otherwise. Increasing $\theta$ results in more pixels being predicted as \emph{invalid}. This approach is motivated by the fact that ground-truth invalid regions are identified during annotation by the uncertainty of their semantic content, which implies that a model should ideally place lower confidence (equivalently higher uncertainty) in predictions on invalid regions than on valid ones, so that the former get invalidated for lower values of $\theta$ than the latter. The formulation of our UIoU metric rewards this behavior as we shall see next. Note that our evaluation does not strictly require soft predictions, as UIoU can be normally computed for fixed, hard predictions $\hat{H}$.

\subsection{UIoU}

We propose UIoU as a generalization of the standard IoU metric for evaluation of semantic segmentation predictions which may contain pixels labeled as \emph{invalid}. UIoU reduces to standard IoU if no pixel is predicted to be invalid, e.g.\ when $\theta = 1/C$.

The calculation of UIoU for class $c$ involves five sets of pixels, which are listed along with their symbols: true positives (TP), false positives (FP), false negatives (FN), true invalids (TI), and false invalids (FI). Based on the ground-truth invalid masks $J$, the ground-truth semantic labelings $H$ and the predicted labels $\hat{H}$ for the set of evaluation images, these five sets are defined as follows:
\begin{align}
\text{TP} &= \{\mathbf{p}: H(\mathbf{p}) = \hat{H}(\mathbf{p}) = c\}, \label{eq:tp}\\
\text{FP} &= \{\mathbf{p}: H(\mathbf{p}) \neq c \text{ and } \hat{H}(\mathbf{p}) = c\}, \label{eq:fp}\\
\text{FN} &= \{\mathbf{p}: H(\mathbf{p}) = c \text{ and } \hat{H}(\mathbf{p}) \notin \{c,\,\text{\emph{invalid}}\}\}, \label{eq:fn}\\
\text{TI} &= \{\mathbf{p}: H(\mathbf{p}) = c \text{ and } \hat{H}(\mathbf{p}) = \text{\emph{invalid} and } J(\mathbf{p}) = 1\}, \label{eq:ti}\\
\text{FI} &= \{\mathbf{p}: H(\mathbf{p}) = c \text{ and } \hat{H}(\mathbf{p}) = \text{\emph{invalid} and } J(\mathbf{p}) = 0\}. \label{eq:fi}
\end{align}
UIoU for class $c$ is then defined as
\begin{equation} \label{eq:uiou}
\text{UIoU} = \frac{|\text{TP}| + |\text{TI}|}{|\text{TP}| + |\text{TI}| + |\text{FP}| + |\text{FN}| + |\text{FI}|}.
\end{equation}
Note that a true invalid prediction results in equal reward to predicting the correct semantic label of the pixel. Moreover, an invalid prediction does not come at no cost: it incurs the same penalty on valid pixels as predicting an incorrect label.

When dealing with multiple classes, we modify our notation to $\text{UIoU}^{(c)}$ (similarly for the five sets of pixels related to class $c$), which we avoided in the previous definitions to reduce clutter. The overall semantic segmentation performance on the evaluation set is reported as the mean UIoU over all $C$ classes. By varying the confidence threshold $\theta$ and using the respective output, we obtain a parametric expression $\text{UIoU}(\theta)$. When $\theta = 1/C$, no pixel is predicted as invalid and thus $\text{UIoU}(1/C) = \text{IoU}$.

We motivate the usage of UIoU instead of standard IoU in case the test set includes ground-truth invalid masks by showing in Th.~\ref{thm:UIoU:greater:iou} that UIoU is guaranteed to be larger than IoU for some $\theta > 1/C$ under the assumption that predictions on invalid regions are associated with lower confidence than those on valid regions, which lies in the heart of our evaluation framework. The proof is in Appendix~\ref{supp:sec:proof}.

\begin{thm} \label{thm:UIoU:greater:iou}
Assume that there exist $\theta_1$, $\theta_2$ such that $\theta_1 < \theta_2$, $\forall p: J(p) = 1 \Rightarrow S_{\tilde{H}(p)}(p) \leq \theta_1$ and $J(p) = 0 \Rightarrow S_{\tilde{H}(p)}(p) \geq \theta_2$. If we additionally assume that $\exists p \in \text{\emph{FN}}^{(c)}(1/C) \cup \text{\emph{FP}}^{(c)}(1/C): J(p) = 1$, then $\text{\emph{IoU}}^{(c)} < \text{\emph{UIoU}}^{(c)}(\theta_1)$.
\end{thm}

\section{The Dark Zurich Dataset}
\label{sec:dark:zurich}

\begin{figure*}[!tb]
    \centering
    \begin{tikzpicture}
    \tikzstyle{every node}=[font=\footnotesize]
    \begin{axis}[
      ybar,
      ymode=log,
      width=\textwidth,
      height=4.5cm,
      xmin=0,
      xmax=26,
      ymin=1e4,
      ymax=1e10,
      ymajorgrids=true,
      ylabel={number of pixels},
      ytick={1e4,1e6,1e8,1e10},
      yticklabels={$10^4$,$10^6$,$10^8$,$10^{10}$,},
      xtick={1.5,5,8.5,13.5,18,21,24.5},
      minor xtick={3,7,10,17,19,23},
      xticklabels = {
        flat,
        construction,
        nature,
        vehicle,
        sky,
        object,
        human,
      },
      major x tick style = {opacity=0},
      minor x tick num = 1,
      xtick pos=left,
      every node near coord/.append style={
      anchor=west,
      rotate=90,
      font=\scriptsize,
      }
    ]

    \addplot[bar shift=0pt,draw=road,          fill opacity=0.9,fill=road!80!white           , nodes near coords=road                 ] plot coordinates{ ( 1,     116888390  ) };
    \addplot[bar shift=0pt,draw=sidewalk,      fill opacity=0.8,fill=sidewalk!80!white       , nodes near coords=sidewalk             ] plot coordinates{ ( 2,     30808566   ) };

    \addplot[bar shift=0pt,draw=building,      fill opacity=0.8,fill=building!80!white       , nodes near coords=build.               ] plot coordinates{ ( 4,     80937811  ) };
    \addplot[bar shift=0pt,draw=fence,         fill opacity=0.8,fill=fence!80!white          , nodes near coords=fence                ] plot coordinates{ ( 5,     6376074   ) };
    \addplot[bar shift=0pt,draw=wall,          fill opacity=0.8,fill=wall!80!white           , nodes near coords=wall                 ] plot coordinates{ ( 6,     10001205  ) };

    \addplot[bar shift=0pt,draw=vegetation,    fill opacity=0.8,fill=vegetation!80!white     , nodes near coords=veget.               ] plot coordinates{ ( 8,    38594654   ) };
    \addplot[bar shift=0pt,draw=terrain,       fill opacity=0.8,fill=terrain!80!white        , nodes near coords=terrain              ] plot coordinates{ ( 9,    3415370    ) };

    \addplot[bar shift=0pt,draw=car,           fill opacity=0.8,fill=car!80!white            , nodes near coords=car           ] plot coordinates{ ( 11,    7442836   ) };
    \addplot[bar shift=0pt,draw=truck,         fill opacity=0.8,fill=truck!80!white          , nodes near coords=truck         ] plot coordinates{ ( 12,    57607     ) };
    \addplot[bar shift=0pt,draw=train,         fill opacity=0.8,fill=train!80!white          , nodes near coords=train         ] plot coordinates{ ( 13,    6316001        ) };
    \addplot[bar shift=0pt,draw=bus,           fill opacity=0.8,fill=bus!80!white            , nodes near coords=bus           ] plot coordinates{ ( 14,    36178     ) };
    \addplot[bar shift=0pt,draw=bicycle,       fill opacity=0.8,fill=bicycle!80!white        , nodes near coords=bicycle       ] plot coordinates{ ( 15,    326467      ) };
    \addplot[bar shift=0pt,draw=motorcycle,    fill opacity=0.8,fill=motorcycle!80!white     , nodes near coords=motorcycle    ] plot coordinates{ ( 16,    336434        ) };

    \addplot[bar shift=0pt,draw=sky,           fill opacity=0.8,fill=sky!80!white            , nodes near coords=sky                  ] plot coordinates{ ( 18,    58742600 ) };

    \addplot[bar shift=0pt,draw=pole,          fill opacity=0.8,fill=pole!80!white           , nodes near coords=pole                 ] plot coordinates{ ( 20,    3461555   ) };
    \addplot[bar shift=0pt,draw=traffic sign,  fill opacity=0.8,fill=traffic sign!80!white   , nodes near coords=traffic sign         ] plot coordinates{ ( 21,    1517787   ) };
    \addplot[bar shift=0pt,draw=traffic light, fill opacity=0.8,fill=traffic light!80!white  , nodes near coords=traffic light        ] plot coordinates{ ( 22,    421527    ) };

    \addplot[bar shift=0pt,draw=person,        fill opacity=0.8,fill=person!80!white         , nodes near coords=person        ] plot coordinates{ ( 24,    986273     ) };
    \addplot[bar shift=0pt,draw=rider,         fill opacity=0.8,fill=rider!80!white          , nodes near coords=rider         ] plot coordinates{ ( 25,    101965        ) };

    \end{axis}
    \end{tikzpicture}
    \caption{Number of annotated pixels per class in \emph{Dark Zurich}.}
    \label{fig:dataset:stats}
\end{figure*}
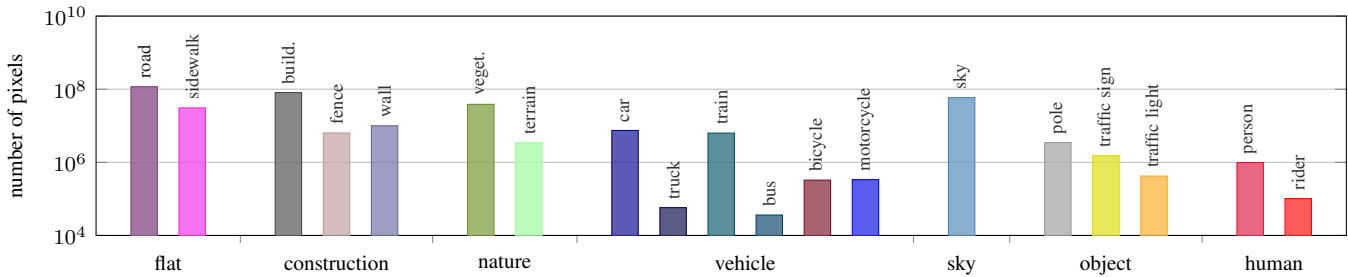

\begin{figure*}[!tb]
    \centering
    \subfloat{\includegraphics[width=0.195\textwidth]{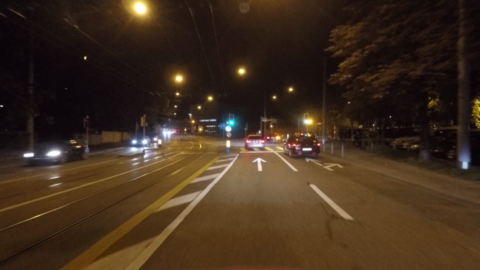}}
    \hfil
    \subfloat{\includegraphics[width=0.195\textwidth]{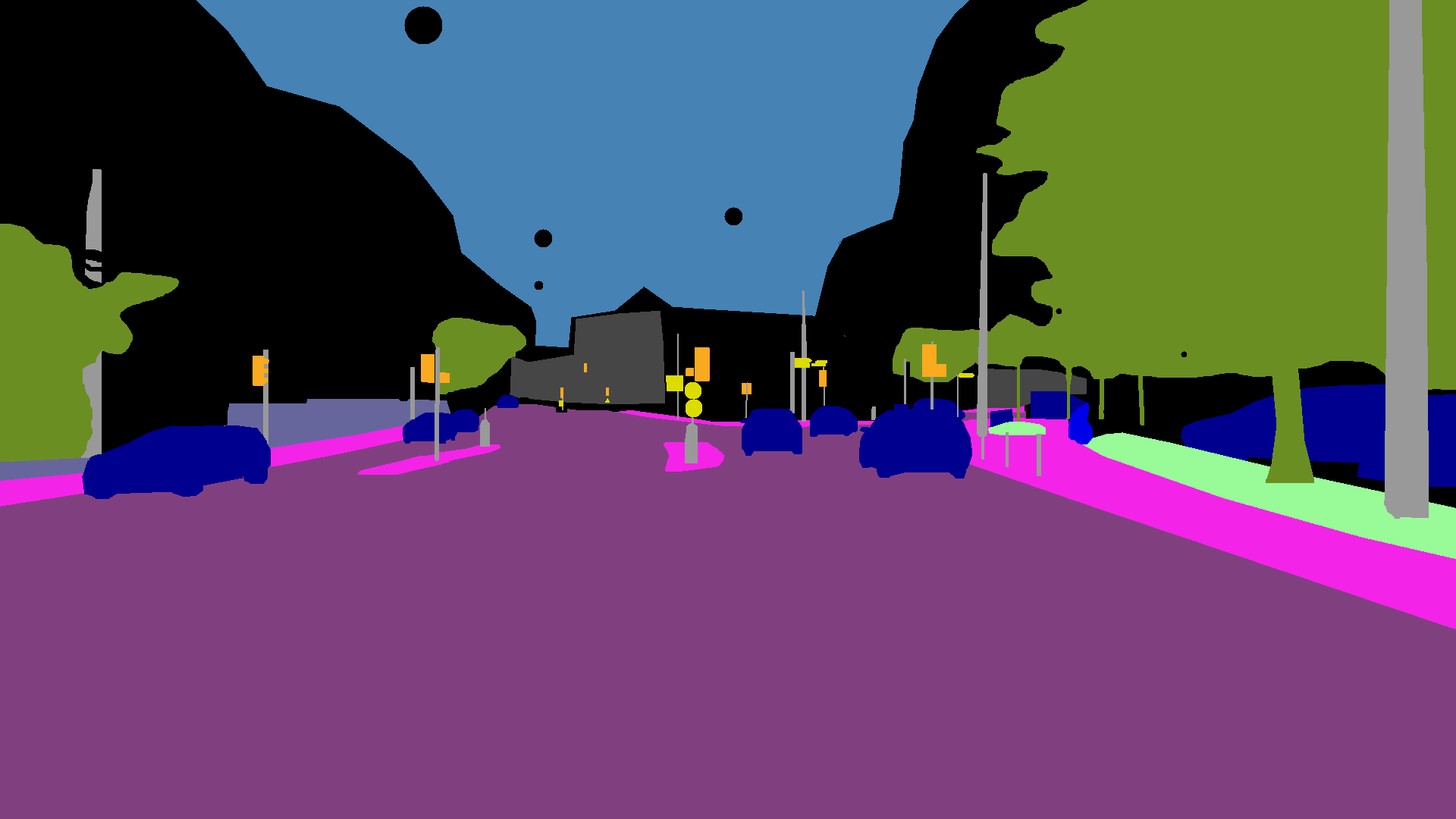}}
    \hfil
    \subfloat{\includegraphics[width=0.195\textwidth]{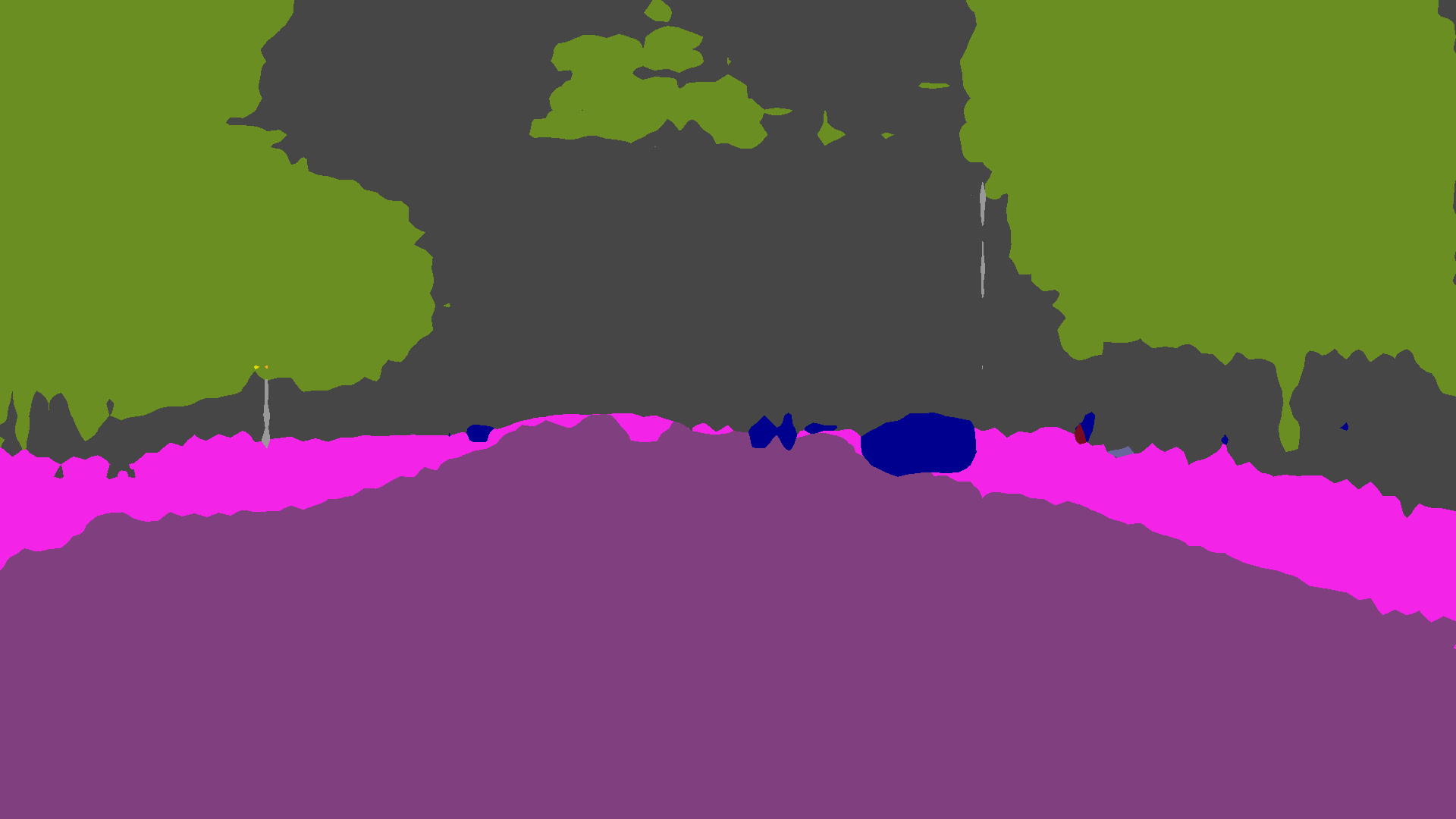}}
    \hfil
    \subfloat{\includegraphics[width=0.195\textwidth]{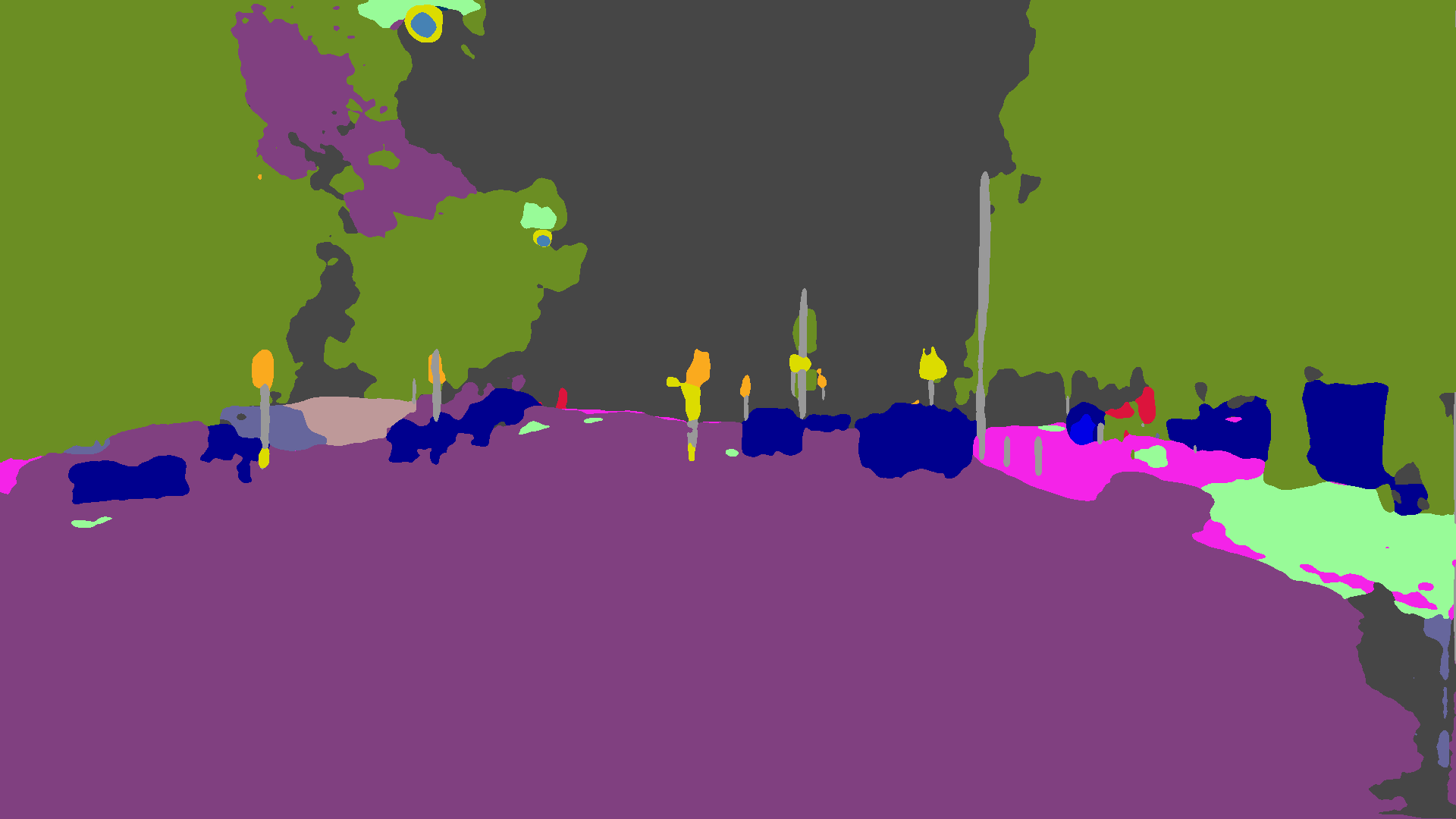}}
    \hfil
    \subfloat{\includegraphics[width=0.195\textwidth]{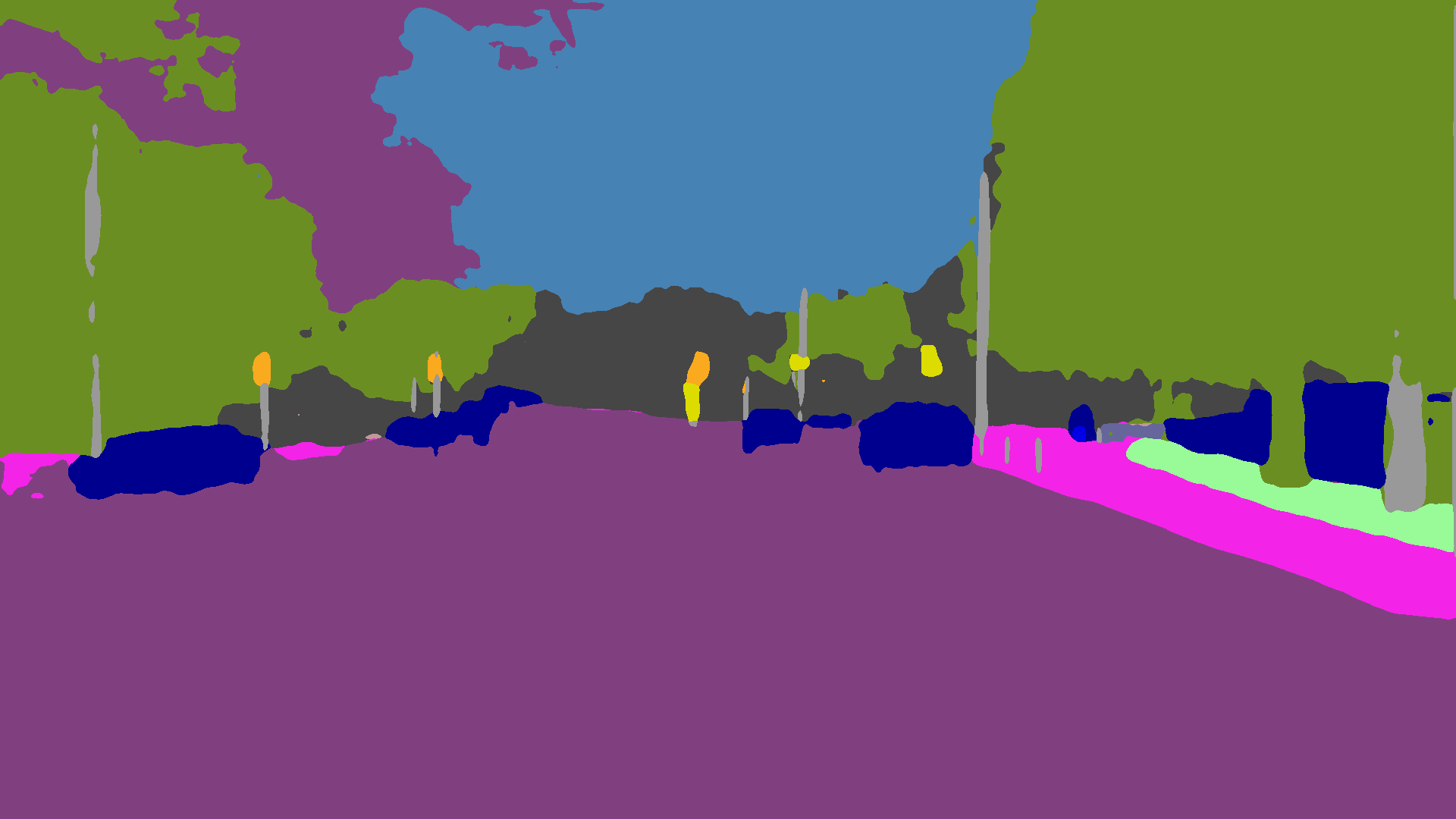}}
    \\
    \vspace{-0.3cm}
    \subfloat{\includegraphics[width=0.195\textwidth]{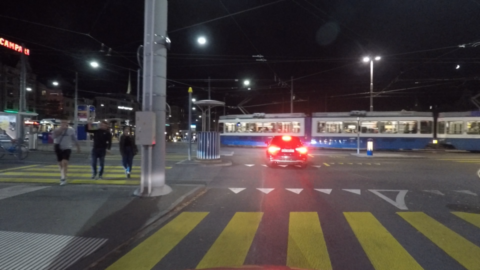}}
    \hfil
    \subfloat{\includegraphics[width=0.195\textwidth]{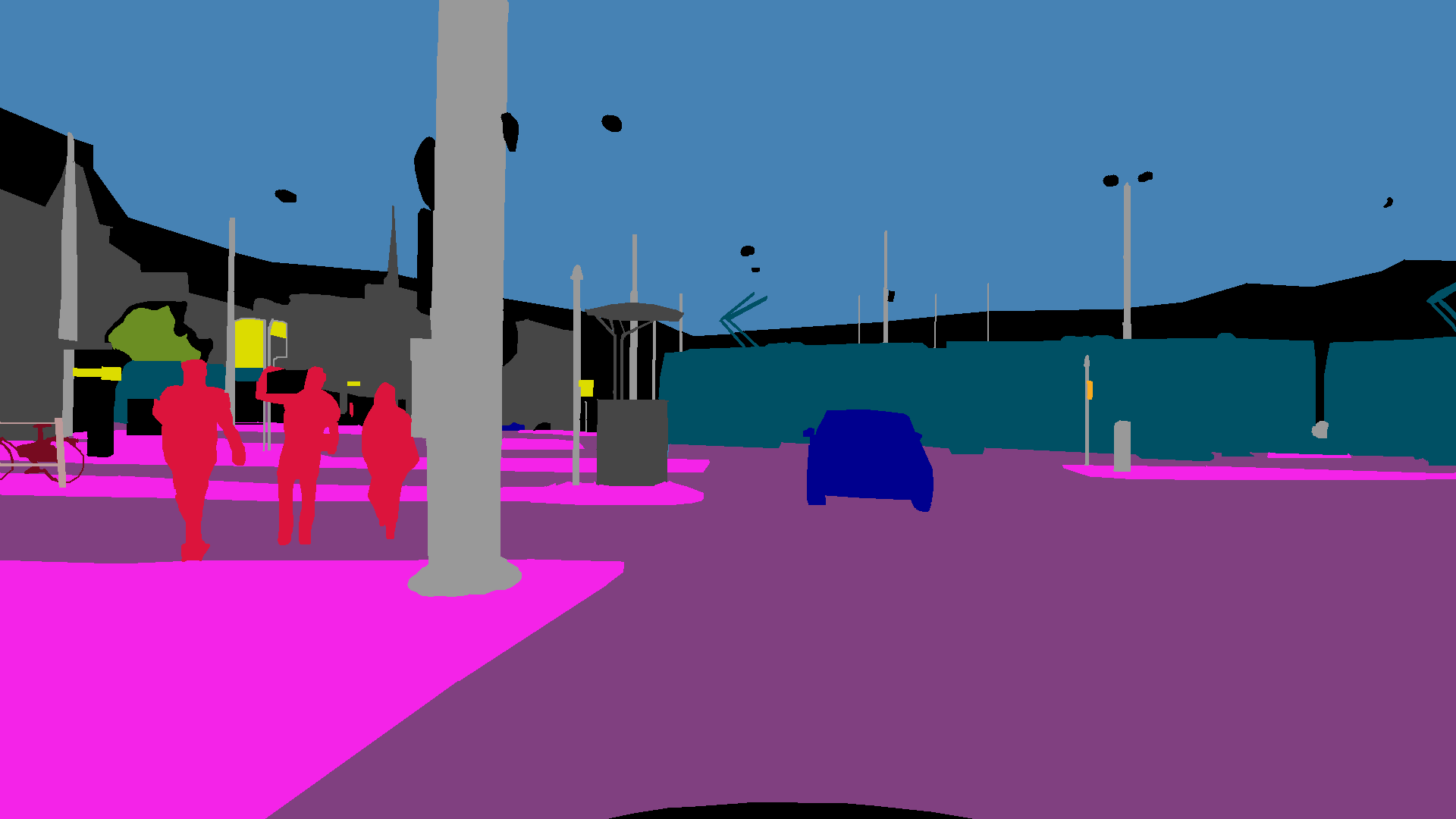}}
    \hfil
    \subfloat{\includegraphics[width=0.195\textwidth]{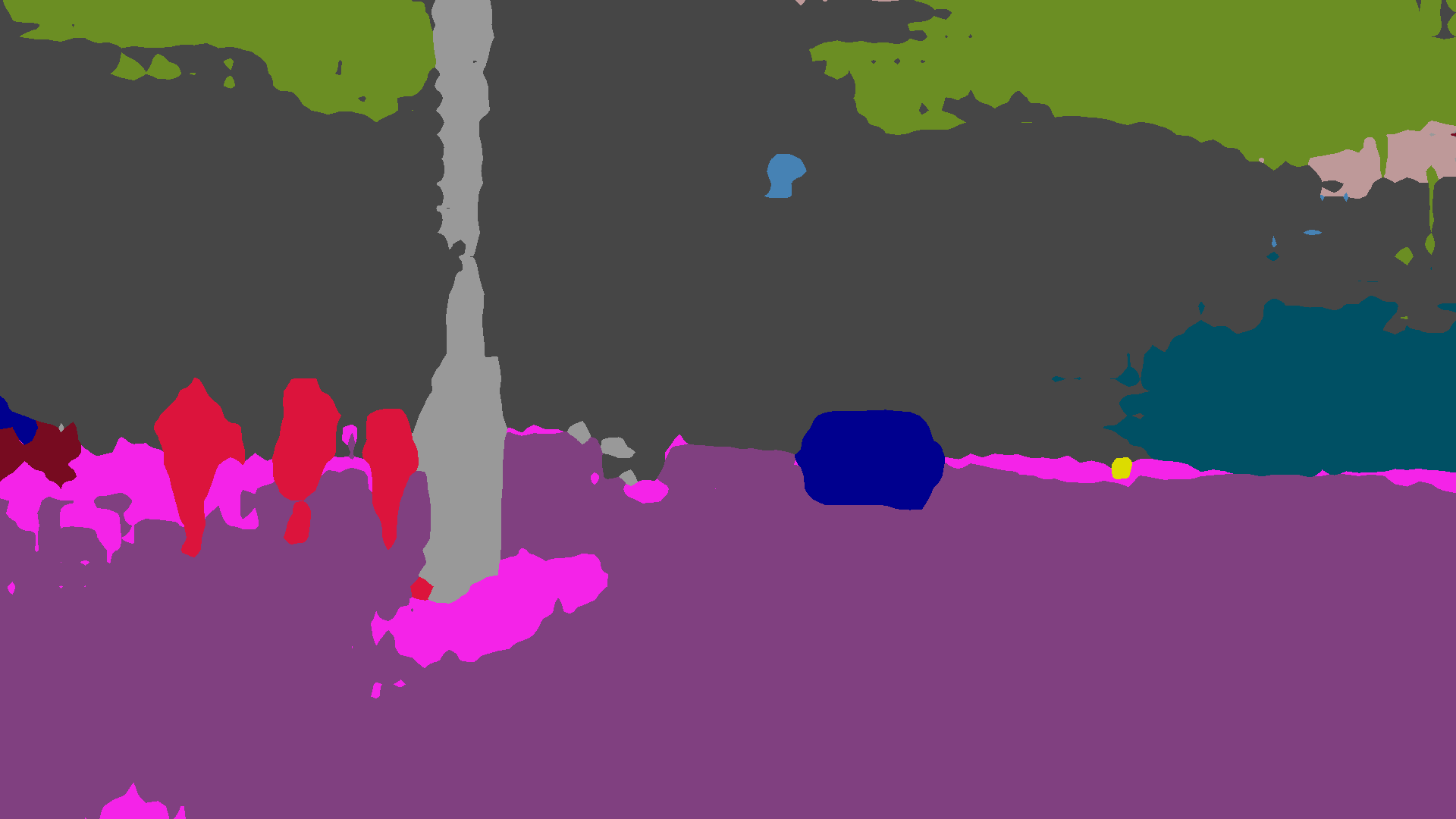}}
    \hfil
    \subfloat{\includegraphics[width=0.195\textwidth]{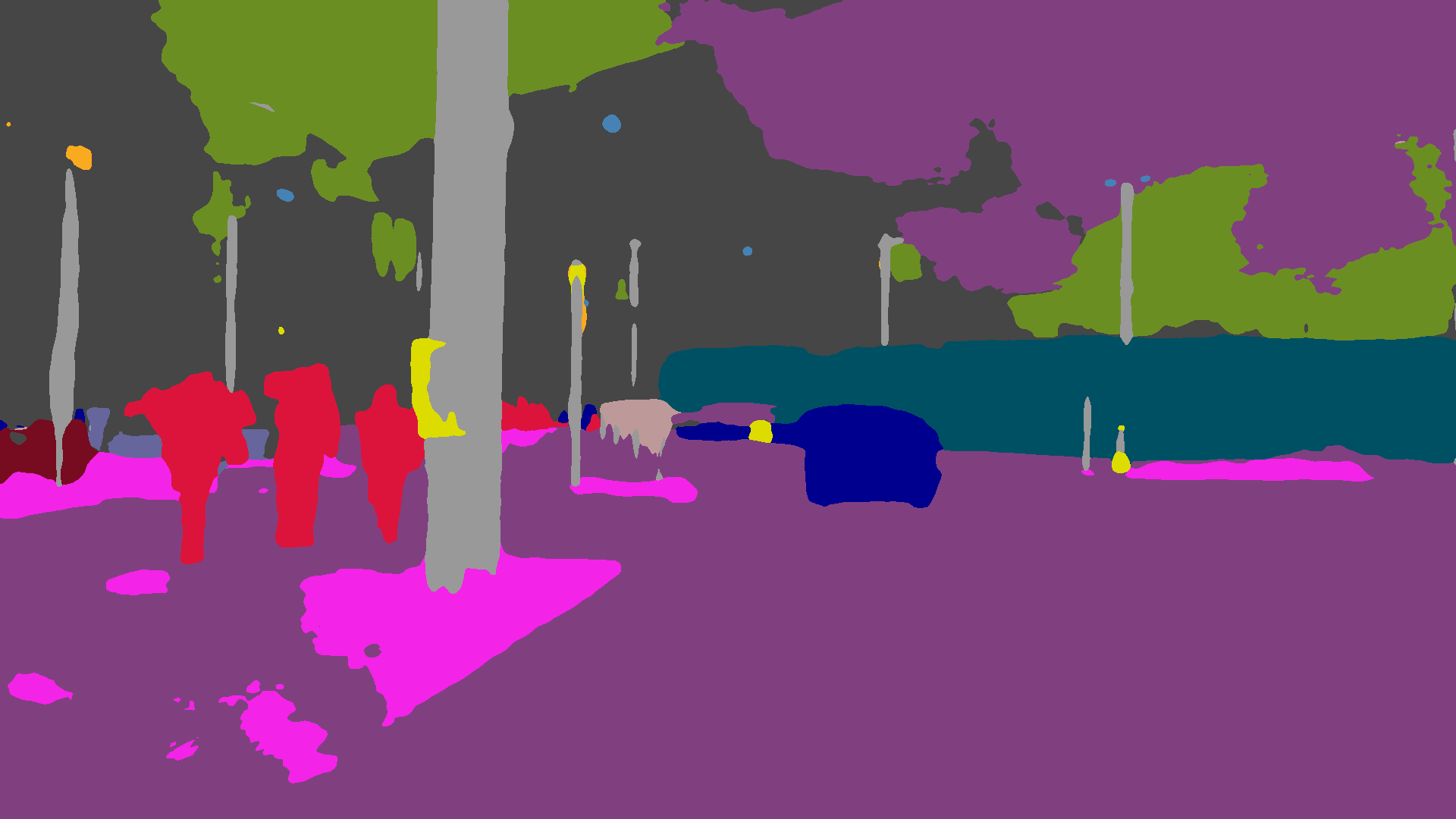}}
    \hfil
    \subfloat{\includegraphics[width=0.195\textwidth]{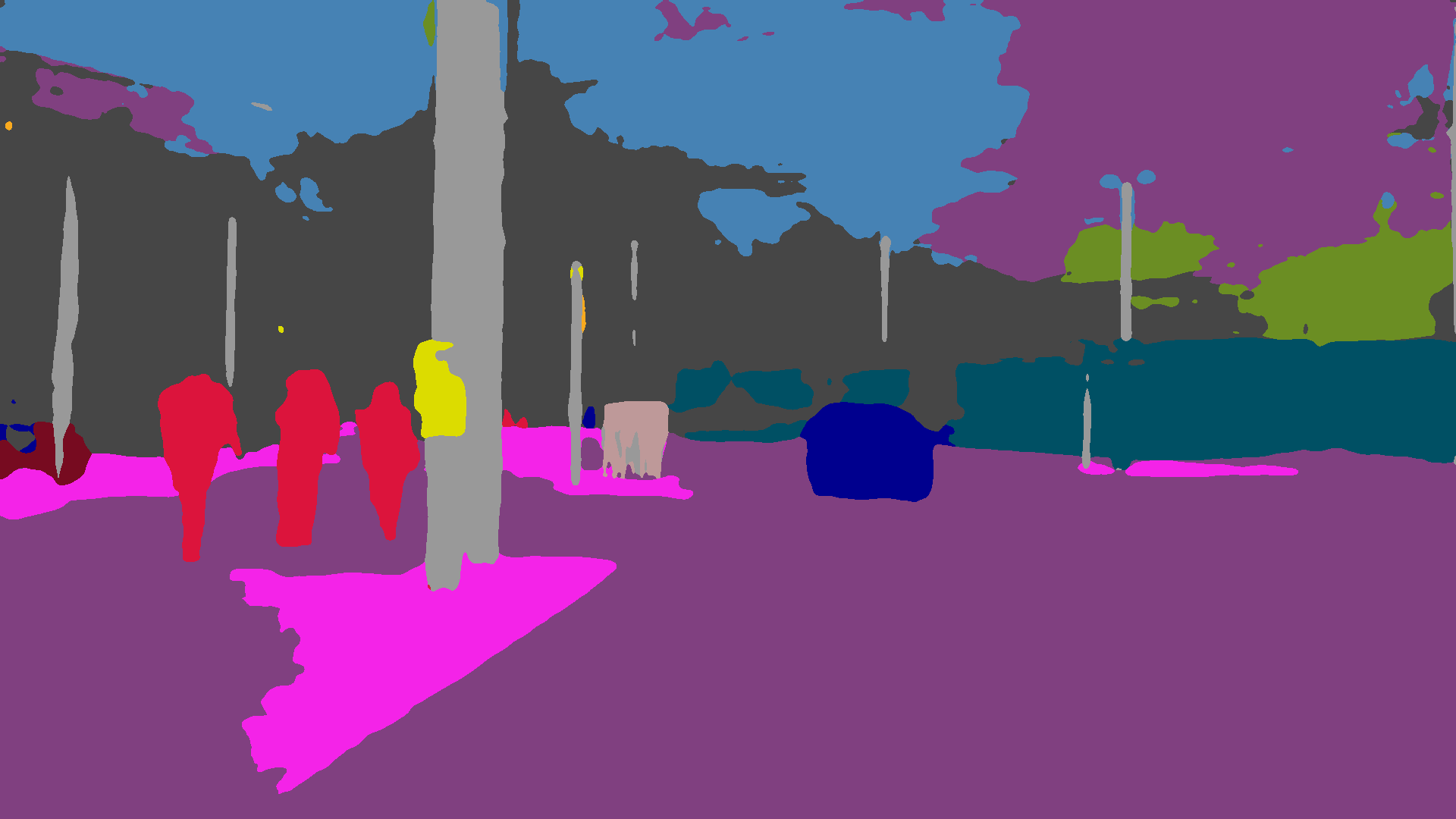}}
    \\
    \vspace{-0.3cm}
    \addtocounter{subfigure}{-10}
    \subfloat[Image]{\includegraphics[width=0.195\textwidth]{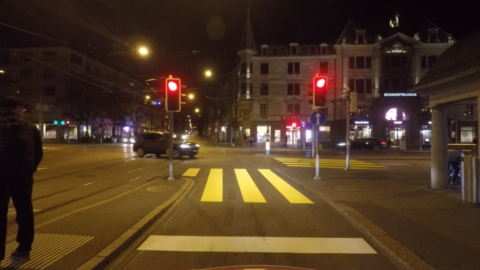}}
    \hfil
    \subfloat[Semantic GT]{\includegraphics[width=0.195\textwidth]{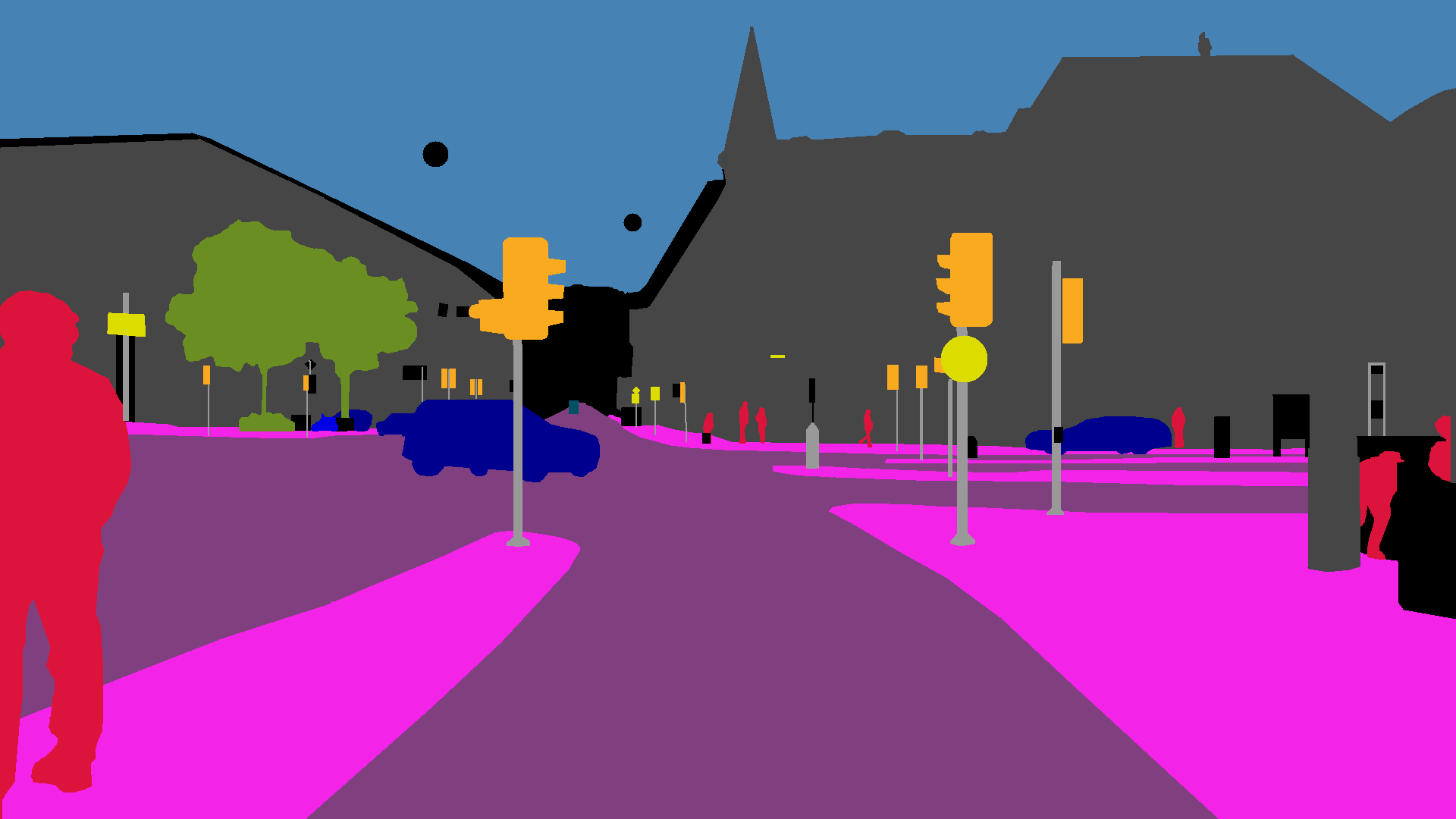}}
    \hfil
    \subfloat[AdaptSegNet~\cite{adapt:structured:output:cvpr18}]{\includegraphics[width=0.195\textwidth]{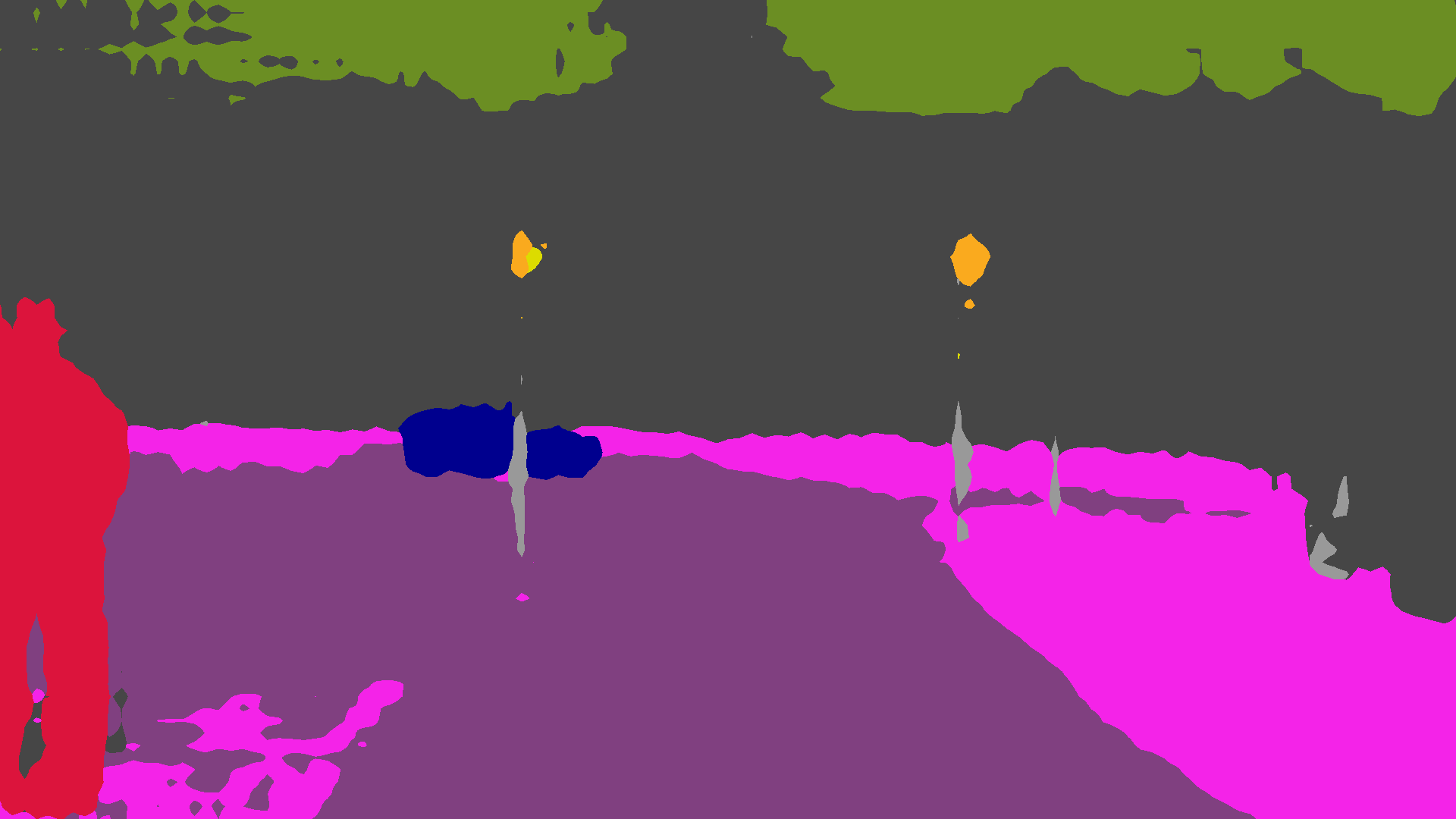}}
    \hfil
    \subfloat[DMAda~\cite{daytime:2:nighttime}]{\includegraphics[width=0.195\textwidth]{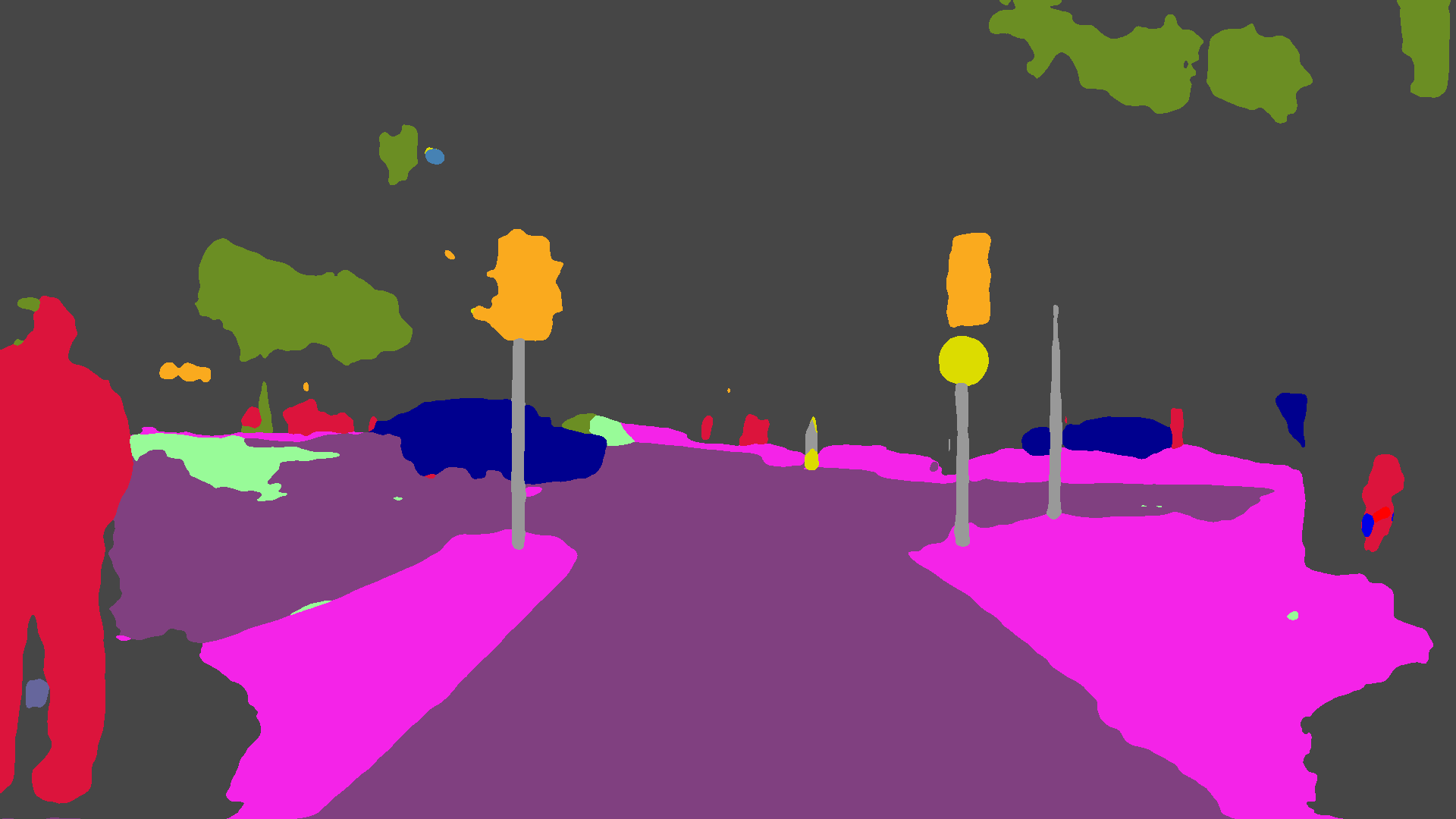}}
    \hfil
    \subfloat[MGCDA (Ours)]{\includegraphics[width=0.195\textwidth]{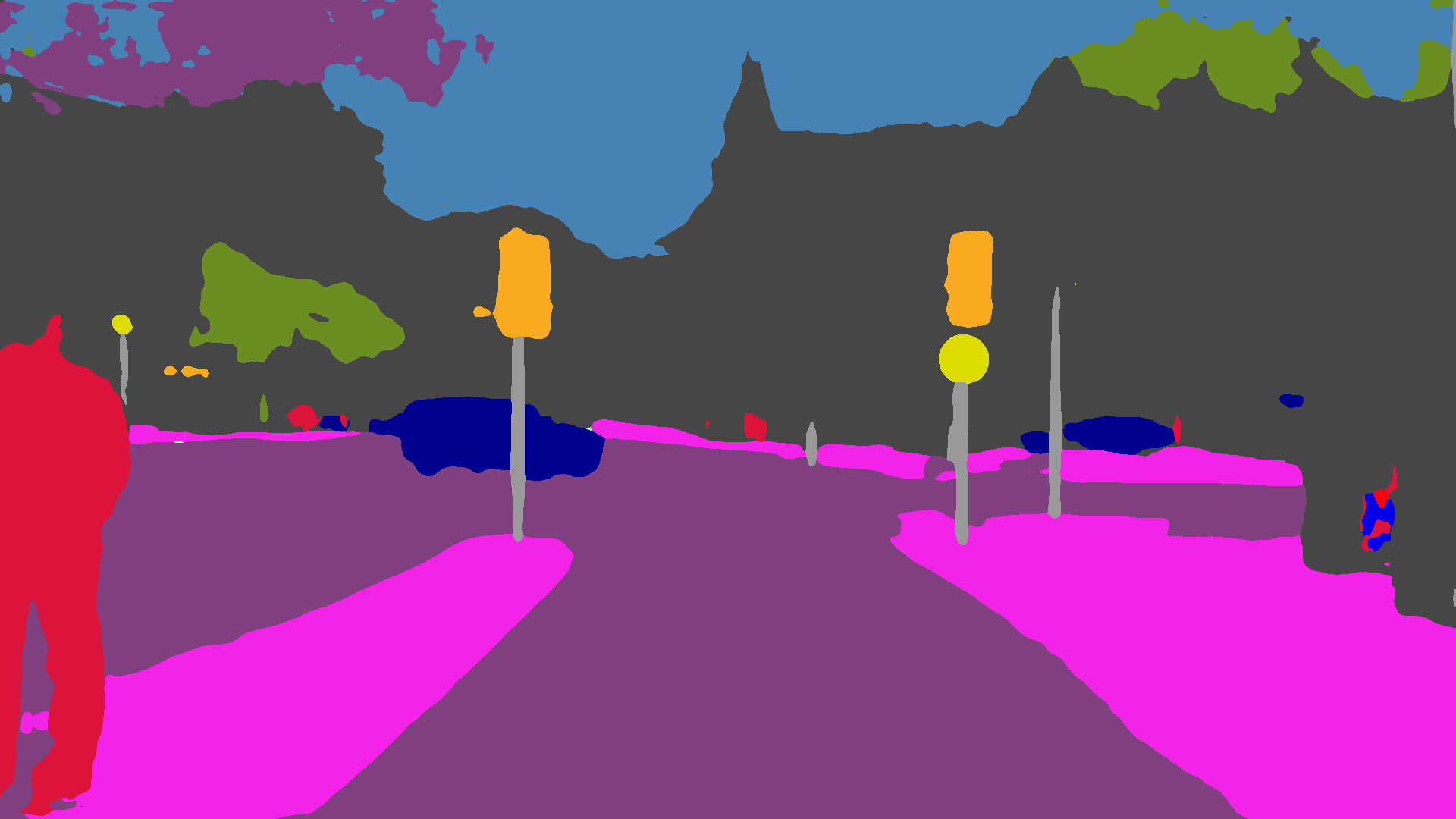}}
    \caption{Qualitative semantic segmentation results on \emph{Dark Zurich-test}. ``AdaptSegNet'' adapts from Cityscapes to \emph{Dark Zurich-night}.}
    \label{fig:sem:seg:dark_zurich}
\end{figure*}

\begin{table}[!tb]
    \caption{Comparison of \emph{Dark Zurich} against related datasets with nighttime semantic annotations. ``Night annot.'': annotated nighttime images, ``Invalid'': can invalid regions get legitimate labels?}
    \label{table:datasets:comparison}
    \centering
    \footnotesize
    \setlength\tabcolsep{1.2pt}
    \begin{tabular}{lccccc}
    \toprule
    Dataset & Night annot. & Classes & Reliable GT & Fine GT & Invalid \\
    \midrule
    WildDash~\cite{wilddash} & 13 & 19 & \yes & \yes & \no \\
    Raincouver~\cite{raincouver} & 95 & 3 & \yes & \no & \no \\
    BDD100K~\cite{BDD100K} & 345 & 19 & \no & \yes & \no \\
    Nighttime Driving~\cite{daytime:2:nighttime} & 50 & 19 & \yes & \no & \no \\
    Dark Zurich & 201 & 19 & \yes & \yes & \yes \\
    \bottomrule
    \end{tabular}
\end{table}

\emph{Dark Zurich} was recorded in Zurich using a 1080p GoPro Hero 5 camera, mounted on top of the front windshield of a car. The collection protocol with multiple drives of several laps to establish correspondences is detailed in Sec.~\ref{sec:mgcda}.

We split \emph{Dark Zurich} and reserve one lap for validation and another lap for testing. The rest of the laps remain unlabeled and are used for training. They comprise 3041 daytime, 2920 twilight and 2416 nighttime images extracted at 1 fps, which are named \emph{Dark Zurich}-\{\emph{day}, \emph{twilight}, \emph{night}\} respectively and correspond to the three sets in the rightmost column of Table~\ref{tab:GCMA:notations}. From the validation and testing night laps, we extract one image every 50m or 20s, whichever comes first, and assign to it the corresponding daytime image to serve as the auxiliary image $I^{\prime}$ in our annotation (cf.\ Sec.~\ref{sec:evaluation:annotation}). We annotate 201 nighttime images (151 from the testing lap and 50 from the validation lap) with fine pixel-level Cityscapes labels and invalid masks following our protocol and name these sets \emph{Dark Zurich-test} and \emph{Dark Zurich-val} respectively. In total, 366.8M pixels have been annotated with semantic labels and 90.2M of these pixels are marked as invalid. Detailed annotation statistics are provided in Fig.~\ref{fig:dataset:stats}. We validate the quality of our annotations by having 20 images annotated twice by different subjects and measuring consistency. 93.5\% of the labeled pixels are consistent in the semantic annotations and respectively 95\% in the invalid masks.
We compare to existing annotated nighttime sets in Table~\ref{table:datasets:comparison}, noting that most large-scale sets for road scene parsing, such as Cityscapes~\cite{Cityscapes} and Mapillary Vistas~\cite{Mapillary}, contain few or no nighttime scenes. Nighttime Driving~\cite{daytime:2:nighttime} and Raincouver~\cite{raincouver} only include \emph{coarse} annotations. \emph{Dark Zurich} contains fifteen times more annotated nighttime images than WildDash~\cite{wilddash}---the only other dataset with fine and \emph{reliable} nighttime annotations. Detailed inspection showed that $\sim$70\% of the 345 densely annotated nighttime images of BDD100K~\cite{BDD100K} contain severe labeling errors which render them unsuitable for evaluation, especially in dark regions we treat as invalid (e.g.\ \emph{sky} is often mislabeled as \emph{building}). Our annotation protocol helps avoid such errors by properly defining invalid regions and using daytime images to aid annotation, and the labeled part of \emph{Dark Zurich} is an initial high-quality benchmark to promote our uncertainty-aware evaluation.

\section{Results}
\label{sec:exp}

\begin{table*}[!tb]
  \caption{Performance comparison of our method with state-of-the-art approaches and daytime-trained baselines on our \emph{Dark Zurich-test} dataset. Cityscapes$\to$\emph{DZ-night} denotes adaptation from Cityscapes to \emph{Dark Zurich-night}.}
  \label{table:exp:main:dark_zurich}
  \centering
  \setlength\tabcolsep{1.9pt}
  \footnotesize
  \begin{tabular}{lcccccccccccccccccccc}
  \toprule
  Method & \ver{road} & \ver{sidew.} & \ver{build.} & \ver{wall} & \ver{fence} & \ver{pole} & \ver{light} & \ver{sign} & \ver{veget.} & \ver{terrain} & \ver{sky} & \ver{person} & \ver{rider} & \ver{car} & \ver{truck} & \ver{bus} & \ver{train} & \ver{motorc.} & \ver{bicycle} & mIoU\\
  \midrule
  RefineNet~\cite{refinenet} & 68.8&23.2&46.8&20.8&12.6&29.8&30.4&26.9&43.1&14.3&0.3&36.9&49.7&63.6&6.8&\best{0.2}&24.0&33.6&9.3     & 28.5\\
  DeepLab-v2~\cite{DeepLab:v2} & 79.0 & 21.8 & 53.0 & 13.3 & 11.2 & 22.5 & 20.2 & 22.1 & 43.5 & 10.4 & 18.0 & 37.4 & 33.8 & 64.1 & 6.4 & 0.0 & 52.3 & 30.4 & 7.4     & 28.8\\
  \midrule
  AdaptSegNet-Cityscapes$\to$\emph{DZ-night}~\cite{adapt:structured:output:cvpr18} & \best{86.1} & 44.2 & 55.1 & 22.2 & 4.8 & 21.1 & 5.6 & 16.7 & 37.2 & 8.4 & 1.2 & 35.9 & 26.7 & 68.2 & 45.1 & 0.0 & 50.1 & 33.9 & 15.6       & 30.4\\
  ADVENT-Cityscapes$\to$\emph{DZ-night}~\cite{advent:adaptation} & 85.8 & 37.9 & 55.5 & 27.7 & 14.5 & 23.1 & 14.0 & 21.1 & 32.1 & 8.7 & 2.0 & 39.9 & 16.6 & 64.0 & 13.8 & 0.0 & \best{58.8} & 28.5 & 20.7    & 29.7\\
  BDL-Cityscapes$\to$\emph{DZ-night}~\cite{bidirectional:learning:adaptation} & 85.3 & 41.1 & 61.9 & \best{32.7} & 17.4 & 20.6 & 11.4 & 21.3 & 29.4 & 8.9 & 1.1 & 37.4 & 22.1 & 63.2 & 28.2 & 0.0 & 47.7 & \best{39.4} & 15.7    & 30.8\\
  DMAda~\cite{daytime:2:nighttime} & 75.5 & 29.1 & 48.6 & 21.3 & 14.3 & 34.3 & 36.8 & 29.9 & 49.4 & 13.8 & 0.4 & 43.3 & 50.2 & 69.4 & 18.4 & 0.0 & 27.6 & 34.9 & 11.9     & 32.1\\
  Ours: GCMA~\cite{GCMA_UIoU:v1} & 81.7 & 46.9 & 58.8 & 22.0 & \best{20.0} & 41.2 & \best{40.5} & \best{41.6} & \best{64.8} & \best{31.0} & 32.1 & \best{53.5} & 47.5 & \best{75.5} & 39.2 & 0.0 & 49.6 & 30.7 & 21.0             & 42.0\\
  Ours: MGCDA & 80.3 & \best{49.3} & \best{66.2} & 7.8 & 11.0 & \best{41.4} & 38.9 & 39.0 & 64.1 & 18.0 & \best{55.8} & 52.1 & \best{53.5} & 74.7 & \best{66.0} & 0.0 & 37.5 & 29.1 & \best{22.7}     & \best{42.5}\\
  \bottomrule
  \\
  \end{tabular}
\end{table*}

Our architecture of choice for implementing MGCDA is RefineNet~\cite{refinenet}. We use the publicly available \emph{RefineNet-res101-Cityscapes} model, trained on Cityscapes, as the baseline model to be adapted to nighttime. Throughout our experiments, we train this model with a constant learning rate of $5 \times 10^{-5}$ on mini-batches of size 1. To obtain the synthetic labeled datasets for MGCDA, we stylize Cityscapes to twilight using a CycleGAN model that is trained to translate Cityscapes to \emph{Dark Zurich-twilight} (respectively to nighttime with \emph{Dark Zurich-night}). The real training datasets for MGCDA are \emph{Dark Zurich-day}, instantiating $\mathcal{D}^1_{ur}$, and \emph{Dark Zurich-twilight}, instantiating $\mathcal{D}^2_{ur}$. Each adaptation step comprises 30k SGD iterations and uses $\mu = 1$. For the second step, we apply our guided refinement to the labels of \emph{Dark Zurich-twilight} that are predicted by model $\phi^2$ fine-tuned in the first step, using the correspondences of \emph{Dark Zurich-twilight} to \emph{Dark Zurich-day}. In particular, we experiment with both variants of our guided refinement, i.e., the original variant which was presented in our conference paper~\cite{GCMA_UIoU:v1} and uses cross bilateral filtering (Sec.~\ref{sec:mgcda:guidance:bilateral}), and the new upgraded variant which uses depth-based warping (Sec.~\ref{sec:mgcda:guidance:warping}). We refer to the original variant of our complete pipeline as GCMA and to the upgraded variant as MGCDA. For MGCDA, we note that for the corresponding image pairs of \emph{Dark Zurich-twilight} and \emph{Dark Zurich-day} on which the RANSAC step of the depth-based warping variant for guided refinement detects less than 14 inliers, we fall back to cross bilateral filtering. This pertains to 1852/2920 pairs. Moreover, MGCDA uses an improved configuration for CycleGAN-based stylization compared to GCMA, which is detailed in Sec.~\ref{sec:exp:cyclegan}.

\subsection{Comparison to Other Adaptation Methods}
\label{sec:exp:other:methods}

\begin{table}[!tb]
  \caption{Performance comparison of our method with state-of-the-art approaches and daytime-trained baselines on Nighttime Driving~\cite{daytime:2:nighttime}. Read as Table~\ref{table:exp:main:dark_zurich}.}
  \label{table:exp:main:nighttime_driving}
  \centering
  \setlength\tabcolsep{4pt}
  \footnotesize
  \begin{tabular}{lc}
  \toprule
  Method & mIoU (\%)\\
  \midrule
  RefineNet~\cite{refinenet} & 31.5\\
  DeepLab-v2~\cite{DeepLab:v2} & 32.6\\
  \midrule
  AdaptSegNet-Cityscapes$\to$\emph{DZ-night}~\cite{adapt:structured:output:cvpr18} & 34.5\\
  ADVENT-Cityscapes$\to$\emph{DZ-night}~\cite{advent:adaptation} & 34.7\\
  BDL-Cityscapes$\to$\emph{DZ-night}~\cite{bidirectional:learning:adaptation} & 34.7\\
  DMAda~\cite{daytime:2:nighttime} & 36.1\\
  Ours: GCMA~\cite{GCMA_UIoU:v1} & 45.6\\
  Ours: MGCDA & \best{49.4}\\
  \bottomrule
  \\
  \end{tabular}
\end{table}

\begin{table}[!tb]
  \caption{Performance comparison of our method with state-of-the-art approaches and daytime-trained baselines on BDD100K-night~\cite{BDD100K}. Read as Table~\ref{table:exp:main:dark_zurich}.}
  \label{table:exp:main:bdd}
  \centering
  \setlength\tabcolsep{4pt}
  \footnotesize
  \begin{tabular}{lc}
  \toprule
  Method & mIoU (\%)\\
  \midrule
  RefineNet~\cite{refinenet} & 26.6\\
  DeepLab-v2~\cite{DeepLab:v2} & 22.9\\
  \midrule
  AdaptSegNet-Cityscapes$\to$\emph{DZ-night}~\cite{adapt:structured:output:cvpr18} & 22.0\\
  ADVENT-Cityscapes$\to$\emph{DZ-night}~\cite{advent:adaptation} & 22.6\\
  BDL-Cityscapes$\to$\emph{DZ-night}~\cite{bidirectional:learning:adaptation} & 22.8\\
  DMAda~\cite{daytime:2:nighttime} & 28.3\\
  Ours: GCMA~\cite{GCMA_UIoU:v1} & 33.2\\
  Ours: MGCDA & \best{34.9}\\
  \bottomrule
  \\
  \end{tabular}
\end{table}

Our first experiment compares MGCDA and GCMA to state-of-the-art approaches for adaptation of semantic segmentation models to nighttime. We evaluate MGCDA and GCMA on \emph{Dark Zurich-test} against the state-of-the-art adaptation approaches AdaptSegNet~\cite{adapt:structured:output:cvpr18}, BDL~\cite{bidirectional:learning:adaptation}, ADVENT~\cite{advent:adaptation} and DMAda~\cite{daytime:2:nighttime} and report standard IoU performance in Table~\ref{table:exp:main:dark_zurich}, including \emph{invalid} pixels which are assigned a legitimate semantic label in the evaluation. We have trained AdaptSegNet, BDL and ADVENT to adapt from Cityscapes to \emph{Dark Zurich-night}. For fair comparison, we also report the performance of the respective baseline Cityscapes models for each method. RefineNet is the common baseline of MGCDA, GCMA and DMAda, while DeepLab-v2~\cite{DeepLab:v2} is the common baseline of AdaptSegNet, BDL and ADVENT. The fact that both baseline models feature a ResNet-101 backbone~\cite{resnet} allows a direct comparison.

Both MGCDA and GCMA significantly outperform the other methods for most classes and achieve a substantial 10\% improvement in the overall mIoU score against the next best method. The improvement with MGCDA and GCMA is pronounced for classes which usually appear dark at nighttime, such as \emph{sky}, \emph{vegetation}, \emph{building} and \emph{person}, indicating that our method successfully handles large domain shifts from its source daytime domain. These findings are supported by visually assessing the predictions of the compared methods, as in the examples of Fig.~\ref{fig:sem:seg:dark_zurich}.

\begin{figure*}[!tb]
    \centering
    \subfloat{\includegraphics[width=0.195\textwidth]{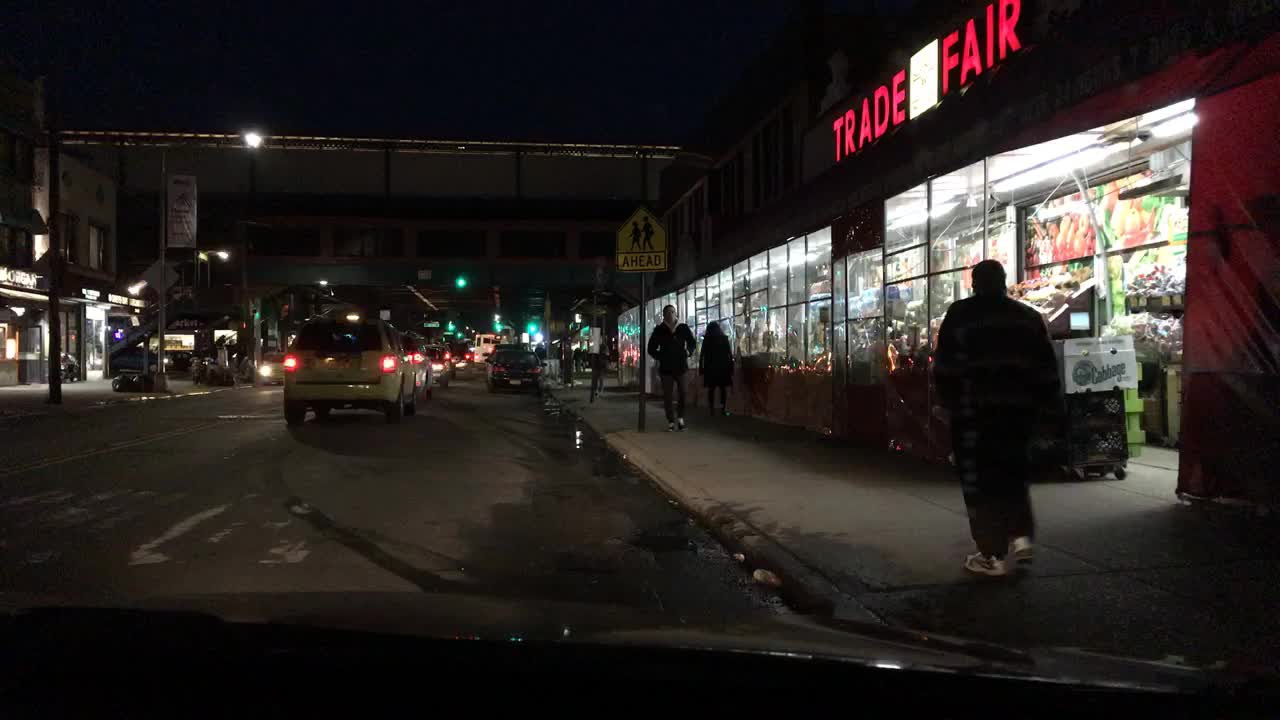}}
    \hfil
    \subfloat{\includegraphics[width=0.195\textwidth]{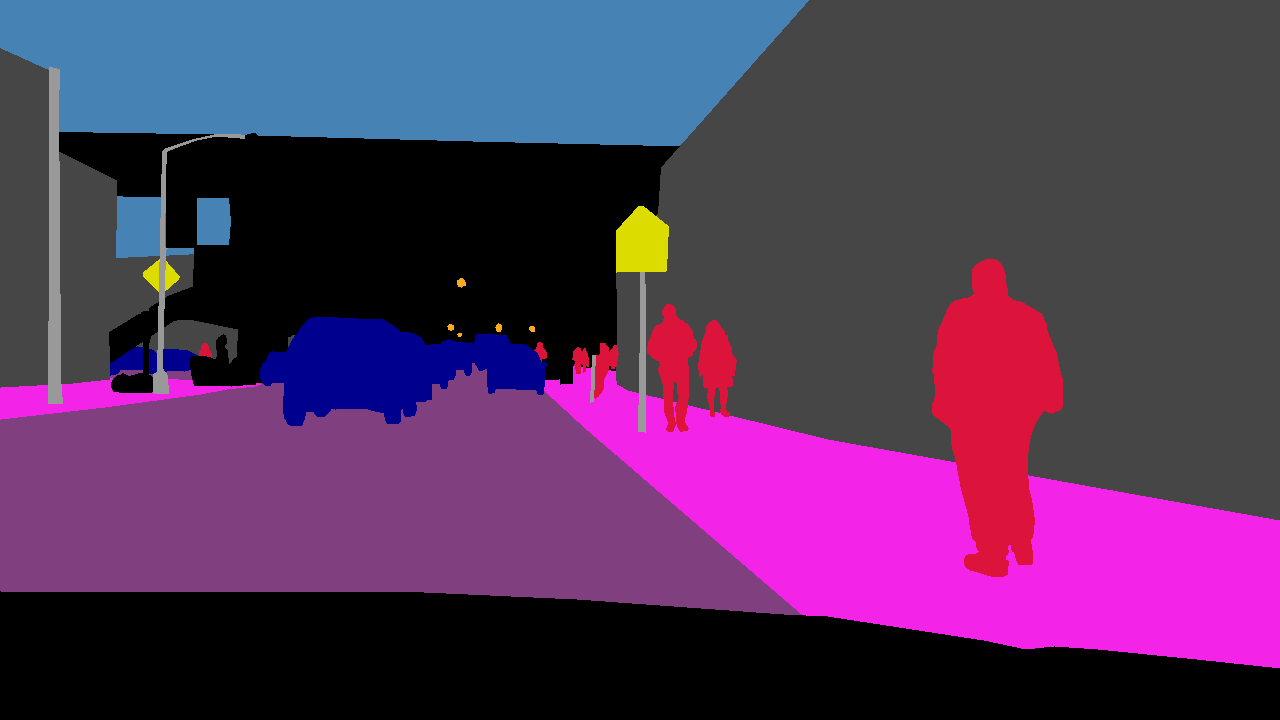}}
    \hfil
    \subfloat{\includegraphics[width=0.195\textwidth]{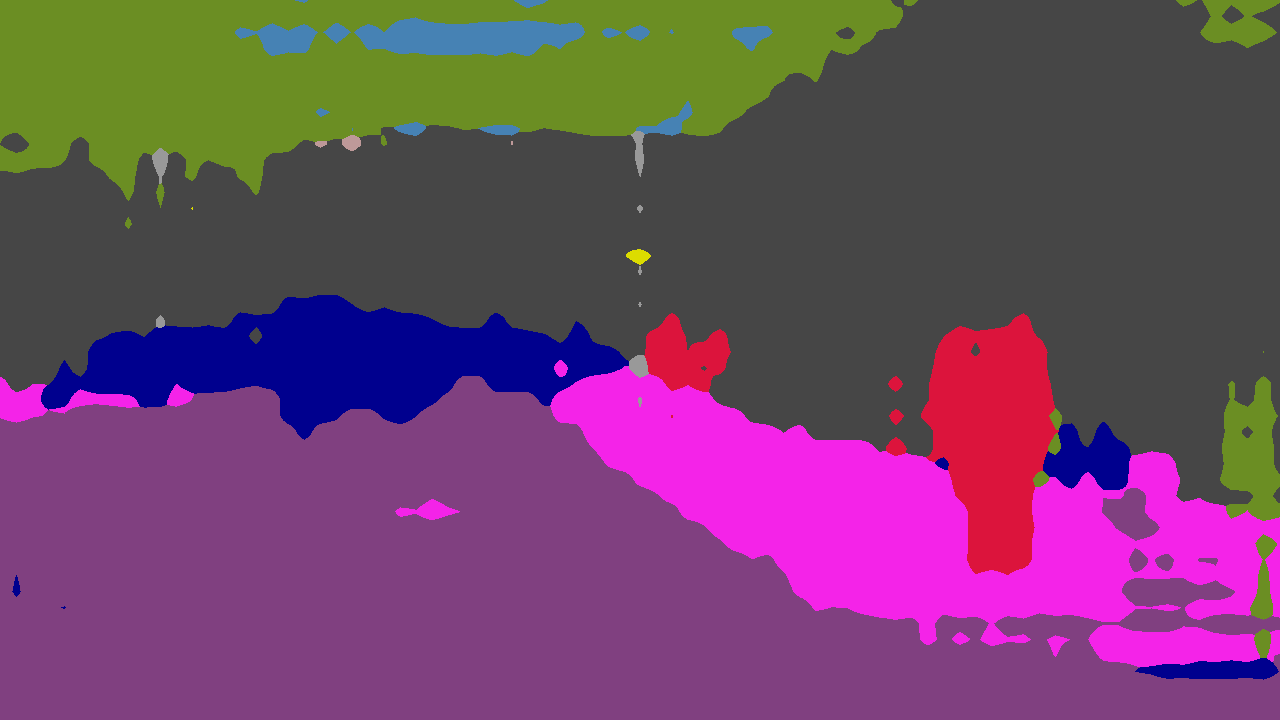}}
    \hfil
    \subfloat{\includegraphics[width=0.195\textwidth]{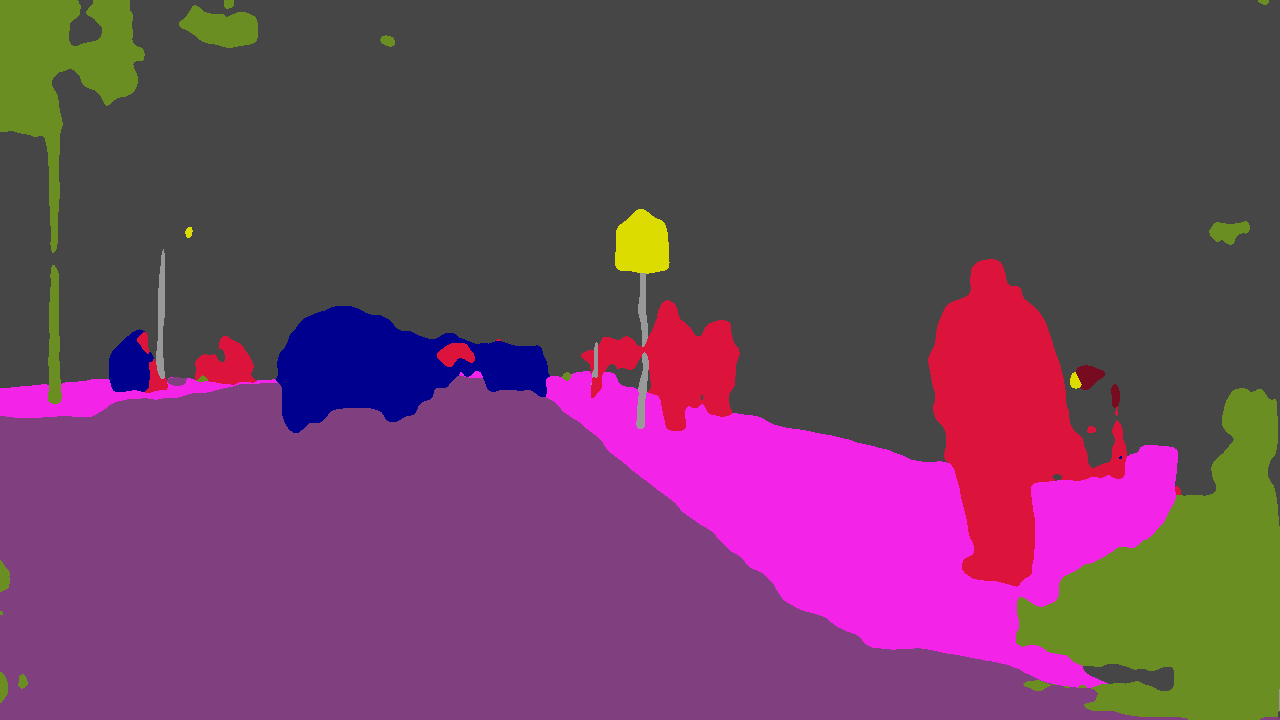}}
    \hfil
    \subfloat{\includegraphics[width=0.195\textwidth]{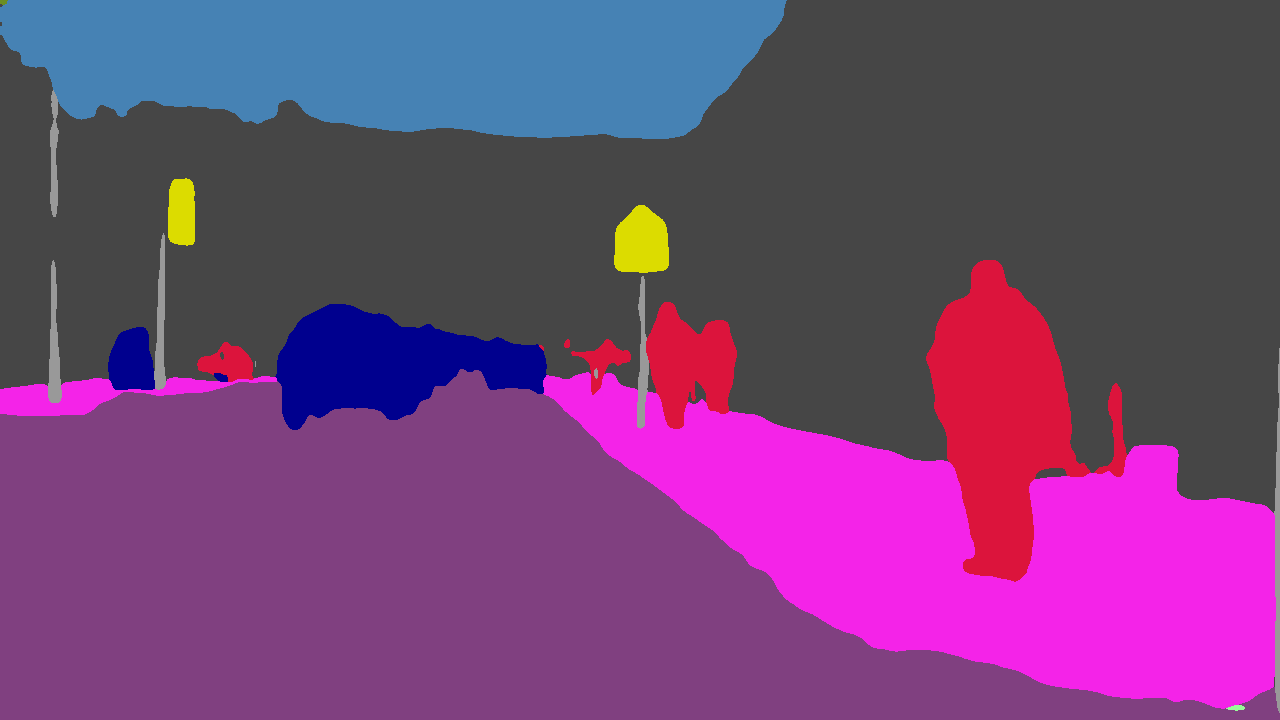}}
    \\
    \vspace{-0.3cm}
    \addtocounter{subfigure}{-5}
    \subfloat[Image]{\includegraphics[width=0.195\textwidth]{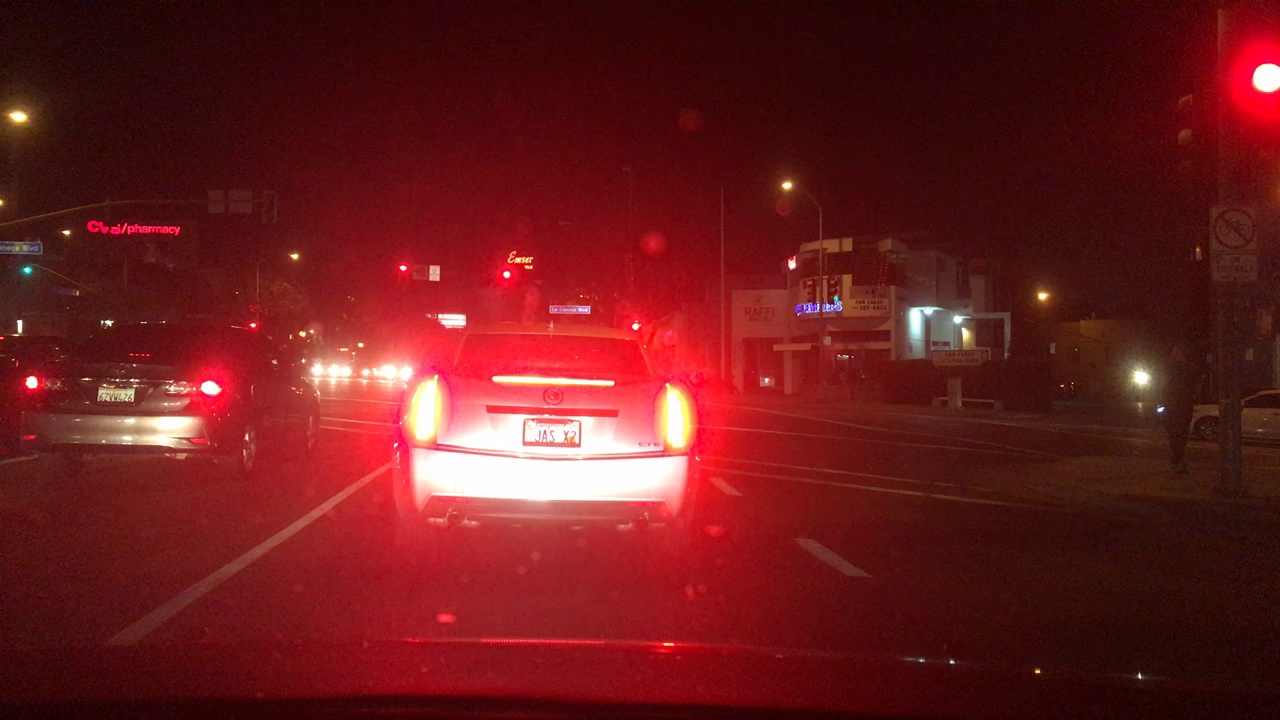}}
    \hfil
    \subfloat[Semantic GT]{\includegraphics[width=0.195\textwidth]{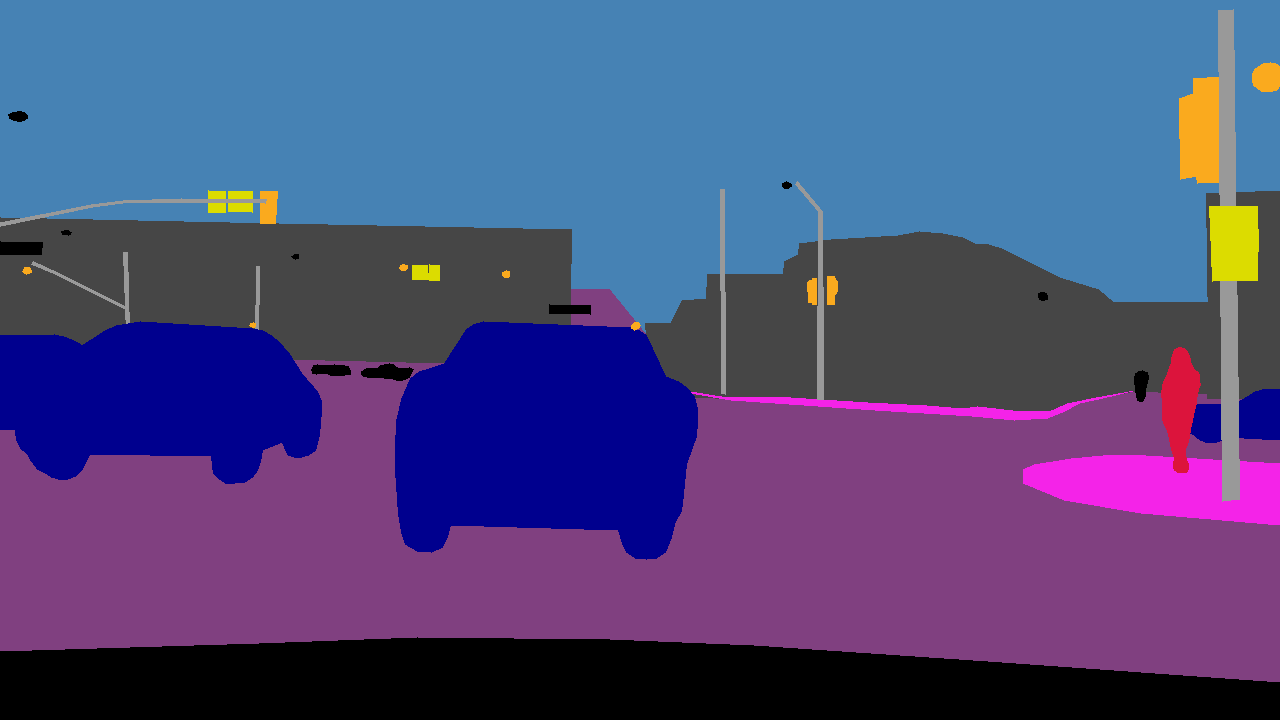}}
    \hfil
    \subfloat[BDL~\cite{bidirectional:learning:adaptation}]{\includegraphics[width=0.195\textwidth]{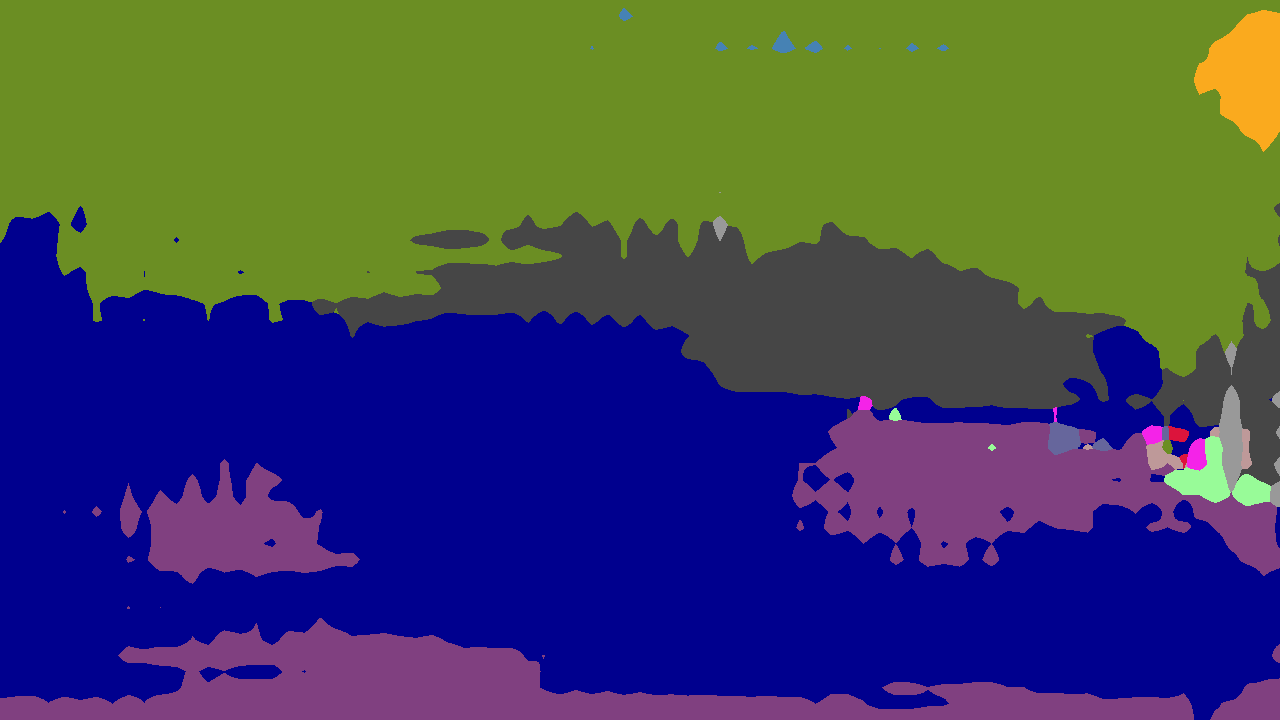}}
    \hfil
    \subfloat[DMAda~\cite{daytime:2:nighttime}]{\includegraphics[width=0.195\textwidth]{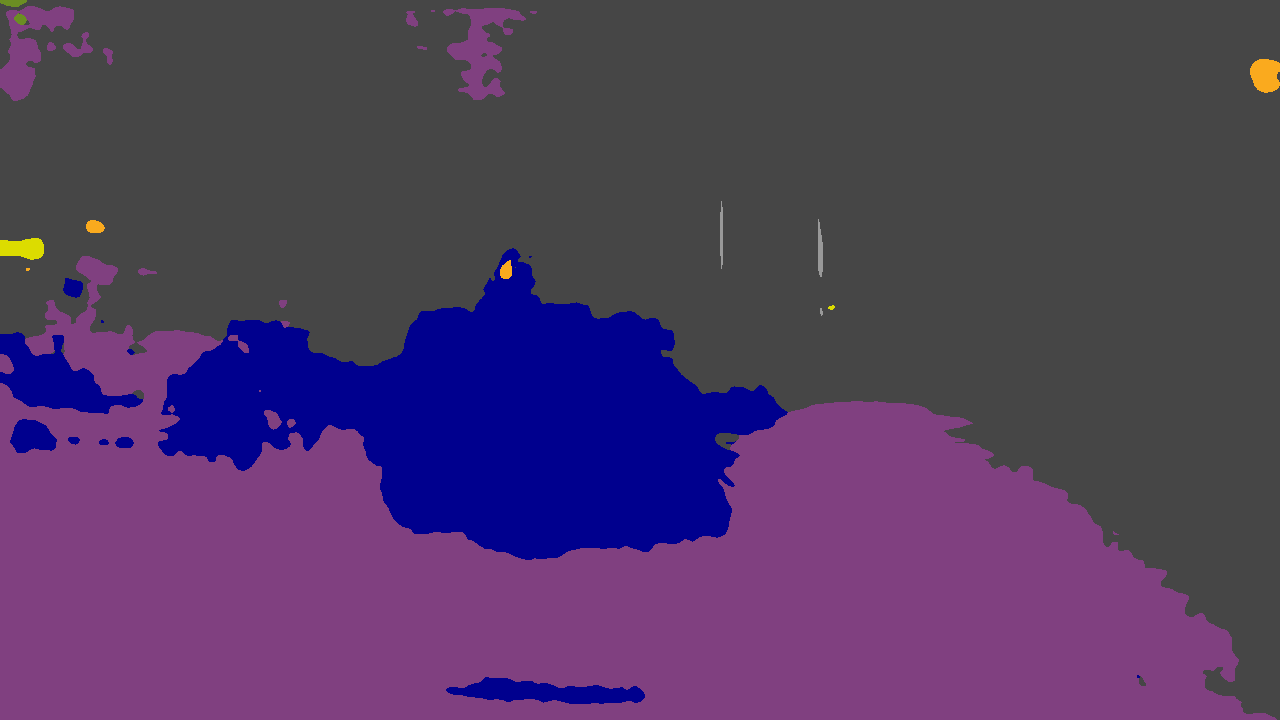}}
    \hfil
    \subfloat[MGCDA (Ours)]{\includegraphics[width=0.195\textwidth]{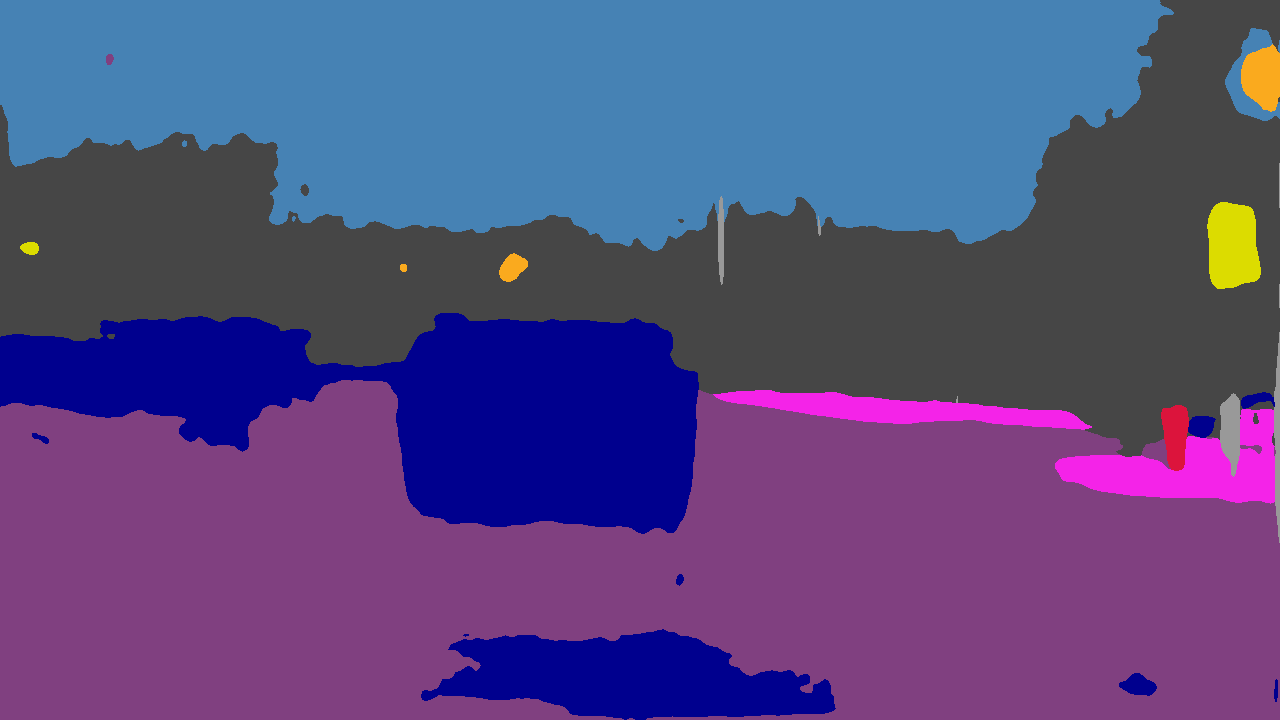}}
    \caption{Qualitative semantic segmentation results on BDD100K-night. ``BDL'' adapts from Cityscapes to \emph{Dark Zurich-night}.}
    \label{fig:sem:seg:bdd}
\end{figure*}

In order to reinforce these conclusions and show that our method generalizes very well to different datasets, we repeat the above comparison on two additional sets. More specifically, we evaluate the various approaches on Nighttime Driving~\cite{daytime:2:nighttime} and report the results in Table~\ref{table:exp:main:nighttime_driving}. In addition, we consider BDD100K~\cite{BDD100K} as a candidate benchmark, even though it presents the difficulty of unreliable ground-truth annotations, as mentioned in Sec.~\ref{sec:dark:zurich}. We overcome this issue by manually identifying a list of 87 images (out of a total of 345) whose annotations are free from obvious errors. We name this subset BDD100K-night and restrict ourselves to it for evaluation. The associated results are reported in Table~\ref{table:exp:main:bdd}. Indeed, MGCDA and GCMA are by far the best-performing adaptation methods on Nighttime Driving and BDD100K-night. The rest of the methods generally deliver only slight improvements compared to their respective daytime baselines. MGCDA improves upon RefineNet by very large margins: 17.9\% on Nighttime Driving and 8.3\% on BDD100K-night. This large improvement on BDD100K-night is achieved even though MGCDA has not been presented with any image from the particular domain of BDD100K during training. Equally importantly, MGCDA brings a significant benefit of 3.8\% on Nighttime Driving and 1.7\% on BDD100K-night compared to GCMA, which supports the utility of the novel geometrically guided refinement via depth-based warping for adaptation thanks to more accurate resulting pseudo-labels. The superiority of MGCDA is demonstrated in the qualitative results of Fig.~\ref{fig:sem:seg:bdd} on BDD100K-night.

\subsection{Image Translation with CycleGAN}
\label{sec:exp:cyclegan}
\begin{figure}
    \centering
    \subfloat{\includegraphics[width=0.33\linewidth]{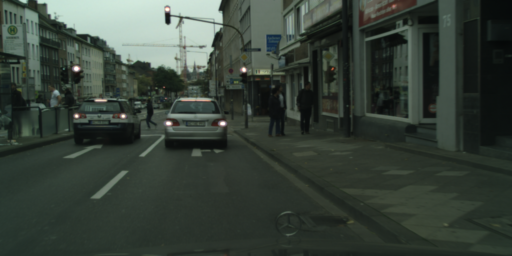}}
    \hfil
    \subfloat{\includegraphics[width=0.33\linewidth]{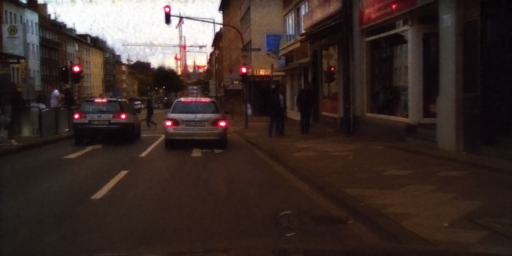}}
    \hfil
    \subfloat{\includegraphics[width=0.33\linewidth]{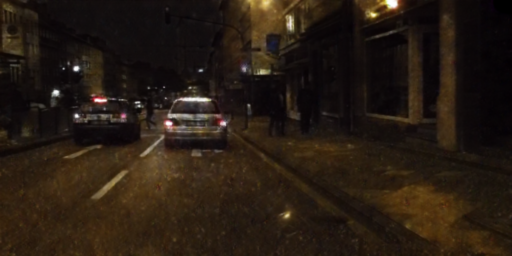}}
    \\
    \vspace{-0.3cm}
    \subfloat{\includegraphics[width=0.33\linewidth]{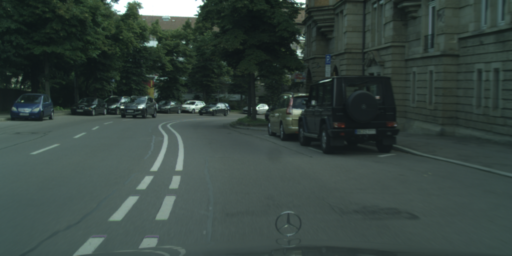}}
    \hfil
    \subfloat{\includegraphics[width=0.33\linewidth]{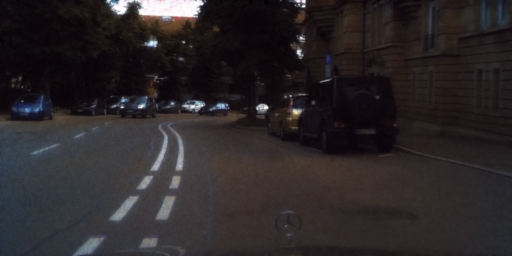}}
    \hfil
    \subfloat{\includegraphics[width=0.33\linewidth]{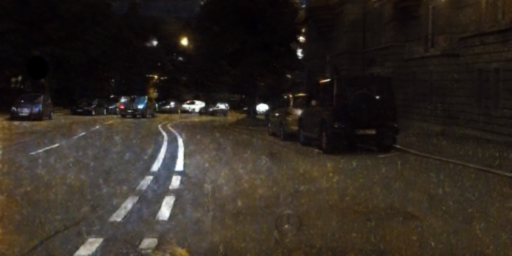}}
    \caption{Comparison of CycleGAN configurations for generation of synthetic stylized data from Cityscapes and \emph{Dark Zurich}. From left to right: input Cityscapes image, stylized nighttime image with training on $256\times{}256$ crops, stylized nighttime image with training on full $360\times{}720$ images.}
    \label{fig:exp:cyclegan}
\end{figure}

Compared to GCMA, in MGCDA we have implemented a different configuration for training and testing CycleGAN to stylize images as twilight or nighttime for generating our synthetic training sets. More specifically, the default CycleGAN configuration, which we used in GCMA, involves training the entire architecture on small $256\times{}256$ crops of the input images, while at test time the full images are passed to the generator. We have observed that this discrepancy between the field of view of the CycleGAN model at training versus test time leads to a smaller degree of translation of the overall image appearance than desired, as shown in Fig.~\ref{fig:exp:cyclegan}. To resolve this issue in the synthetic data stream of our pipeline, we downsize input images from both domains to $360\times{}720$ resolution, so that the entire architecture fits into GPU memory, and train CycleGAN on the \emph{full} downsized images. At test time, the same downsized images as in training are input to the generator, and the stylized $360\times{}720$ outputs are upsampled to the original resolution using joint bilateral upsampling~\cite{joint:bilateral:upsampling}. In this way, the generator is presented with the entire pattern of appearance changes across different image regions and it is able to learn better the global shift in illumination between the two domains, as can be seen in Fig.~\ref{fig:exp:cyclegan}. Apart from this visual comparison, we also demonstrate in Table~\ref{table:exp:cyclegan:dark_zurich} the induced improvement in image translation from full-image training in the target context of semantic segmentation adaptation, using Cityscapes images stylized as nighttime with the two examined CycleGAN variants to adapt the baseline RefineNet model to nighttime in a single step. We therefore use full-image CycleGAN training for generating the synthetic images in our upgraded MGCDA pipeline.

\begin{table}[!tb]
  \caption{Comparison on \emph{Dark Zurich-test} of CycleGAN configurations for generation of synthetic stylized data to adapt to nighttime. CycleGAN-crops stands for the default training of CycleGAN with $256\times{}256$ crops, whereas CycleGAN-full stands for training of CycleGAN with full $360\times{}720$ images.}
  \label{table:exp:cyclegan:dark_zurich}
  \centering
  \setlength\tabcolsep{4pt}
  \footnotesize
  \begin{tabular}{lc}
  \toprule
  Method & mIoU (\%)\\
  \midrule
  Daytime baseline: RefineNet~\cite{refinenet} & 28.5\\
  \midrule
  Previous state of the art: DMAda~\cite{daytime:2:nighttime} & 32.1\\
  \midrule
  CycleGAN-crops adaptation & 37.1\\
  CycleGAN-full adaptation & \best{40.2}\\
  \bottomrule
  \\
  \end{tabular}
\end{table}

\subsection{Ablation Study for MGCDA}
\label{sec:exp:ablation:gcma}

\begin{table}[!tb]
  \caption{Ablation study of the components of MGCDA on \emph{Dark Zurich-test}, reporting mIoU (\%). CBF stands for cross bilateral filtering and DBW for depth-based warping. W/o twilight means not using data from the intermediate domain. W/o curriculum means single-stage adaptation using all available data. The third and fourth rows are with reference to the respective previous row. The fifth and sixth rows are both with reference to the fourth row.}
  \label{table:exp:ablation:dark_zurich}
  \centering
  \setlength\tabcolsep{3pt}
  \footnotesize
  \begin{tabular}{lc}
  \toprule
  Daytime-trained baseline: RefineNet~\cite{refinenet} & 28.5\\
  \midrule
  Adaptation to nighttime (w/o twilight/curriculum/guided refinement) & 33.6\\
  +twilight (w/o curriculum/guided refinement) & 34.0\\
  +curriculum (MGCDA w/o guided refinement) & 38.2\\
  \midrule
  +guided refinement-CBF & 40.3\\
  +guided refinement-DBW (MGCDA) & \best{42.5}\\
  \bottomrule
  \\
  \end{tabular}
\end{table}

\begin{table}[!tb]
  \caption{Comparison on \emph{Dark Zurich-test} focusing on the usage of map guidance at test time and reporting mIoU (\%).}
  \label{table:exp:test:guidance:dark_zurich}
  \centering
  \setlength\tabcolsep{4pt}
  \footnotesize
  \begin{tabular}{lcc}
  \toprule
  Method & w/o test guidance & w/ test guidance\\
  \midrule
  DMAda~\cite{daytime:2:nighttime} & 32.1 & \best{34.8}\\
  Ours: MGCDA & 42.5 & \best{44.1}\\
  \bottomrule
  \\
  \end{tabular}
\end{table}

We measure the individual effect of the main components of MGCDA in Table~\ref{table:exp:ablation:dark_zurich} by evaluating its ablated versions on \emph{Dark Zurich-test}. The naive baseline adapting directly to nighttime delivers only a slight benefit compared to the daytime-trained model. By comparison, adaptation to nighttime with our two-stage curriculum training involving twilight data delivers a much larger benefit of 9.7\%, thanks to the fact that pseudo-labels of real images are always inferred by models already adapted to the respective domain. On the contrary, the corresponding adaptation variant that is trained without a curriculum, i.e.\ in one stage with 60k SGD iterations (equal to the total iterations for the two stages of MGCDA) and data of varying darkness presented in mixed order, cannot profit from data of intermediate darkness. Applying our guided segmentation refinement in its original, cross bilateral filtering variant that we have used in GCMA significantly improves upon the basic curriculum approach. Finally, the upgraded, depth-based warping variant of our guided refinement that we use in MGCDA brings an additional 2.2\% benefit over the original variant, as it corrects even more errors in the pseudo-labels of the real images, which helps compute more reliable gradients from the corrected loss during the subsequent training. Note that only with the fully fledged MGCDA with our novel geometrically guided refinement are we able to adequately leverage real data and significantly improve upon the powerful synthetic-only adaptation baseline with CycleGAN from Table~\ref{table:exp:cyclegan:dark_zurich}.

\subsection{Map Guidance at Test Time}
\label{sec:exp:test:guidance}
\begin{figure}
    \centering
    \subfloat{\includegraphics[width=0.245\linewidth]{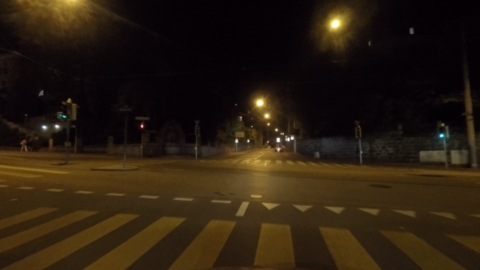}}
    \hfil
    \subfloat{\includegraphics[width=0.245\linewidth]{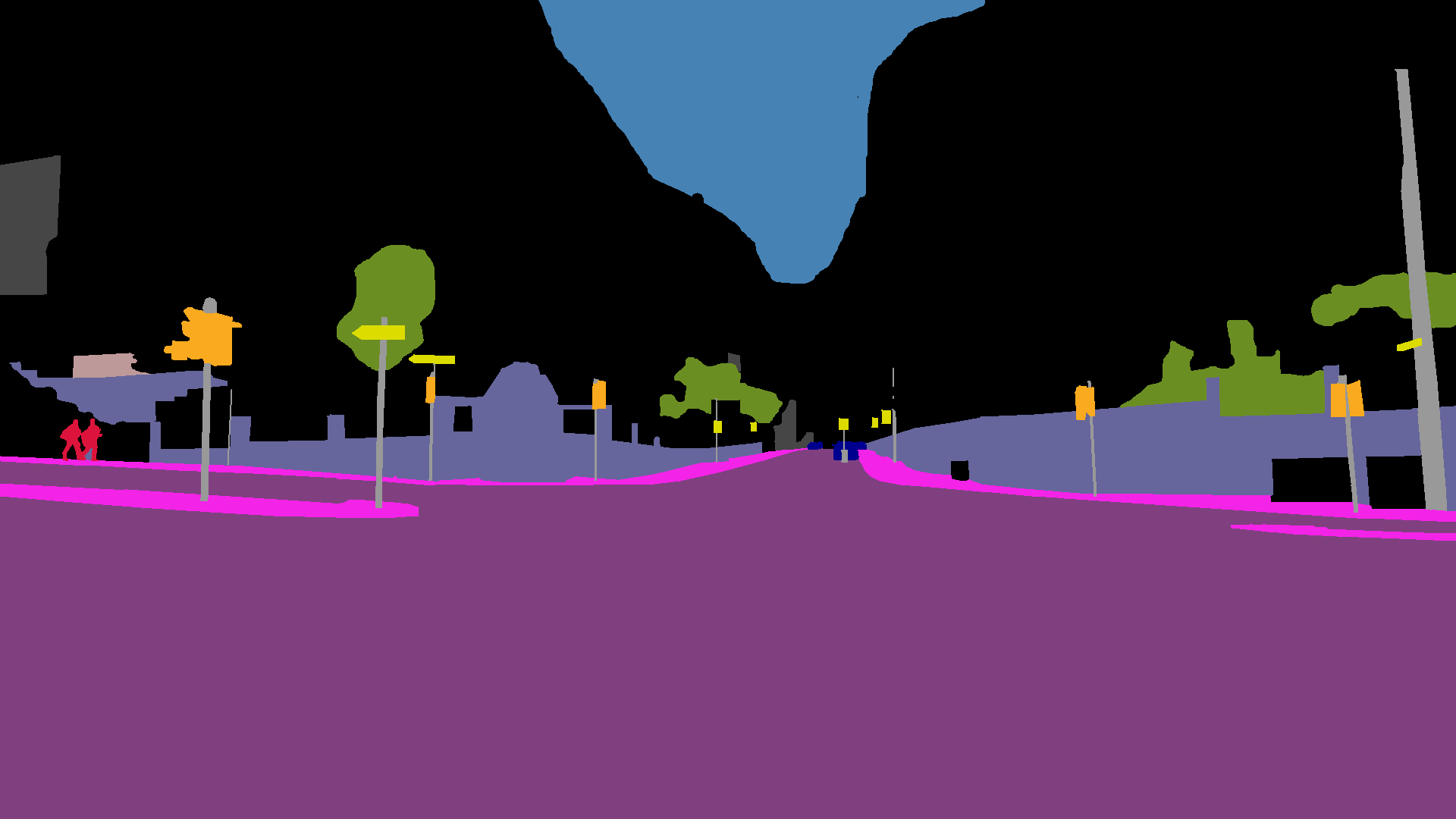}}
    \hfil
    \subfloat{\includegraphics[width=0.245\linewidth]{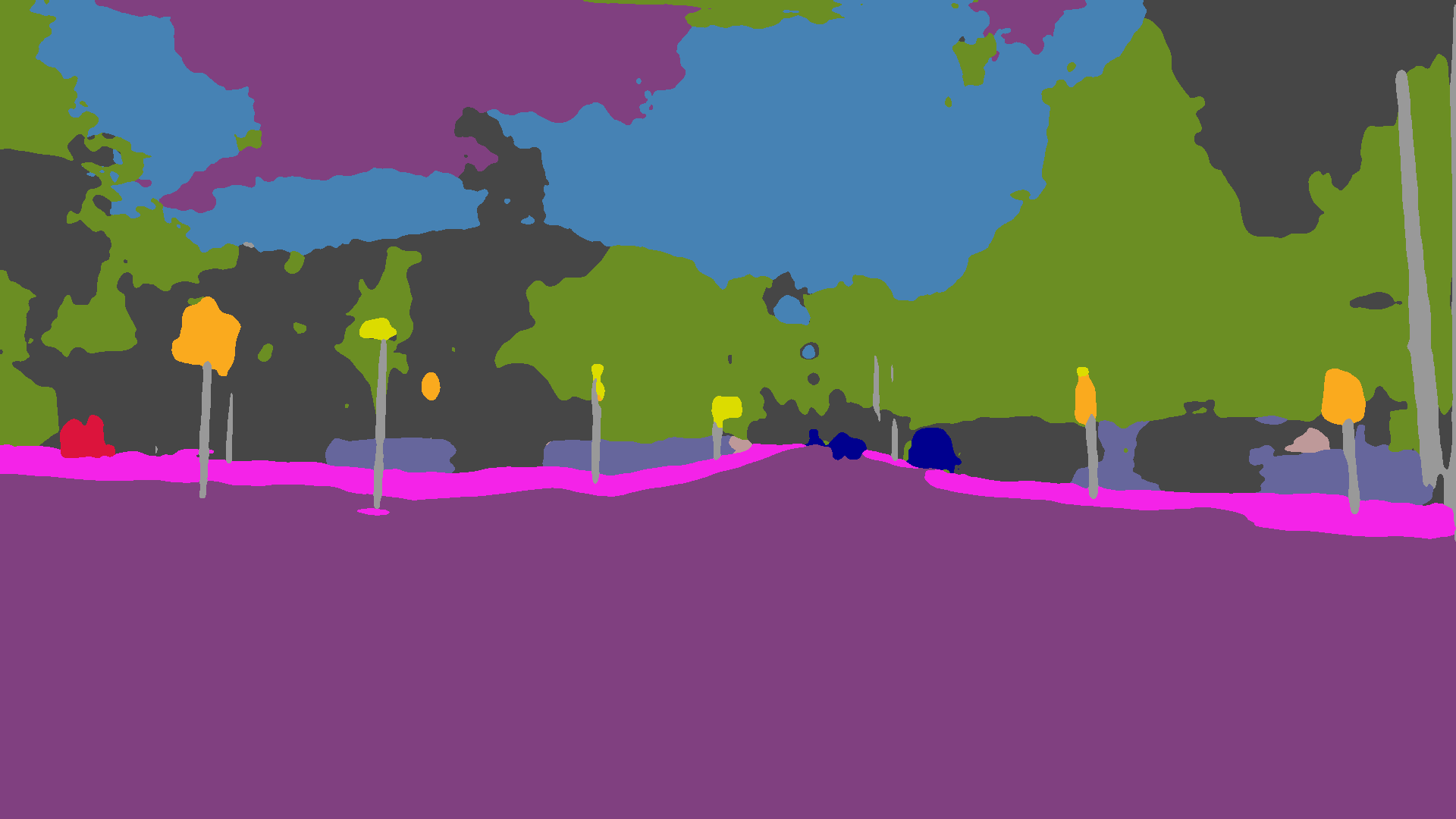}}
    \hfil
    \subfloat{\includegraphics[width=0.245\linewidth]{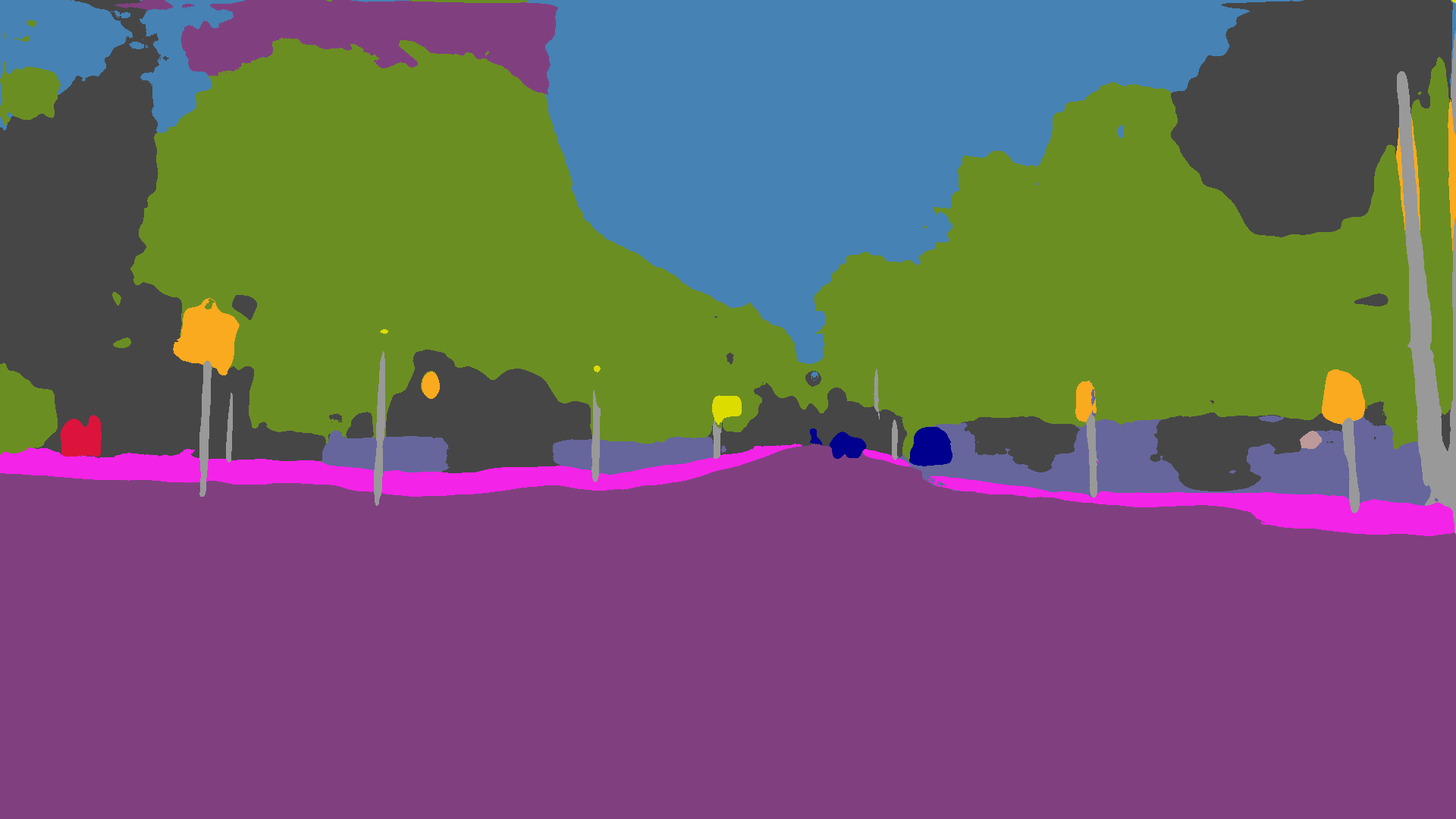}}
    \\
    \vspace{-0.3cm}
    \addtocounter{subfigure}{-4}
    \subfloat[Image]{\includegraphics[width=0.245\linewidth]{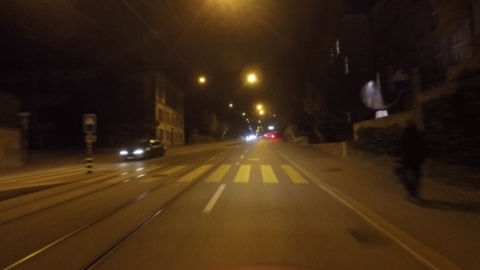}}
    \hfil
    \subfloat[Semantic GT]{\includegraphics[width=0.245\linewidth]{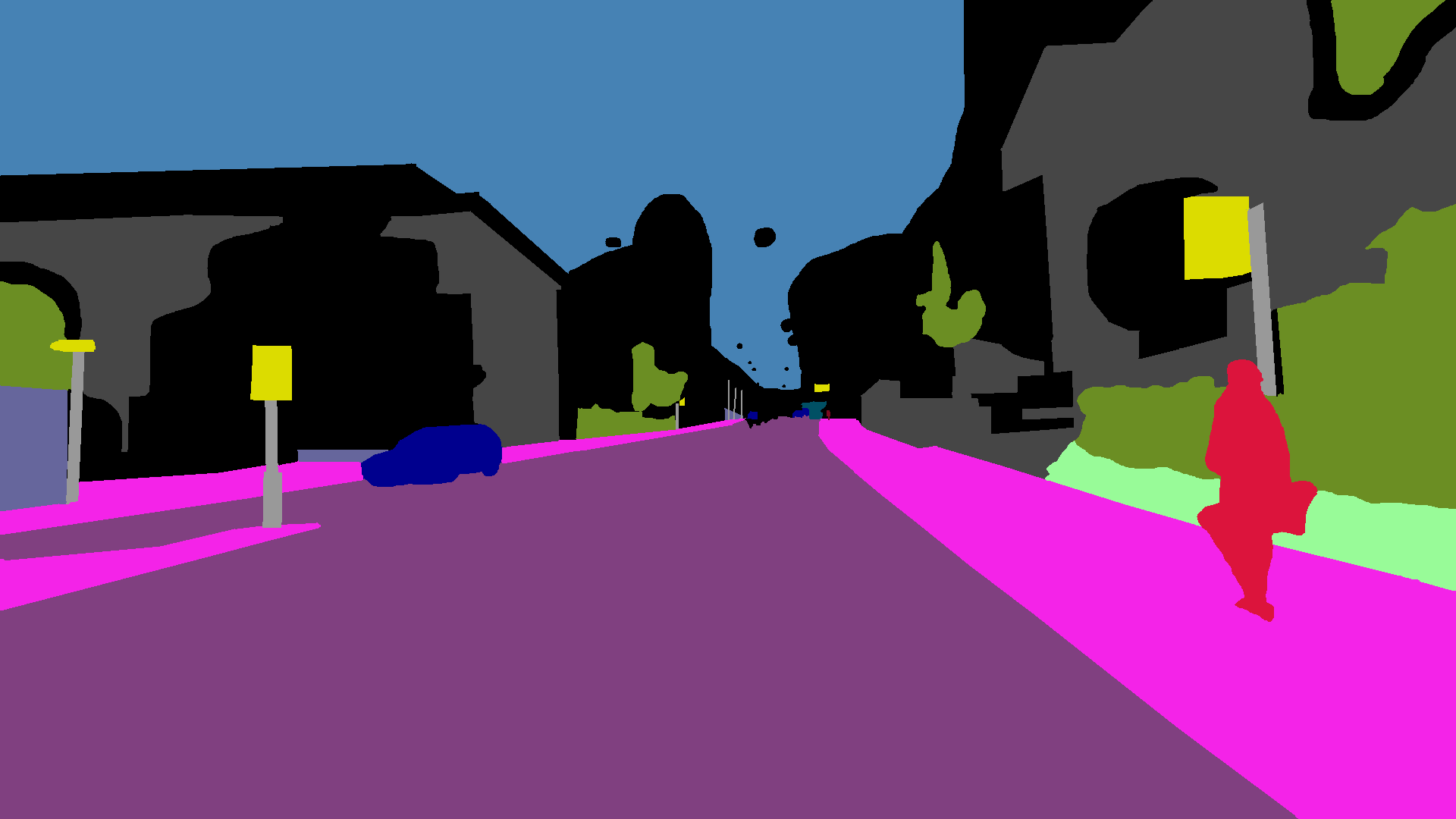}}
    \hfil
    \subfloat[MGCDA]{\includegraphics[width=0.245\linewidth]{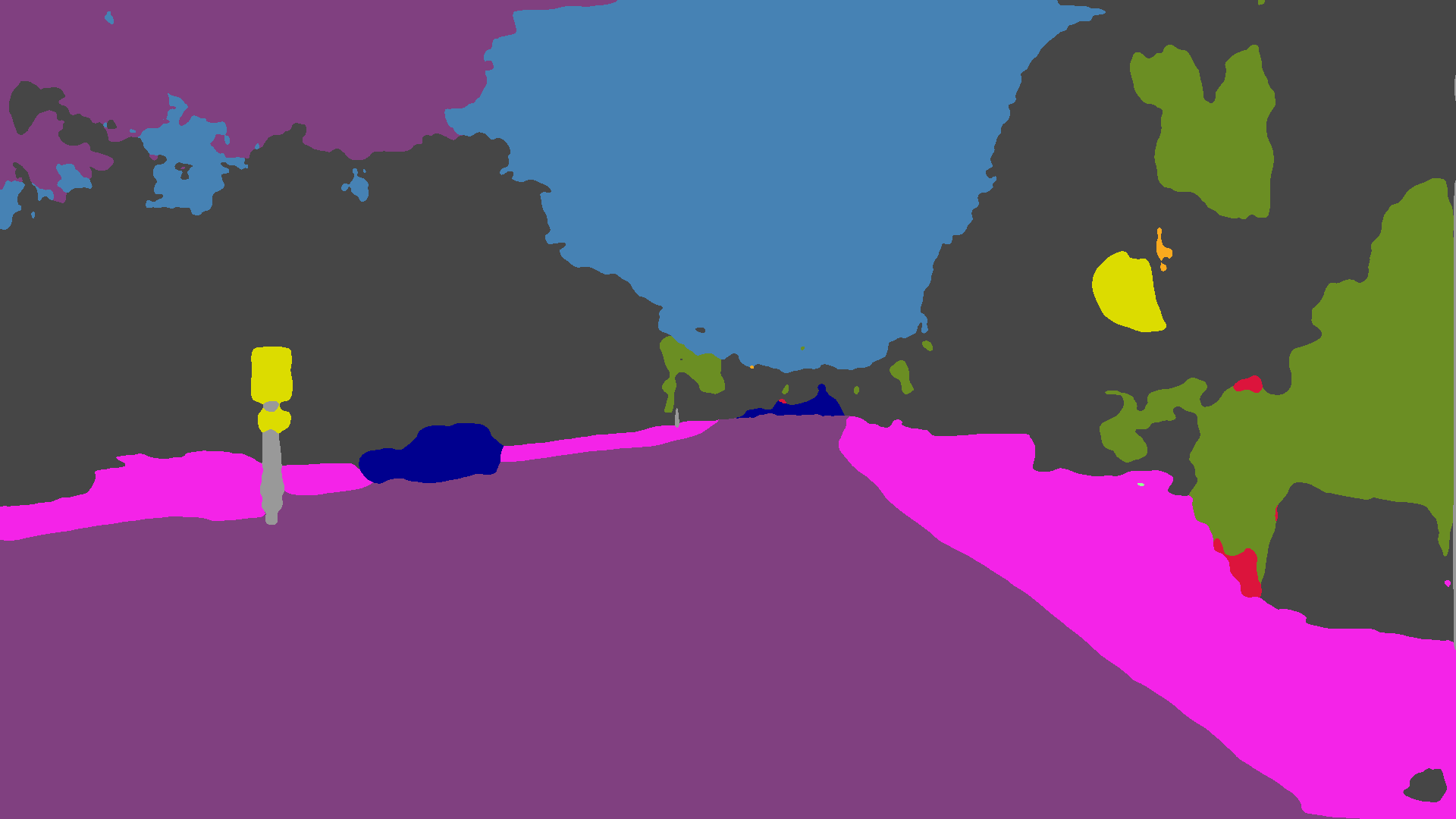}}
    \hfil
    \subfloat[MGCDA+TG]{\includegraphics[width=0.245\linewidth]{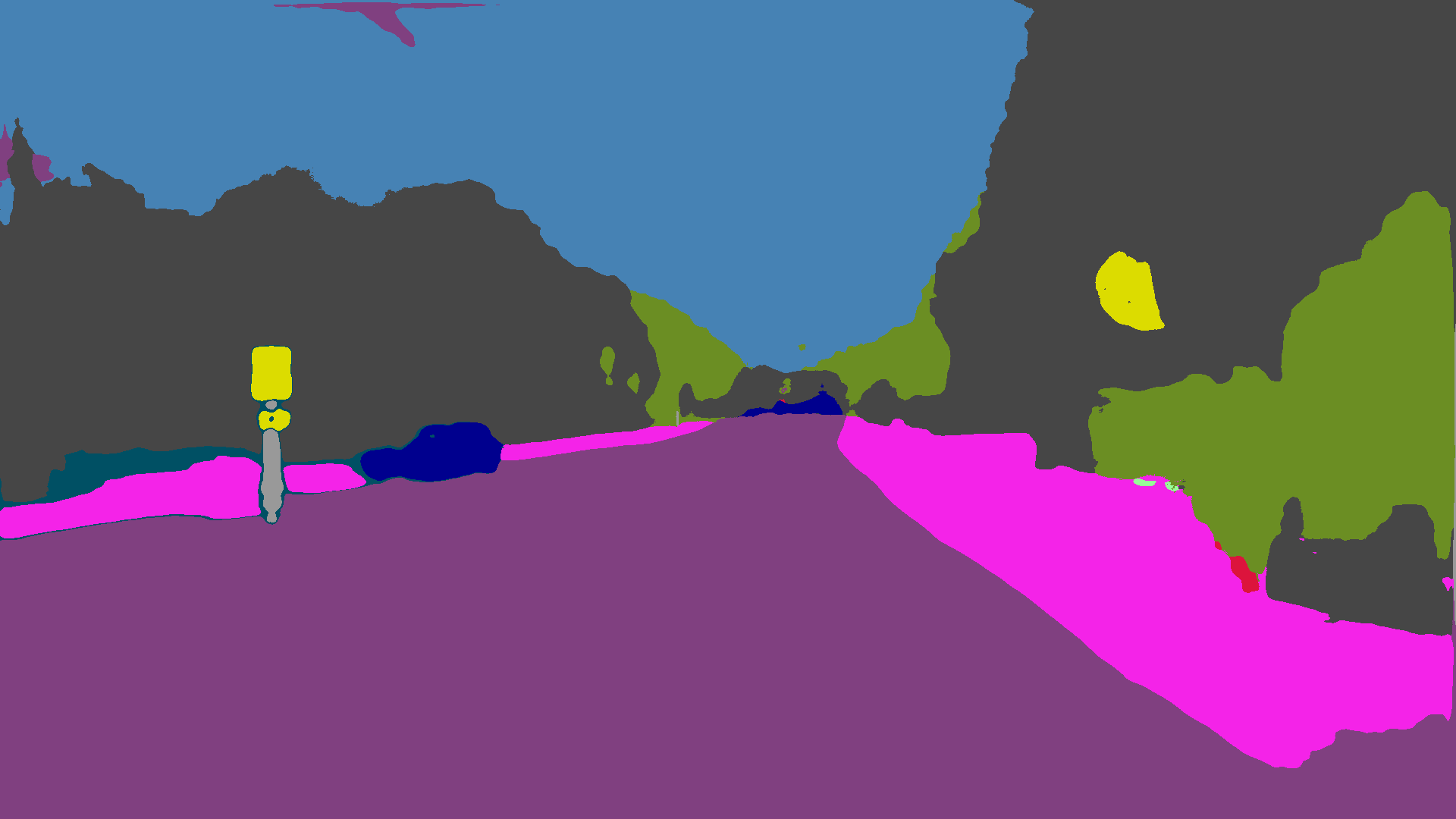}}
    \caption{Qualitative comparison on \emph{Dark Zurich-test} of MGCDA without and with map guidance at test time. TG stands for test-time guidance.}
    \label{fig:exp:test:guidance}
\end{figure}

In the exposition of our MGCDA method as well as in the preceding experiments, we have considered map guidance for segmentation refinement only in the training stage. However, guidance from maps is fully relevant at test time too, when e.g.\ the semantic segmentation model is deployed on an autonomous vehicle. To investigate this scenario, we consider two models, corresponding to MGCDA and DMAda~\cite{daytime:2:nighttime}, and compare in Table~\ref{table:exp:test:guidance:dark_zurich} the performance on \emph{Dark Zurich-test} using 1) the original predictions of the models, and 2) the refined predictions that are obtained after guided refinement using the predictions of RefineNet~\cite{refinenet} on the corresponding daytime images. The performance of both models is boosted significantly with the use of map guidance for refining the initial predictions, showing that our proposed geometrically guided segmentation refinement is applicable to and beneficial for more general semantic segmentation settings beyond our MGCDA framework. A visual comparison for map guidance at test time with MGCDA is included in Fig.~\ref{fig:exp:test:guidance}.

\subsection{Comparison with Preprocessing Baselines}
\label{sec:exp:preprocessing}
\begin{figure}
    \centering
    \subfloat{\includegraphics[width=0.16\linewidth]{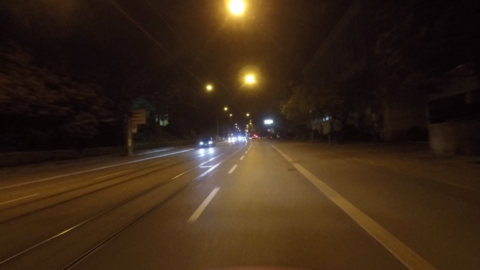}}
    \hfil
    \subfloat{\includegraphics[width=0.16\linewidth]{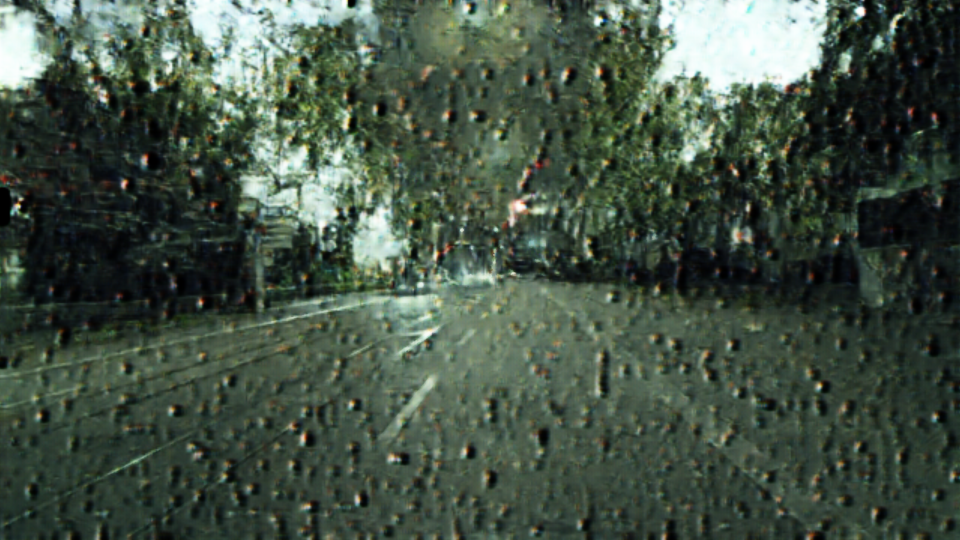}}
    \hfil
    \subfloat{\includegraphics[width=0.16\linewidth]{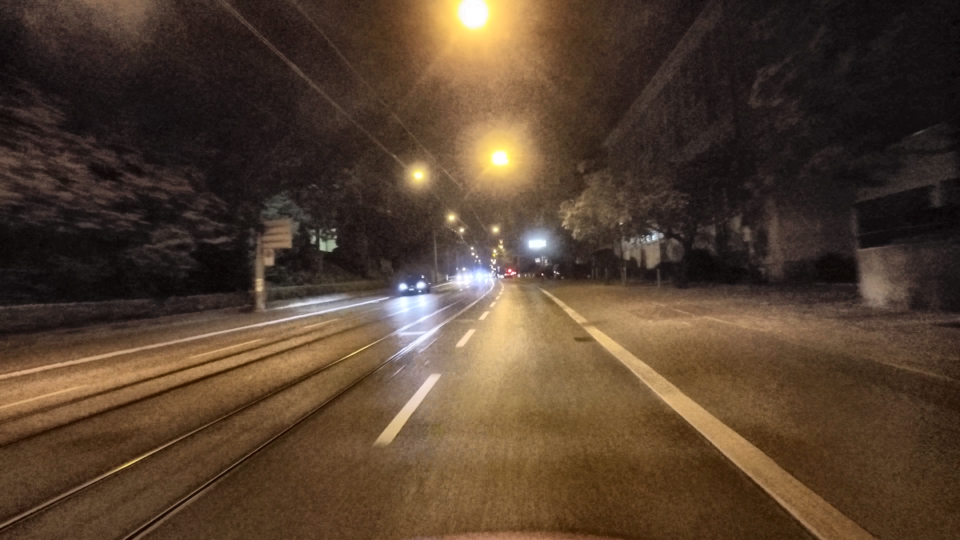}}
    \hfil
    \subfloat{\includegraphics[width=0.16\linewidth]{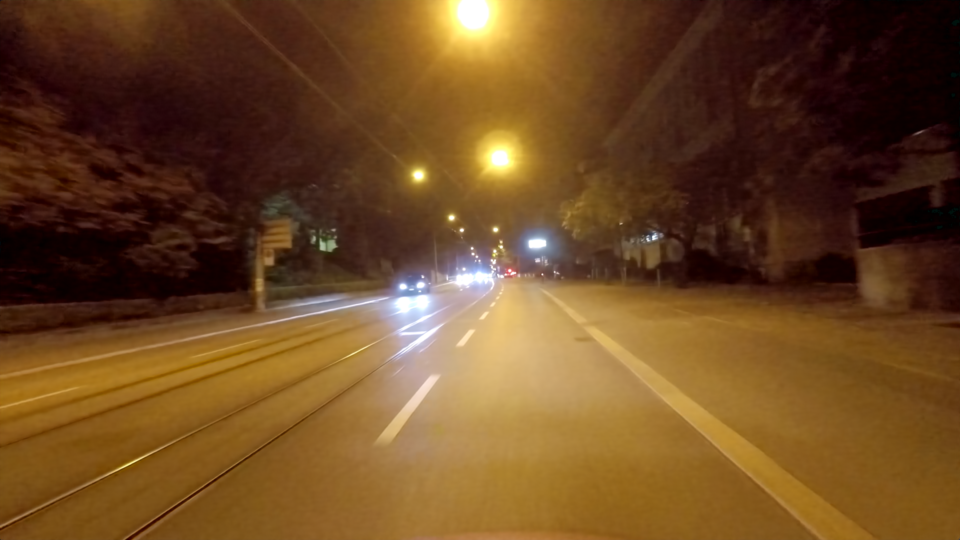}}
    \hfil
    \subfloat{\includegraphics[width=0.16\linewidth]{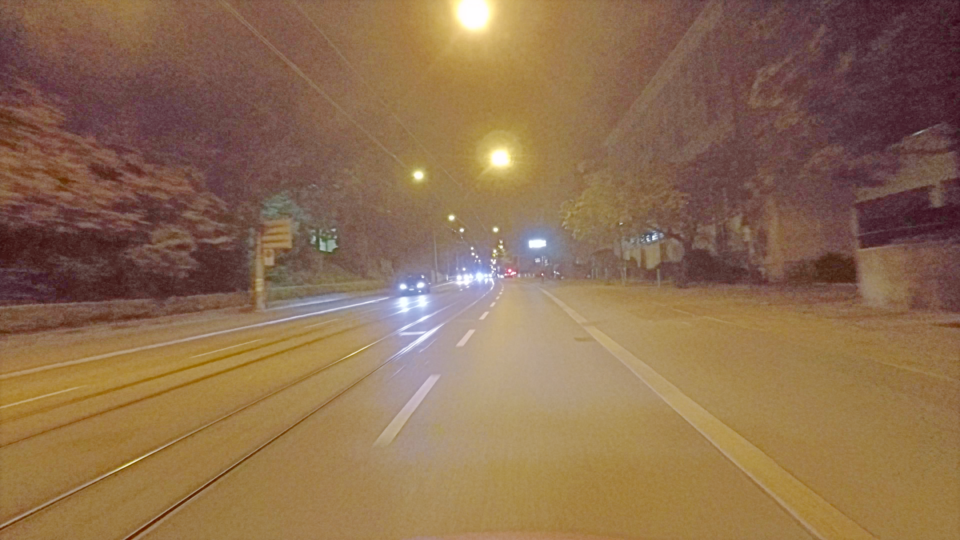}}
    \hfil
    \subfloat{\includegraphics[width=0.16\linewidth]{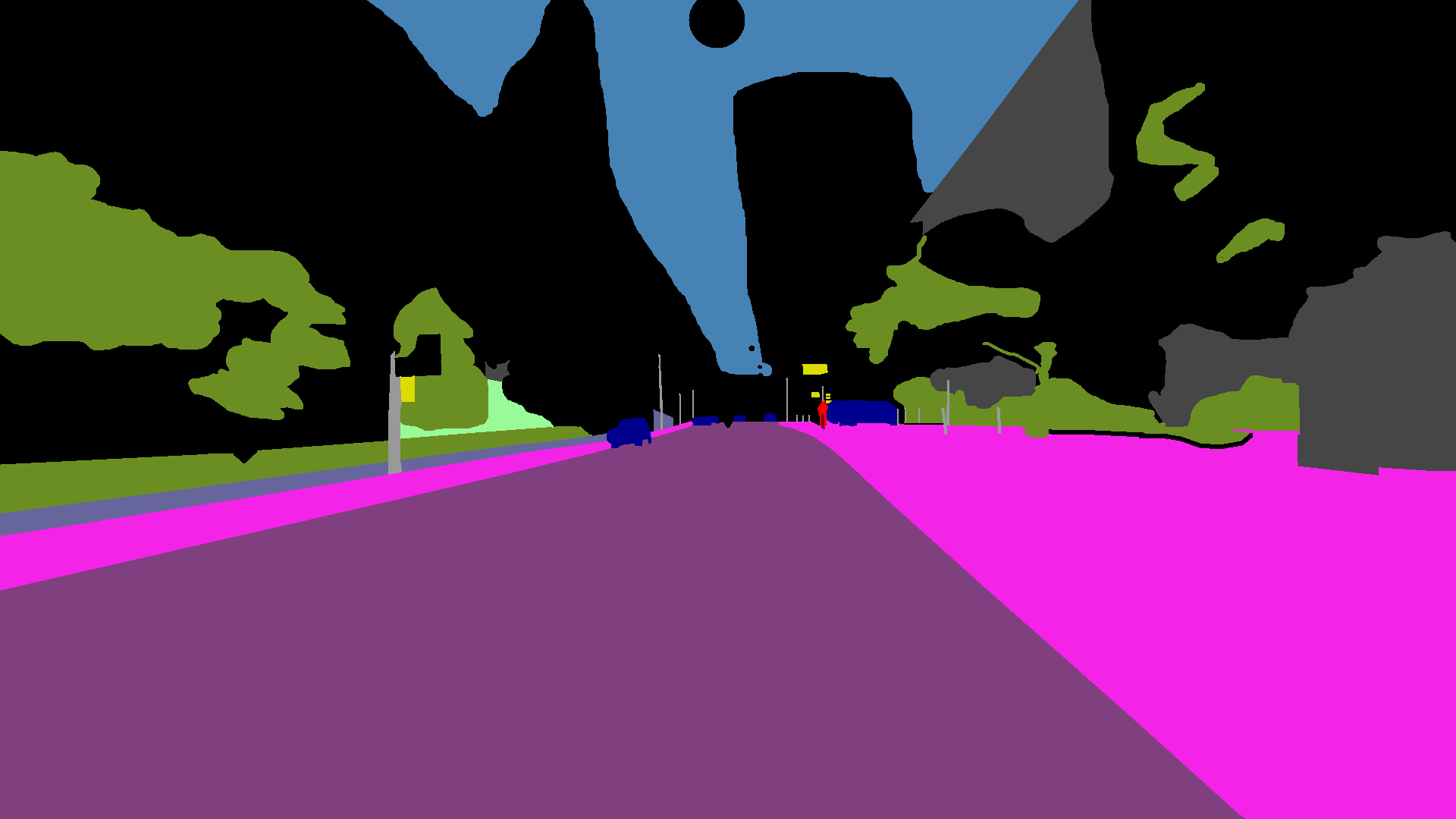}}
    \\
    \vspace{-0.3cm}
    \subfloat{\includegraphics[width=0.16\linewidth]{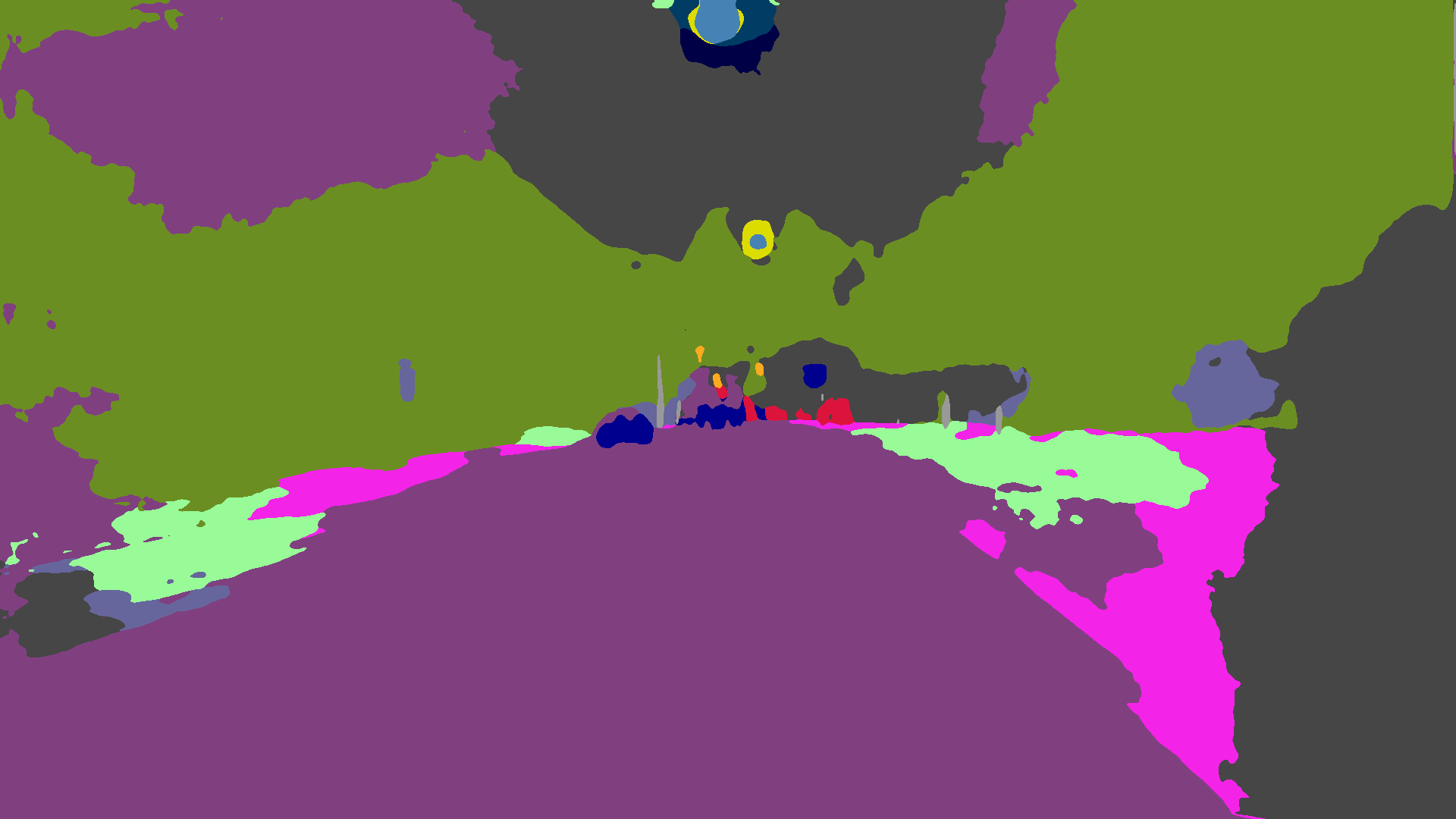}}
    \hfil
    \subfloat{\includegraphics[width=0.16\linewidth]{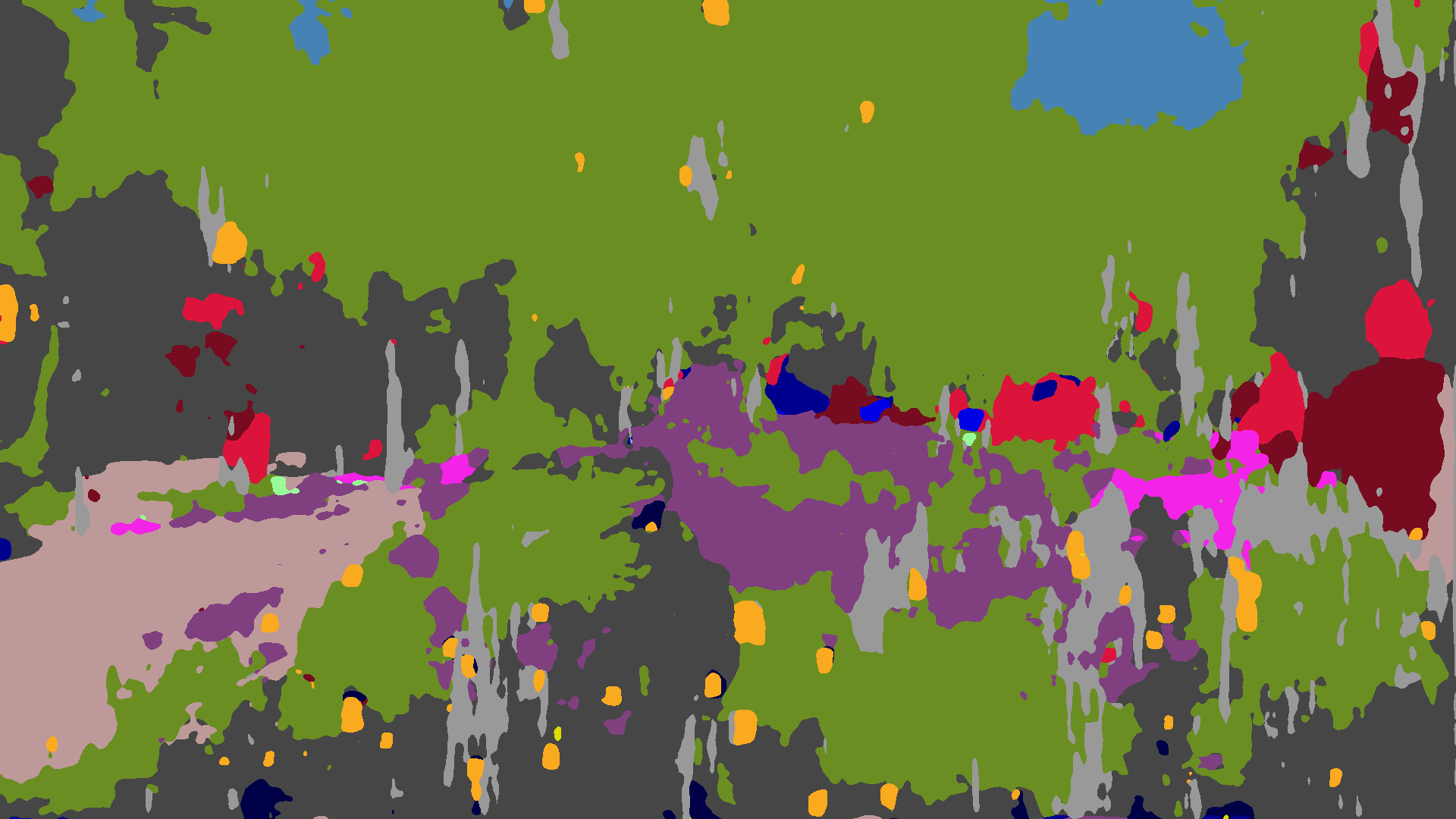}}
    \hfil
    \subfloat{\includegraphics[width=0.16\linewidth]{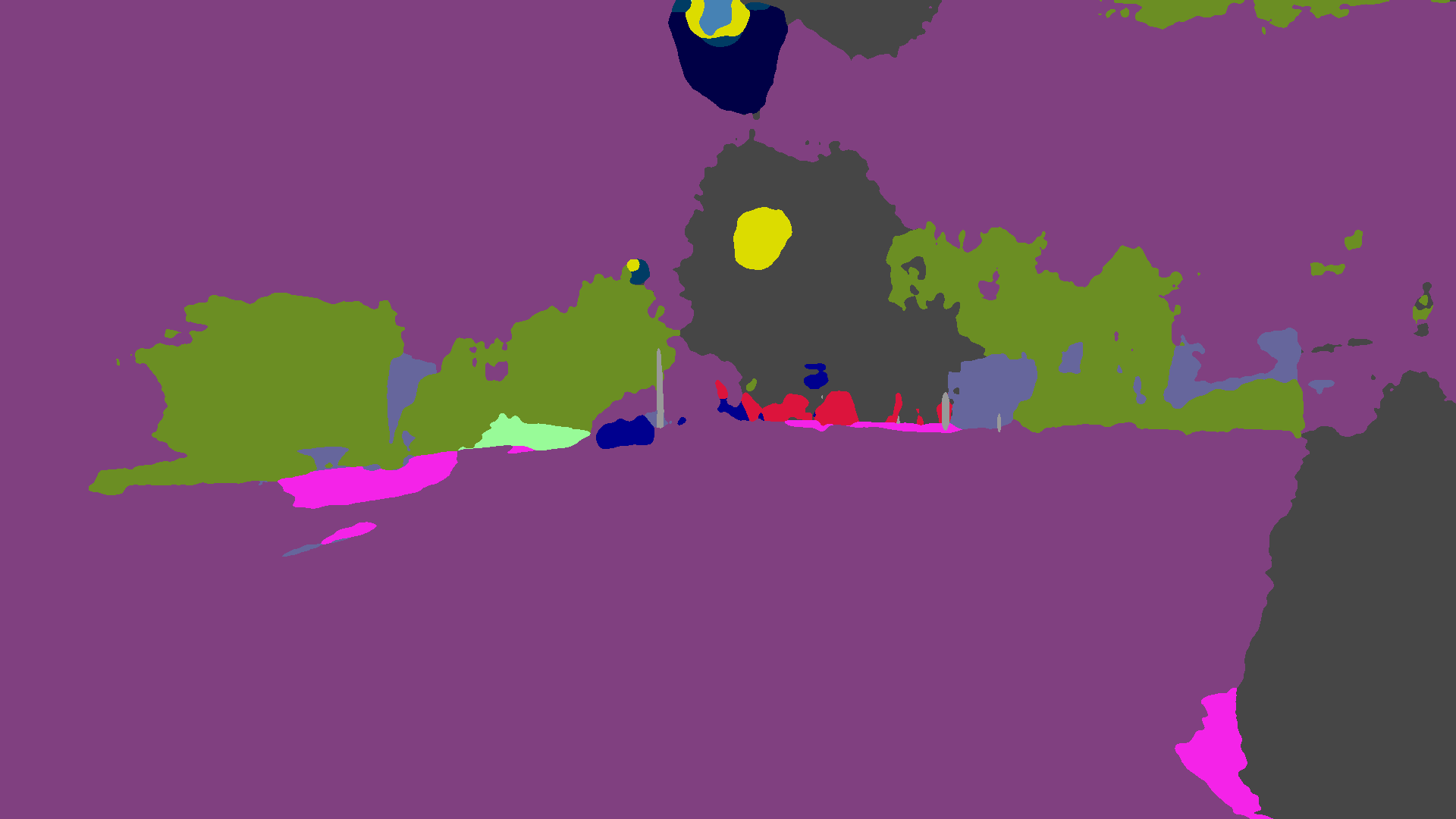}}
    \hfil
    \subfloat{\includegraphics[width=0.16\linewidth]{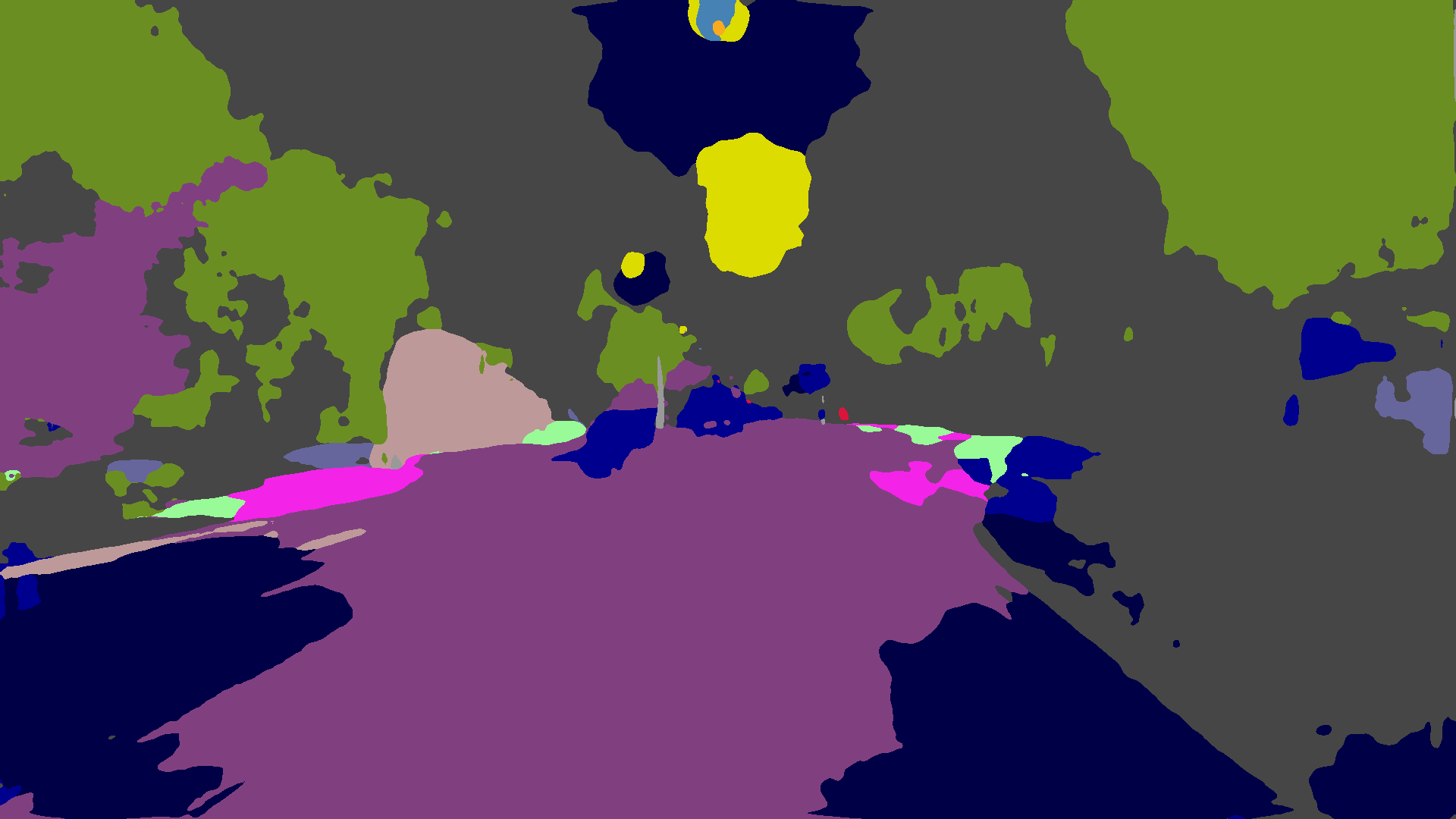}}
    \hfil
    \subfloat{\includegraphics[width=0.16\linewidth]{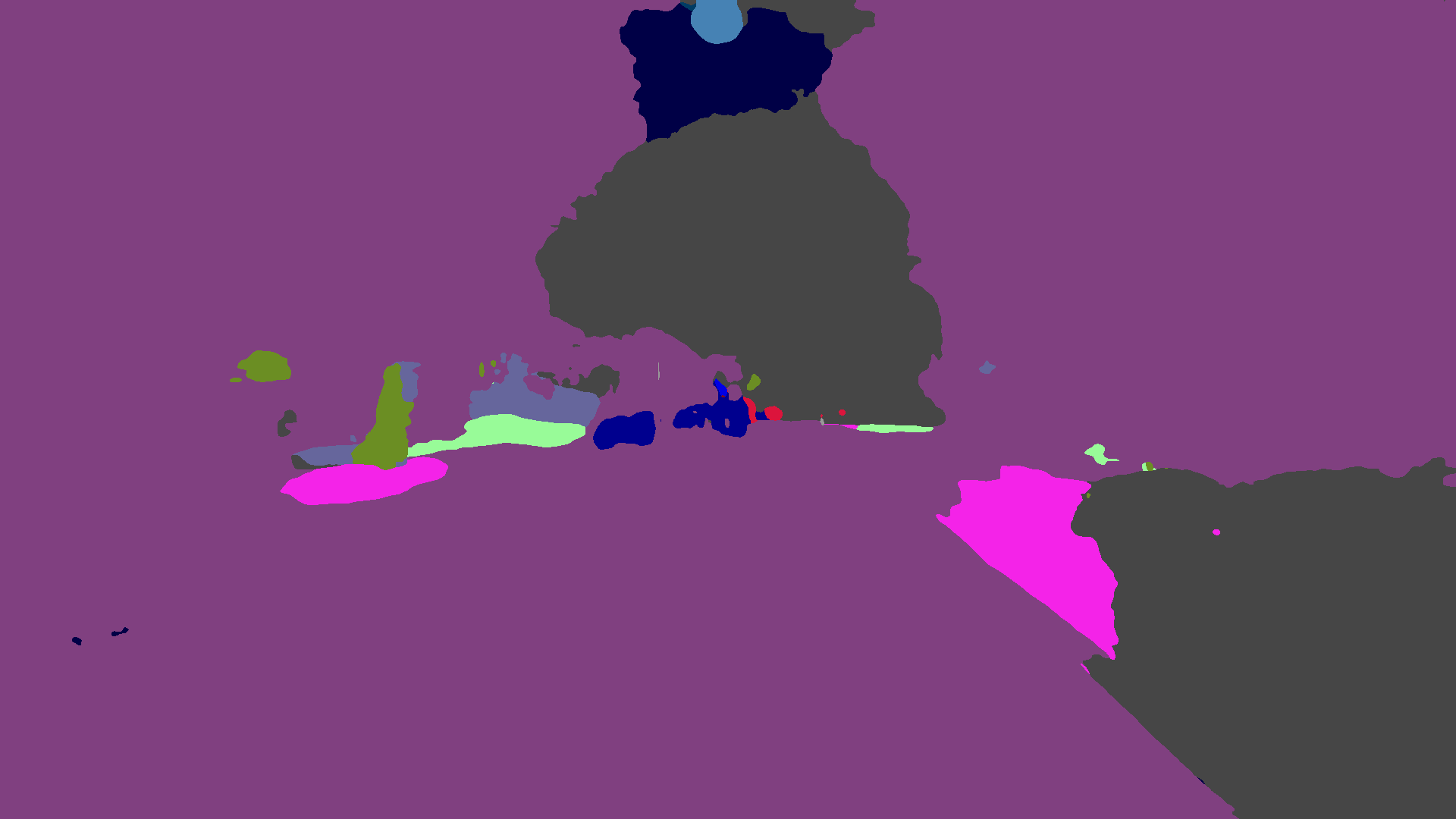}}
    \hfil
    \subfloat{\includegraphics[width=0.16\linewidth]{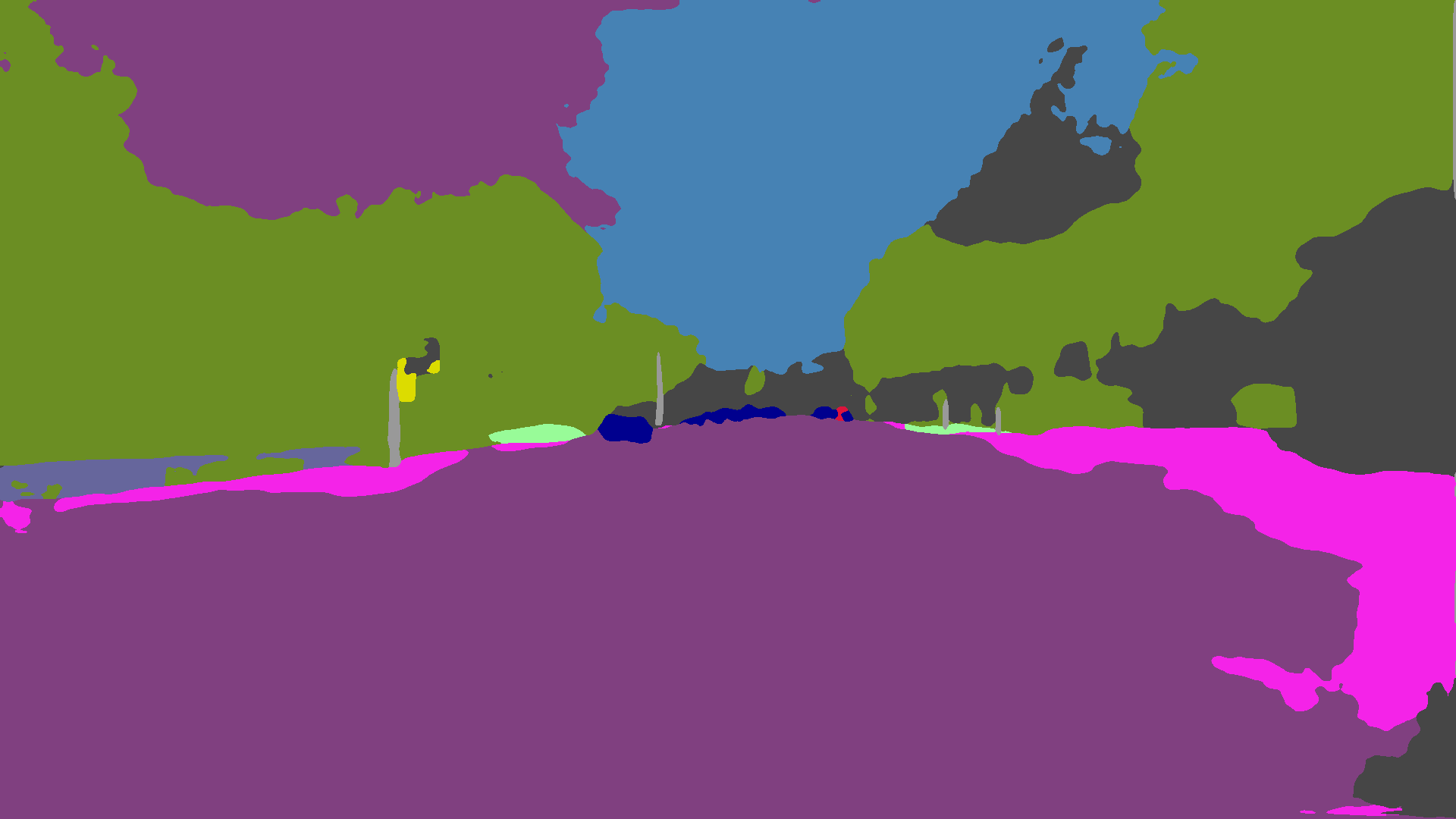}}
    \caption{Qualitative example from \emph{Dark Zurich-test} of effect of preprocessing on semantic predictions. From first to fifth column, left to right: image (top row) and prediction of daytime baseline (bottom row) for no preprocessing, CycleGAN, CLAHE, MBLLEN, and ZeroDCE. Sixth column: semantic GT (top row) and prediction of MGCDA (bottom row).}
    \label{fig:exp:preprocessing}
\end{figure}

For the sake of completeness, we consider the straightforward alternative to our approach of applying a preprocessing step to the images at test time and then using a pre-trained daytime segmentation model on the processed images to get the predictions. Such preprocessing can be accomplished via different approaches. We select the following representative methods for our experiment: ZeroDCE~\cite{ZeroDCE} and MBLLEN~\cite{MBLLEN} for low-light image enhancement, CLAHE~\cite{adaptive:histogram:equalization} for histogram equalization, and CycleGAN~\cite{cycleGAN} for image translation from nighttime to daytime. In Table~\ref{table:exp:preprocessing}, we compare the performance of the daytime RefineNet model on \emph{Dark Zurich-test} without preprocessing versus preprocessing with each of the aforementioned methods. A visual comparison is provided in Fig.~\ref{fig:exp:preprocessing}. Although preprocessing can generally enhance the contrast and visibility in nighttime images, it does not help improve the performance of the daytime model. The reason is the domain mismatch between the preprocessed nighttime images at test time and the daytime images presented to the segmentation model at training time. The finding that contrary to expectation, mere test-time preprocessing does not benefit segmentation at nighttime is in congruence with the results of~\cite{SFSU_synthetic,CMAda:IJCV2020}, where dehazing preprocessing is also found not to help segmentation in fog. To close the aforementioned domain gap, we additionally experiment with enhancement both at training and test time with the state-of-the-art ZeroDCE, adapting the daytime model on ZeroDCE-enhanced versions of synthetic nighttime Cityscapes images generated with CycleGAN. The adapted model achieves 38.4\% mIoU, significantly outperforming the daytime model but falling short of the model from Table~\ref{table:exp:cyclegan:dark_zurich} adapted on synthetic nighttime images without enhancement, as additional artifacts are introduced to the images, which makes the image domain harder. This comparison reveals the limitations of preprocessing in serving adaptation of segmentation to nighttime and stresses the need for a more sophisticated learning scheme such as MGCDA.

\begin{table}[!tb]
  \caption{Comparison on \emph{Dark Zurich-test} of different preprocessing baselines, using the daytime RefineNet~\cite{refinenet} model for predictions and reporting mIoU (\%). No PP stands for no preprocessing.}
  \label{table:exp:preprocessing}
  \centering
  \setlength\tabcolsep{4pt}
  \footnotesize
  \begin{tabular}{ccccc}
  \toprule
  No PP & CycleGAN~\cite{cycleGAN} & CLAHE~\cite{adaptive:histogram:equalization} & MBLLEN~\cite{MBLLEN} & ZeroDCE~\cite{ZeroDCE}\\
  \midrule
  28.5 & 7.2 & 23.0 & 22.4 & 17.9\\
  \bottomrule
  \\
  \end{tabular}
\end{table}

\subsection{Comparisons with UIoU}
\label{sec:exp:uiou}
\begin{figure}[!tb]
    \centering
    \includegraphics[clip,width=\linewidth,trim=25mm 100mm 25mm 100mm]{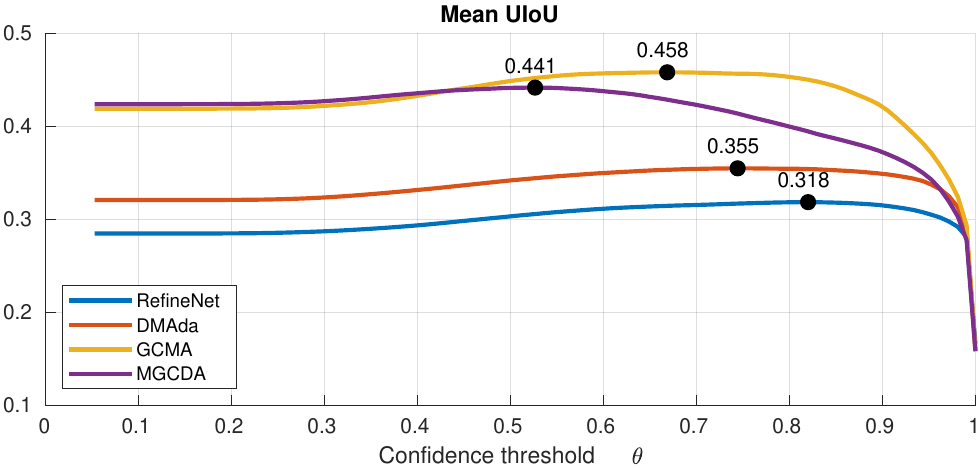}
    \caption{Uncertainty-aware evaluation of RefineNet~\cite{refinenet}, DMAda~\cite{daytime:2:nighttime}, GCMA and MGCDA on \emph{Dark Zurich-test}. We evaluate mean UIoU across the entire range $[1/C,\,1]$ of confidence threshold $\theta$. For each method, the point at which mean UIoU is maximized is marked black and labeled with this maximum mean UIoU value.}
    \label{fig:exp:uiou}
\end{figure}

In Fig.~\ref{fig:exp:uiou}, we use our novel UIoU metric to evaluate MGCDA and GCMA against DMAda and our baseline RefineNet model on \emph{Dark Zurich-test} for varying confidence threshold $\theta$ and plot the resulting mean UIoU$(\theta)$ curves. Note that standard mean IoU can be read out from the leftmost point of each curve. First, our expectation based on Th.~\ref{thm:UIoU:greater:iou} is confirmed for all methods, i.e.\ maximum UIoU values over the range of $\theta$ are larger than IoU by ca.\ 2--3\%. This implies that on \emph{Dark Zurich-test}, these models generally have lower confidence on invalid regions than valid ones, although they do not explicitly model uncertainty, as there is no distinction between valid and invalid regions during training. Modeling confidence explicitly along the lines of \cite{uncertainty:bayesian,simultaneous:segmentation:outliers} could further increase UIoU and is an interesting direction for future work. Second, the comparative performance of the methods is the same across all values of $\theta$, except the pair of MGCDA and GCMA. MGCDA slightly outperforms GCMA for low values of $\theta$, but for high values GCMA achieves significantly higher UIoU, which implies that MGCDA places relatively low confidence on valid regions, which prevents its UIoU from increasing further as $\theta$ increases. In any case, MGCDA and GCMA substantially outperform the other two methods. Overall, UIoU is generally consistent with standard IoU and is a suitable substitute of the latter in adverse settings where selective invalidation of the predictions is relevant.

\section{Conclusion}
\label{sec:conclusion}

In this paper, we have introduced MGCDA, a method to gradually adapt semantic segmentation models from daytime to nighttime with stylized data and unlabeled real data of increasing darkness, as well as UIoU, a novel evaluation metric for semantic segmentation designed for images with indiscernible content. We have also presented \emph{Dark Zurich}, a large-scale dataset of real scenes captured at multiple times of day with cross-time-of-day correspondences, and annotated 201 nighttime scenes of it with a new protocol which enables our evaluation. Detailed evaluation with standard IoU on real nighttime sets demonstrates the merit of MGCDA, which substantially improves upon competing state-of-the-art methods. Finally, evaluation on our benchmark with UIoU shows that invalidating predictions is useful when the input includes ambiguous content.

\appendices
\section{Proof of Theorem \ref{thm:UIoU:greater:iou}}
\label{supp:sec:proof}

\begin{proof}
For brevity in the proof, we drop the class superscript $(c)$ which is used in the statement of the theorem.

Firstly, we draw an association between pixel sets related to the standard $\text{IoU} = \text{UIoU}(1/C)$ and their counterparts for UIoU defined in \eqref{eq:tp}--\eqref{eq:fi}. In particular, the following holds true:
\begin{multline} \label{eq:proof:true:positive:false:negative:iou}
|\text{TP}(1/C)| + |\text{FN}(1/C)| \\
= |\text{TP}(\theta)| + |\text{FN}(\theta)| + |\text{TI}(\theta)| + |\text{FI}(\theta)|,\,\forall \theta \in [1/C,\,1].
\end{multline}
The first assumption of Th.~\ref{thm:UIoU:greater:iou} implies that $\text{FI}(\theta_1) = \emptyset$, because $\forall \theta < \theta_2$ (including $\theta_1$) there exists no false invalid pixel for the examined class. Thus, applying \eqref{eq:proof:true:positive:false:negative:iou} for $\theta = \theta_1$ leads to
\begin{equation} \label{eq:proof:true:positive:iou:theta1}
|\text{TP}(1/C)| \\
= |\text{TP}(\theta_1)| + |\text{TI}(\theta_1)| + |\text{FN}(\theta_1)| - |\text{FN}(1/C)|.
\end{equation}

Secondly, we plug the proposition of the first assumption of the theorem into the proposition of the second assumption to obtain
\begin{equation} \label{eq:proof:nonempty:improved:init}
\left(\text{FN}(1/C) \cup \text{FP}(1/C)\right) \setminus \left(\text{FN}(\theta_1) \cup \text{FP}(\theta_1)\right) \neq \emptyset.
\end{equation}
We further elaborate on \eqref{eq:proof:nonempty:improved:init} by observing that $\text{FN}(1/C) \cap \text{FP}(1/C) = \emptyset$, $\text{FN}(\theta_1) \subseteq \text{FN}(1/C)$ and $\text{FP}(\theta_1) \subseteq \text{FP}(1/C)$ to arrive at
\begin{equation} \label{eq:proof:nonempty:improved}
(|\text{FN}(1/C)| - |\text{FN}(\theta_1)|) + (|\text{FP}(1/C)| - |\text{FP}(\theta_1)|) > 0.
\end{equation}

Both terms on the left-hand side of \eqref{eq:proof:nonempty:improved} are nonnegative based on our previous observations, while at the same time \eqref{eq:proof:nonempty:improved} implies that at least one of the two is strictly positive. To complete the proof, we distinguish between the two corresponding cases.

\begin{figure*}[!tb]
    \centering
    \subfloat{\includegraphics[width=0.195\textwidth]{figures/GP010364_frame_000382_image.png}}
    \hfil
    \subfloat{\includegraphics[width=0.195\textwidth]{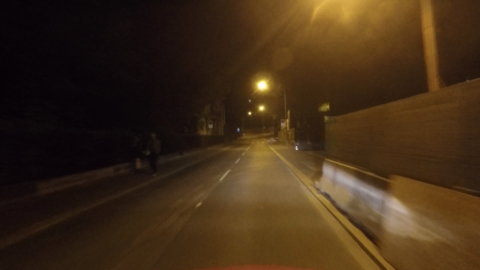}}
    \hfil
    \subfloat{\includegraphics[width=0.195\textwidth]{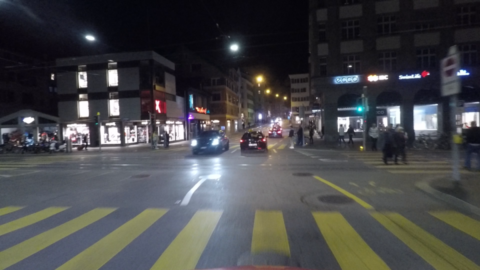}}
    \hfil
    \subfloat{\includegraphics[width=0.195\textwidth]{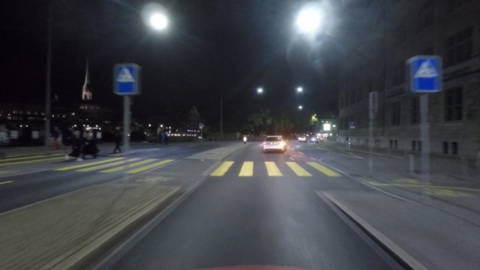}}
    \hfil
    \subfloat{\includegraphics[width=0.195\textwidth]{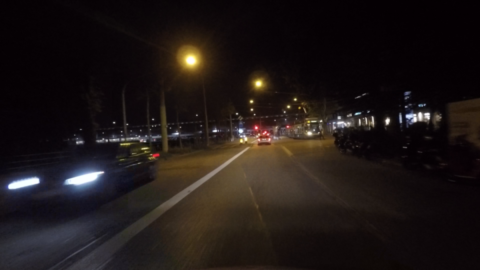}}
    \\
    \vspace{-0.3cm}
    \subfloat{\includegraphics[width=0.195\textwidth]{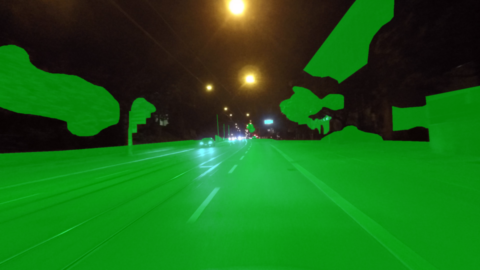}}
    \hfil
    \subfloat{\includegraphics[width=0.195\textwidth]{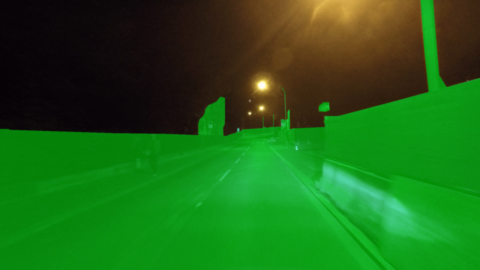}}
    \hfil
    \subfloat{\includegraphics[width=0.195\textwidth]{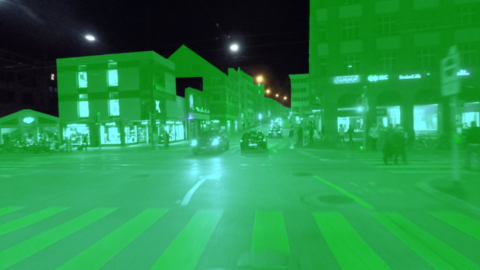}}
    \hfil
    \subfloat{\includegraphics[width=0.195\textwidth]{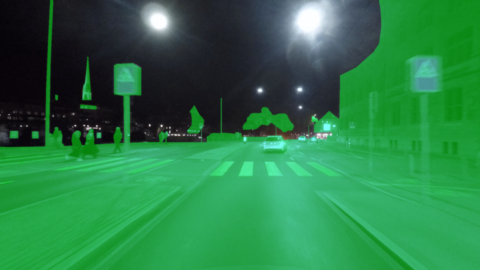}}
    \hfil
    \subfloat{\includegraphics[width=0.195\textwidth]{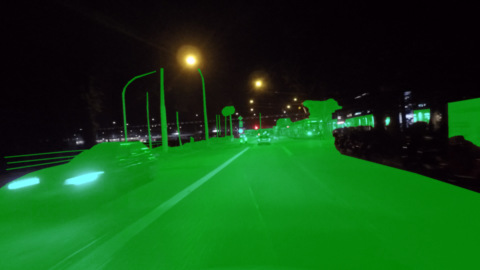}}
    \\
    \vspace{-0.3cm}
    \subfloat{\includegraphics[width=0.195\textwidth]{figures/GP010364_frame_000382_gt.png}}
    \hfil
    \subfloat{\includegraphics[width=0.195\textwidth]{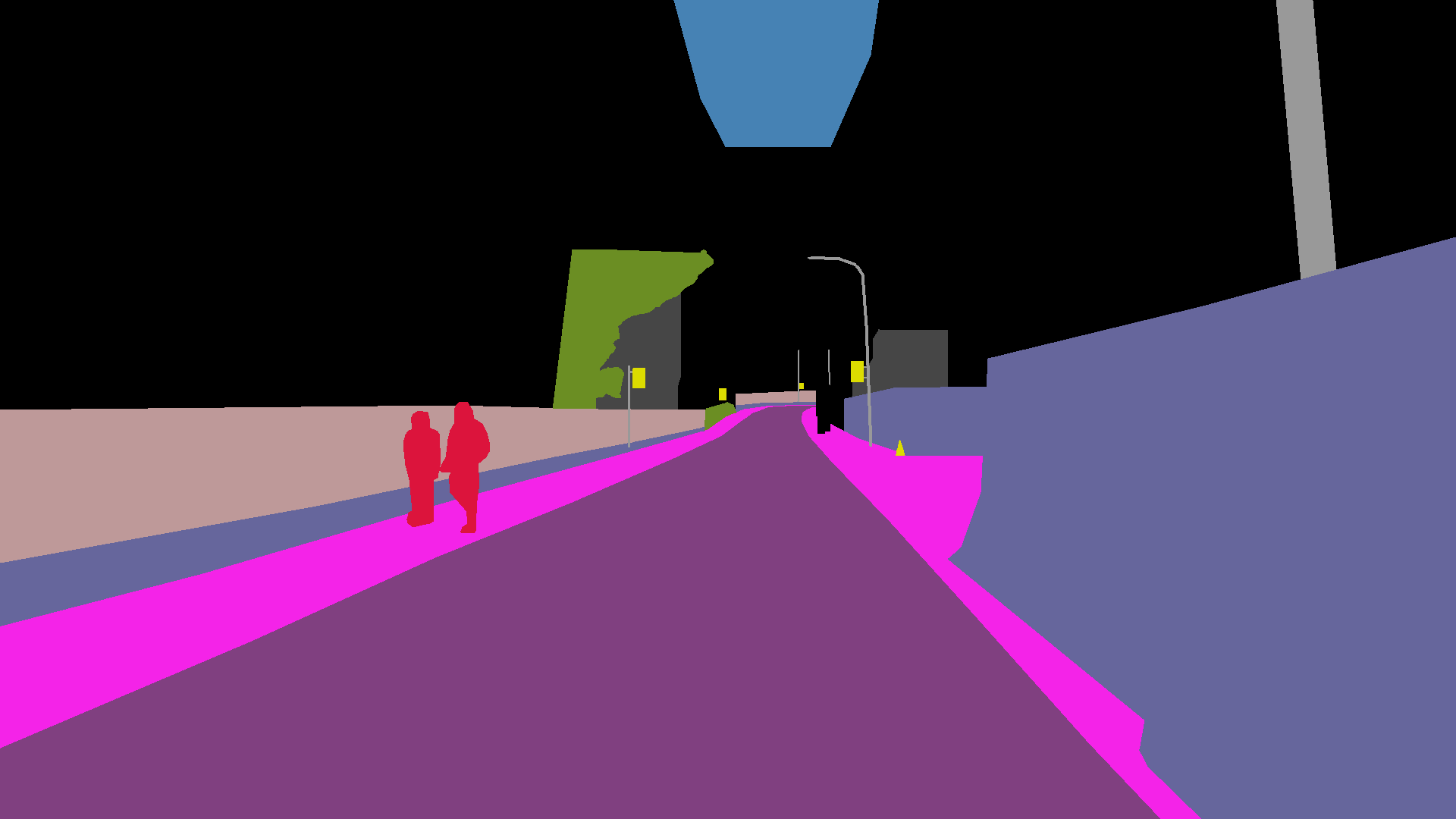}}
    \hfil
    \subfloat{\includegraphics[width=0.195\textwidth]{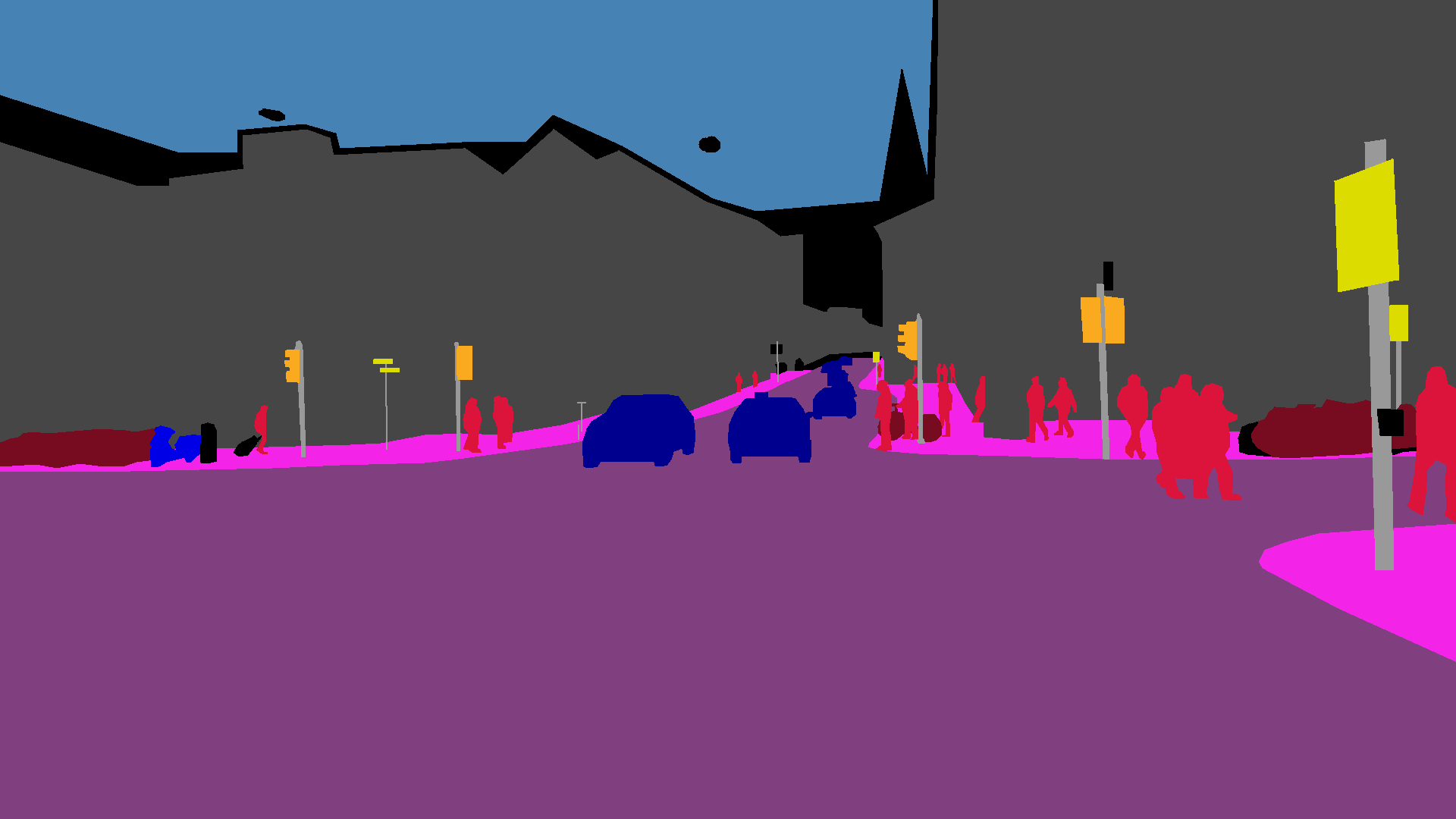}}
    \hfil
    \subfloat{\includegraphics[width=0.195\textwidth]{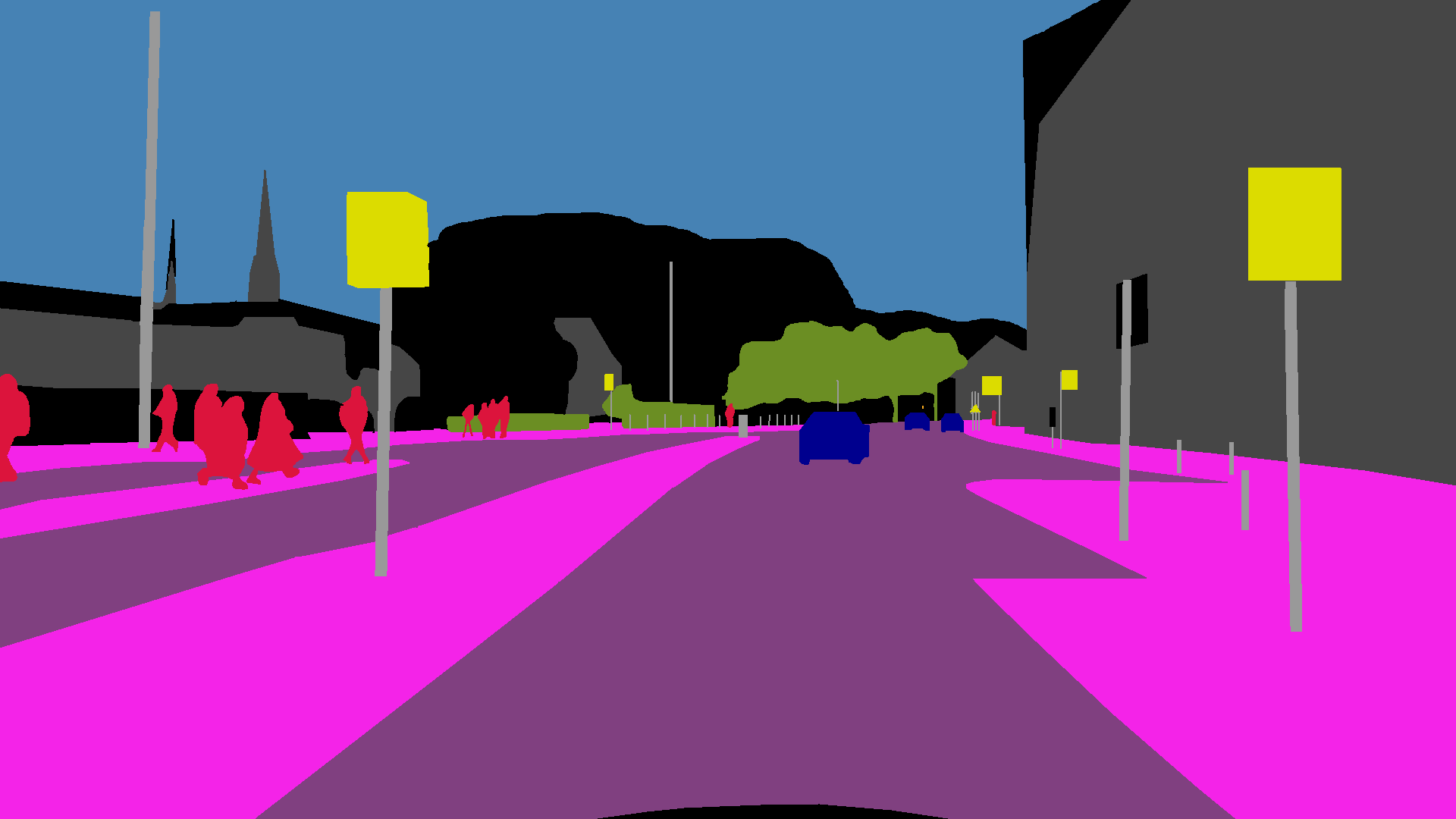}}
    \hfil
    \subfloat{\includegraphics[width=0.195\textwidth]{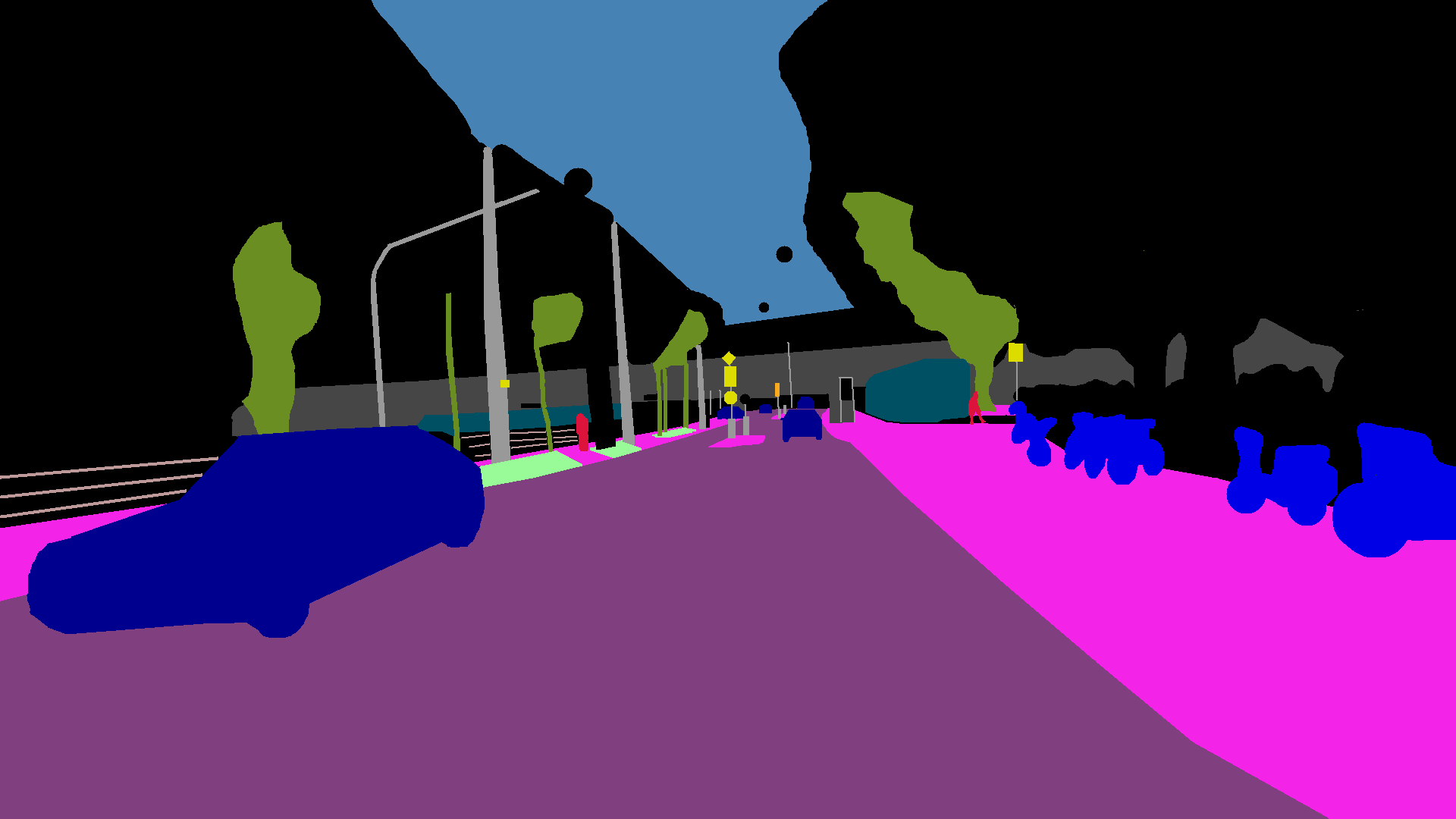}}
    \\
    \vspace{-0.3cm}
    \subfloat{\includegraphics[width=0.195\textwidth]{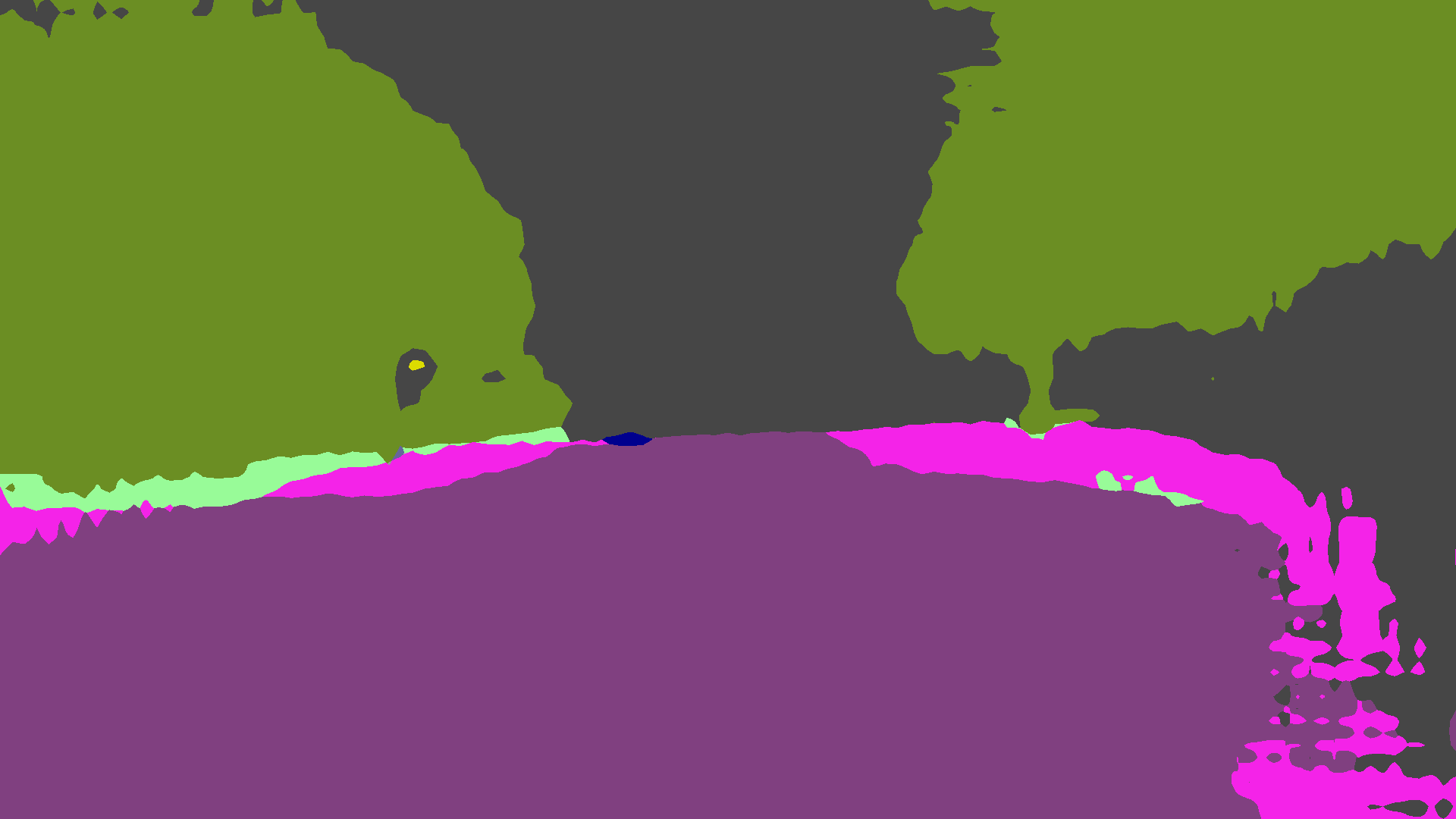}}
    \hfil
    \subfloat{\includegraphics[width=0.195\textwidth]{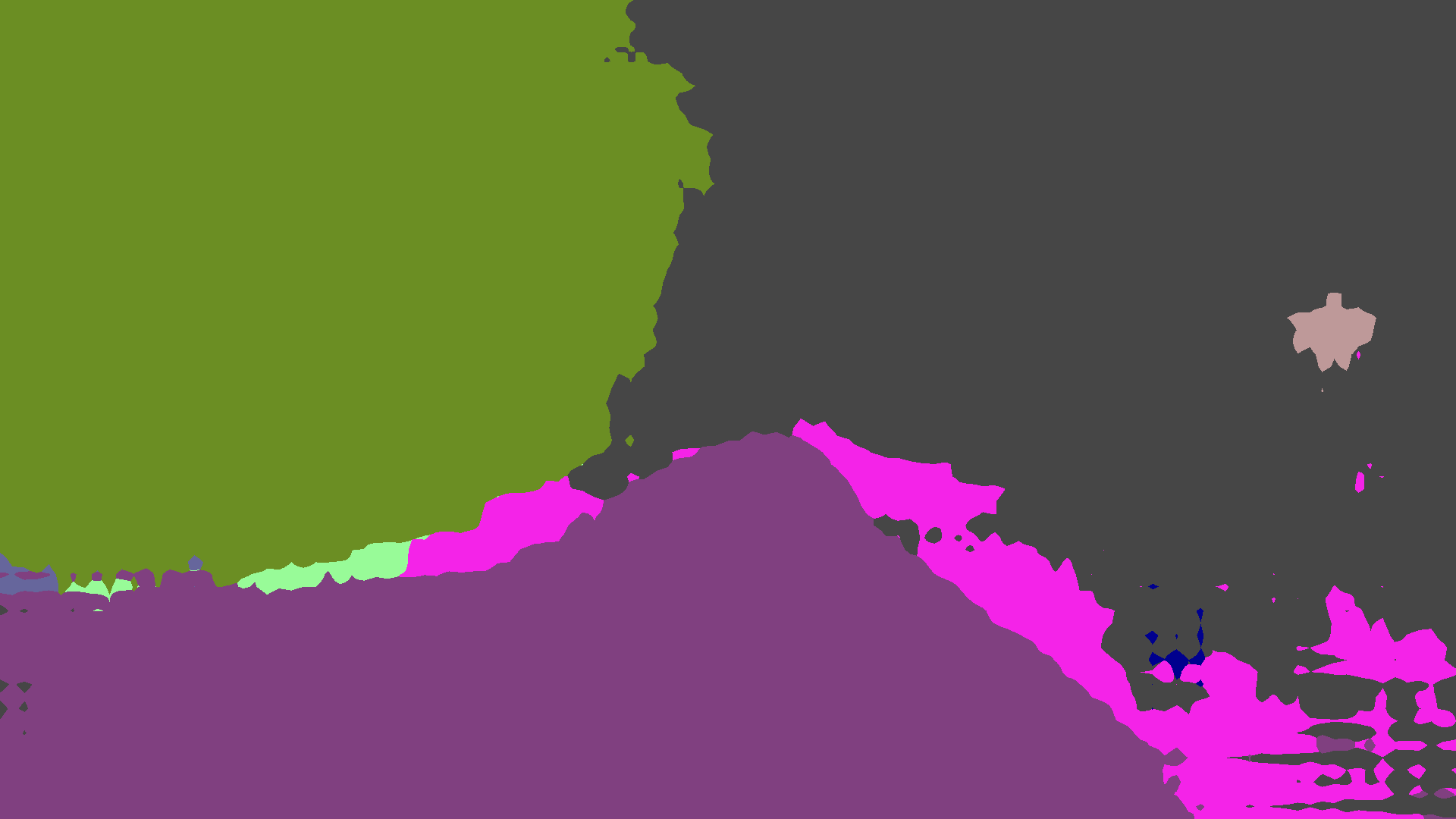}}
    \hfil
    \subfloat{\includegraphics[width=0.195\textwidth]{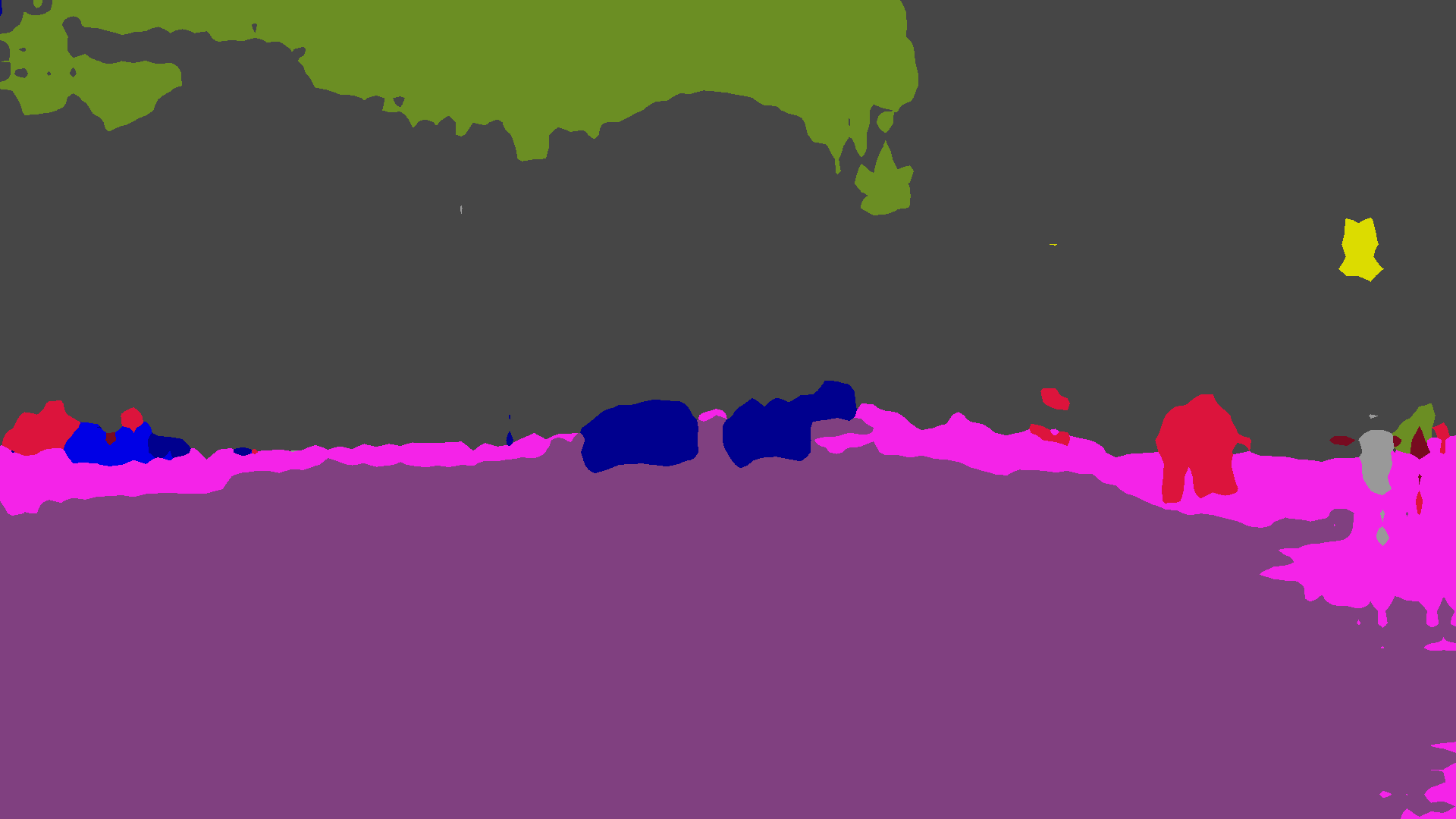}}
    \hfil
    \subfloat{\includegraphics[width=0.195\textwidth]{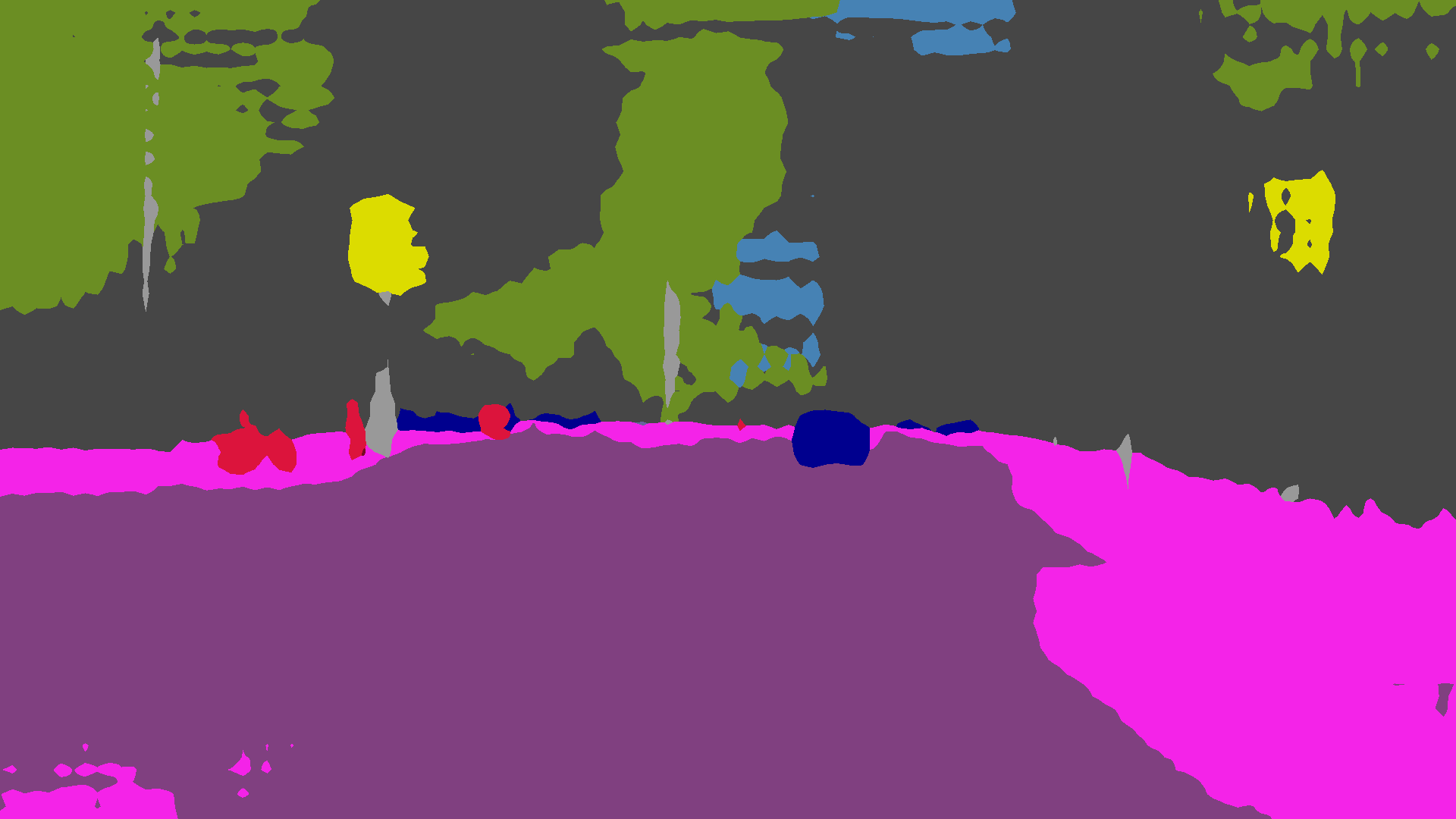}}
    \hfil
    \subfloat{\includegraphics[width=0.195\textwidth]{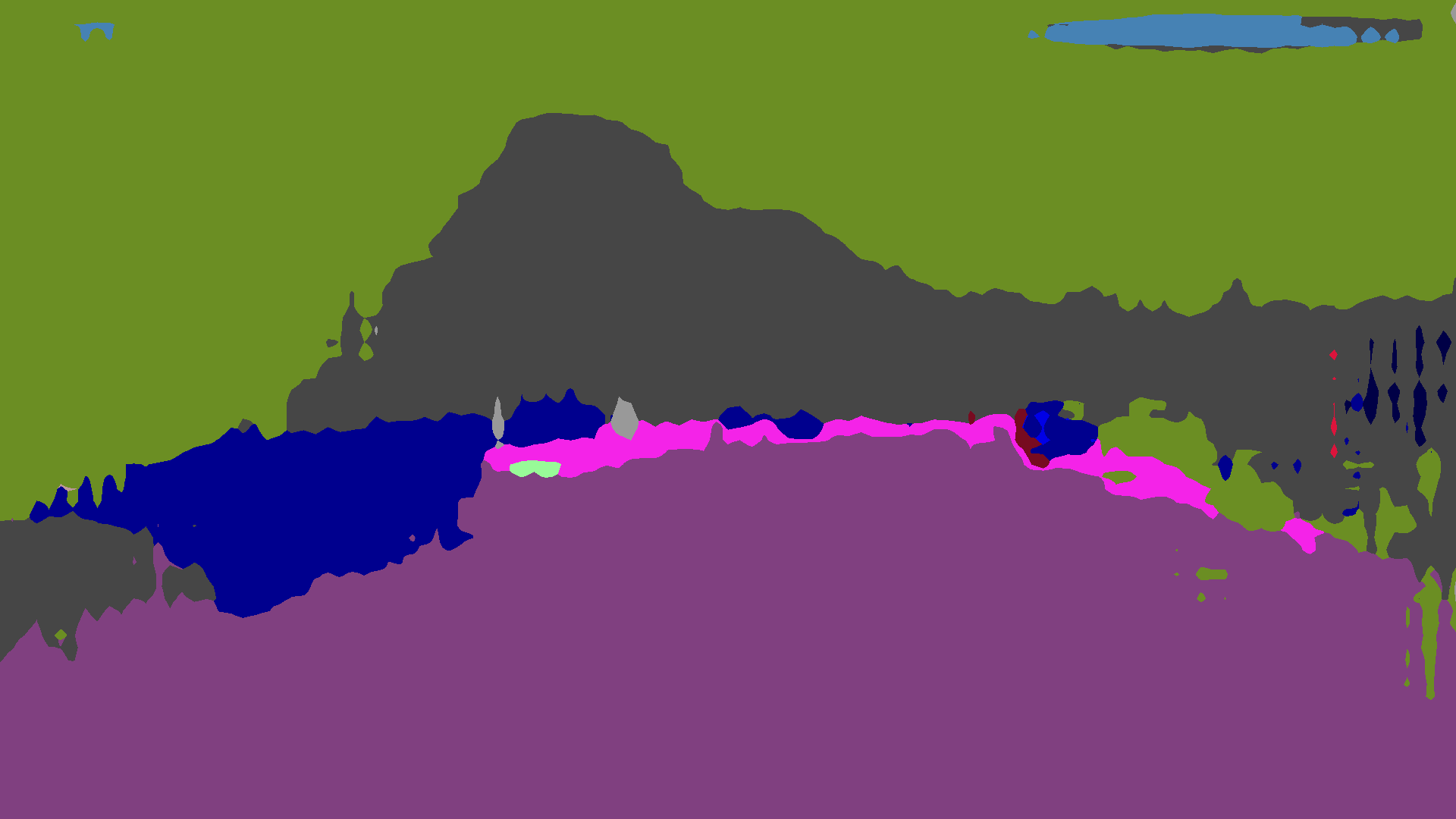}}
    \\
    \vspace{-0.3cm}
    \subfloat{\includegraphics[width=0.195\textwidth]{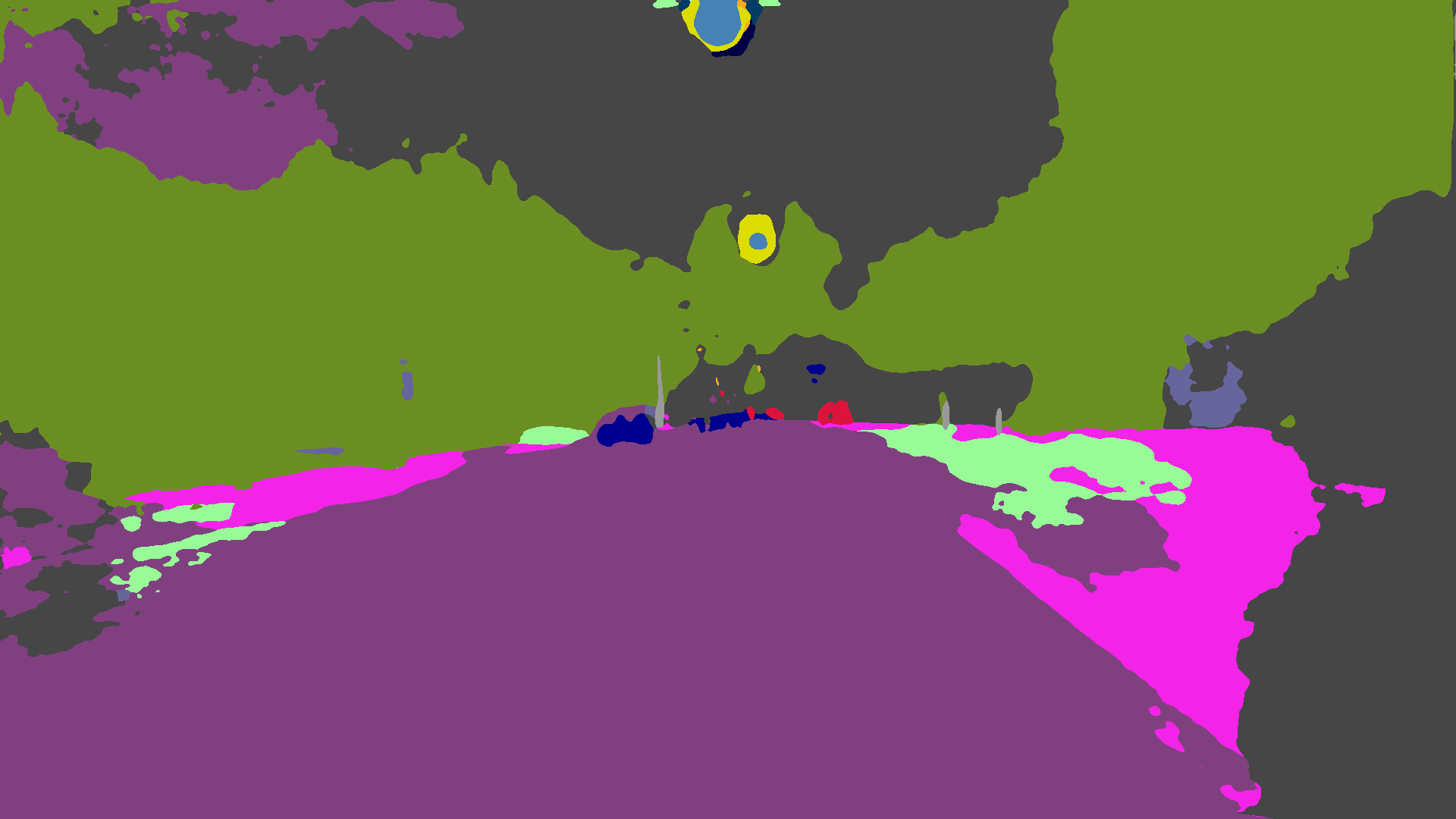}}
    \hfil
    \subfloat{\includegraphics[width=0.195\textwidth]{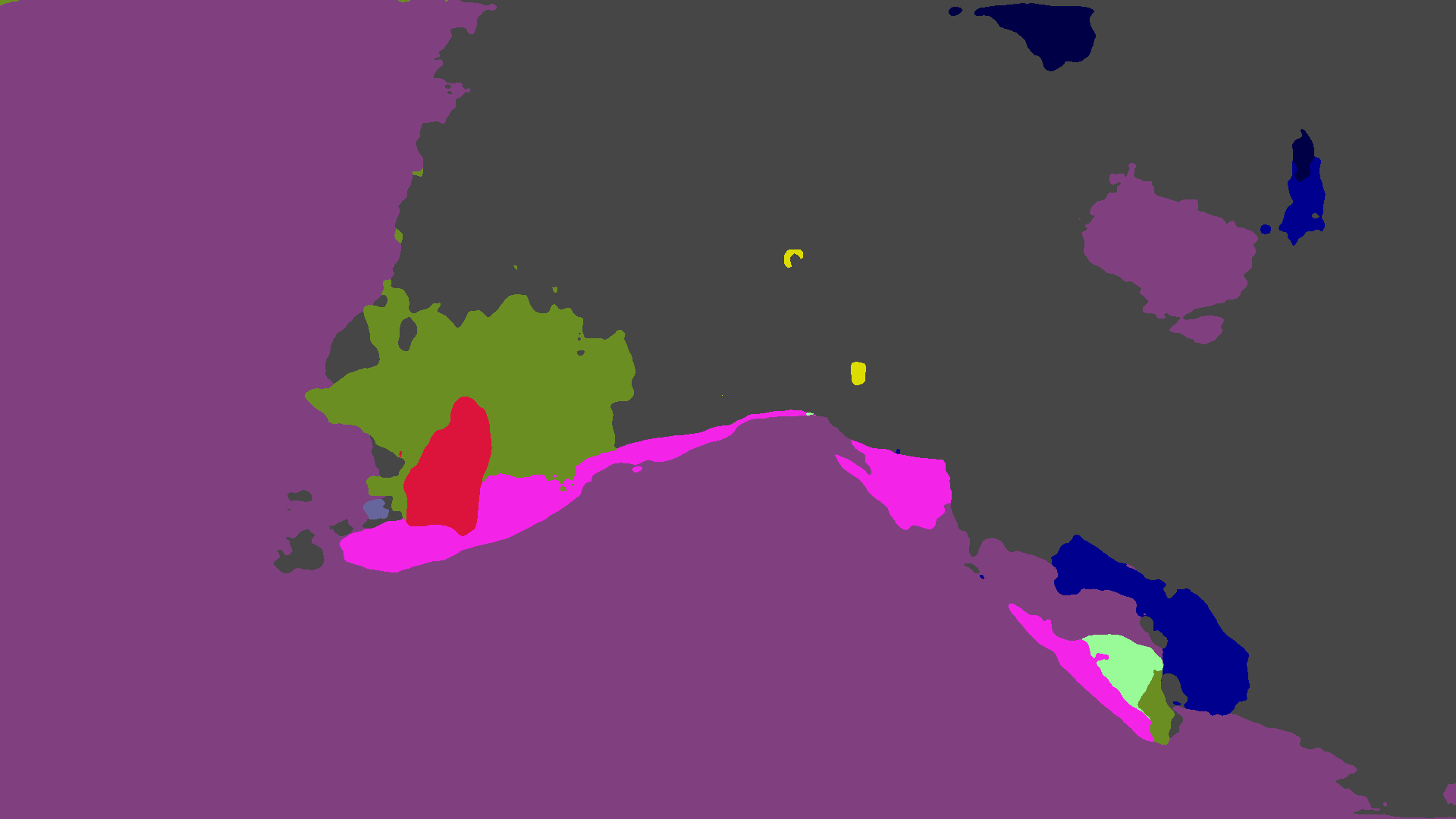}}
    \hfil
    \subfloat{\includegraphics[width=0.195\textwidth]{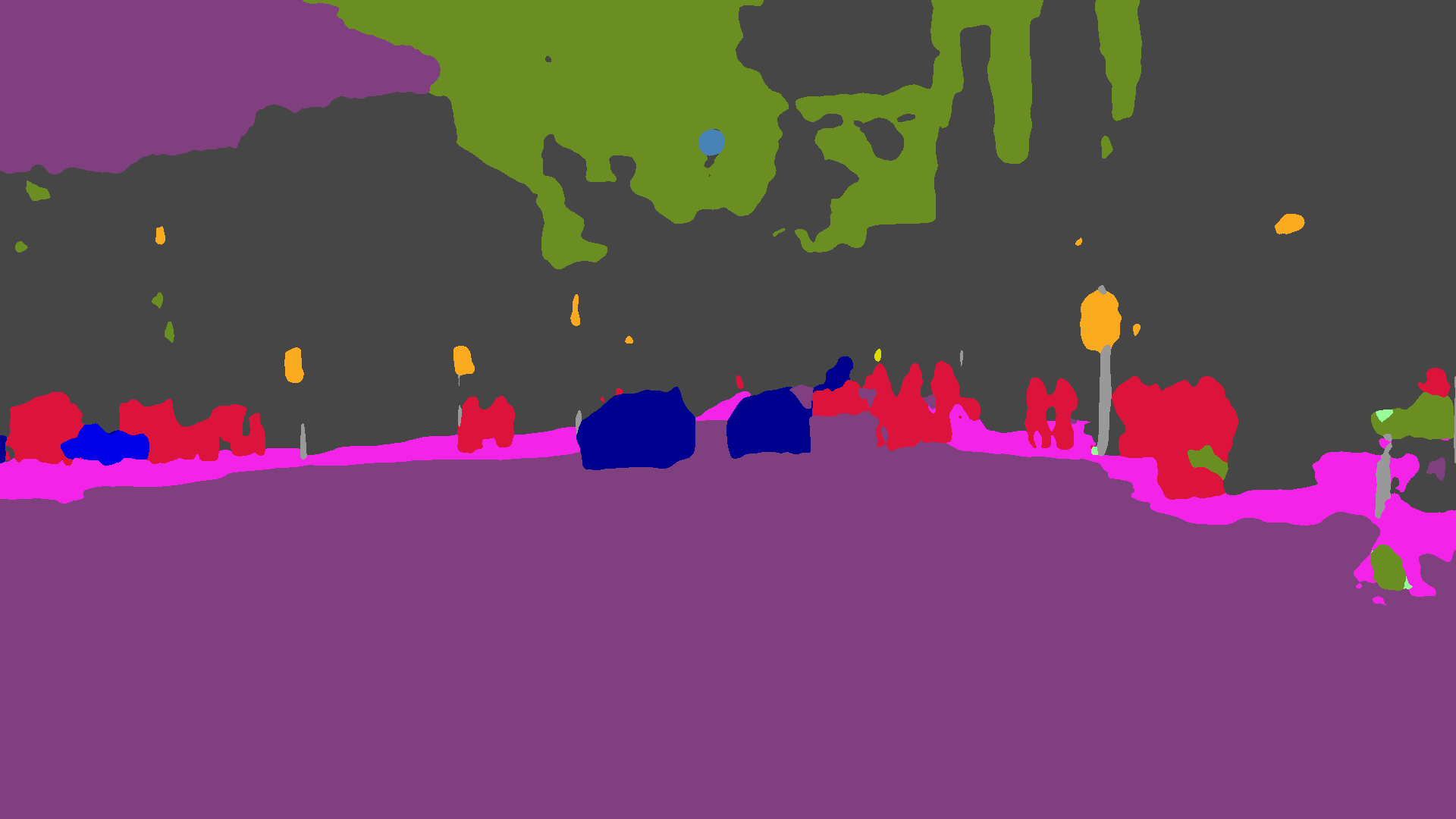}}
    \hfil
    \subfloat{\includegraphics[width=0.195\textwidth]{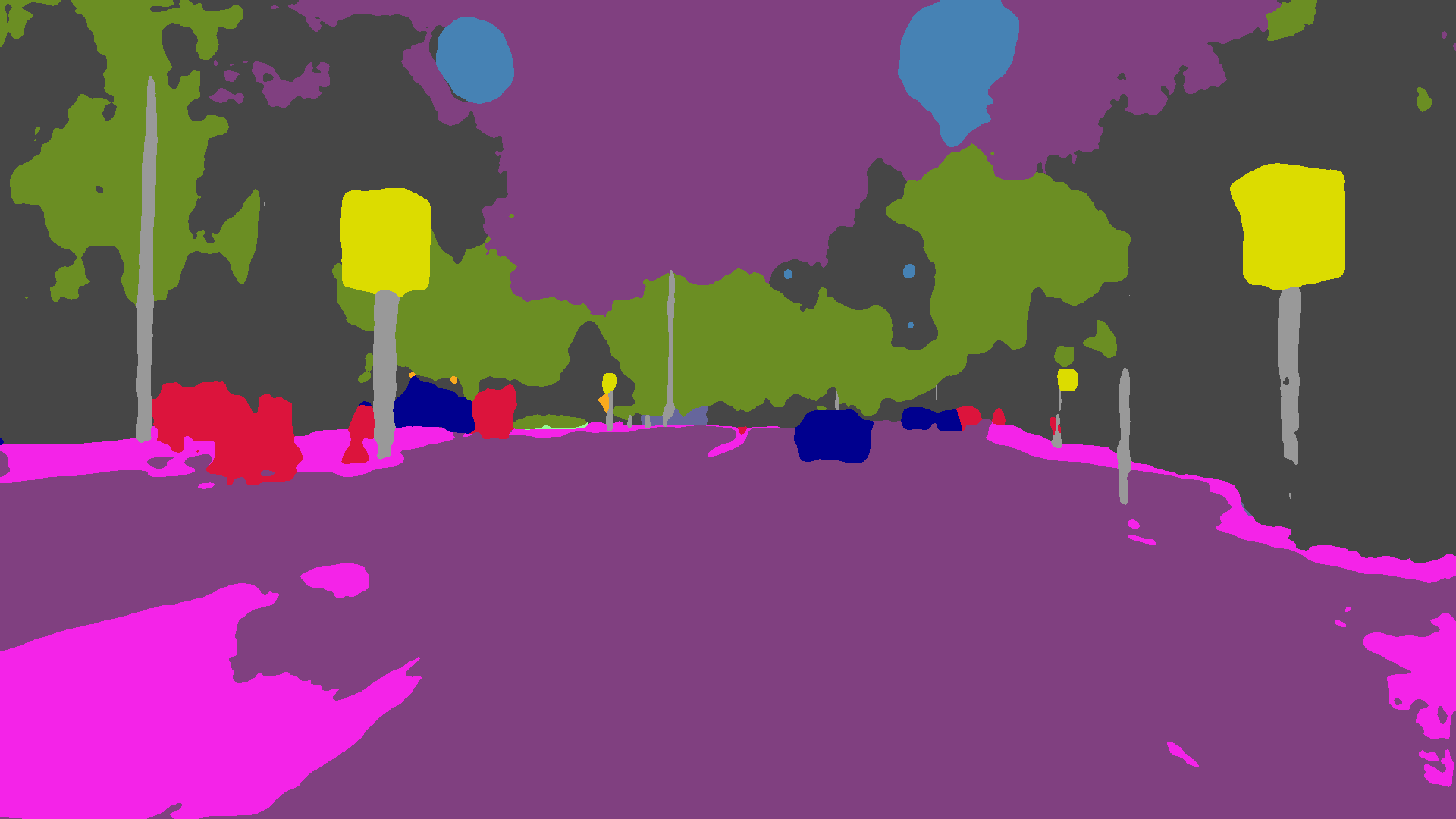}}
    \hfil
    \subfloat{\includegraphics[width=0.195\textwidth]{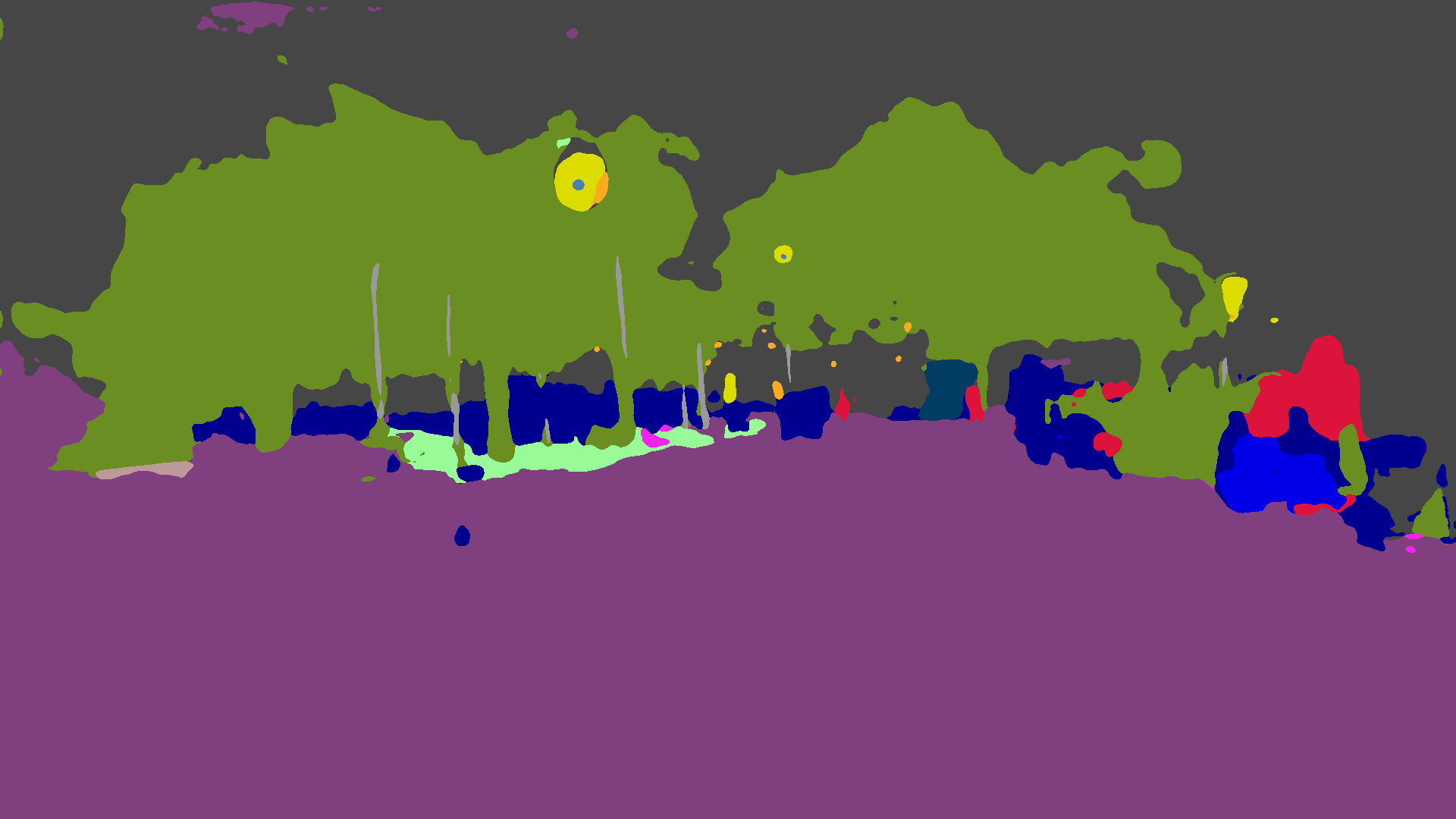}}
    \\
    \vspace{-0.3cm}
    \subfloat{\includegraphics[width=0.195\textwidth]{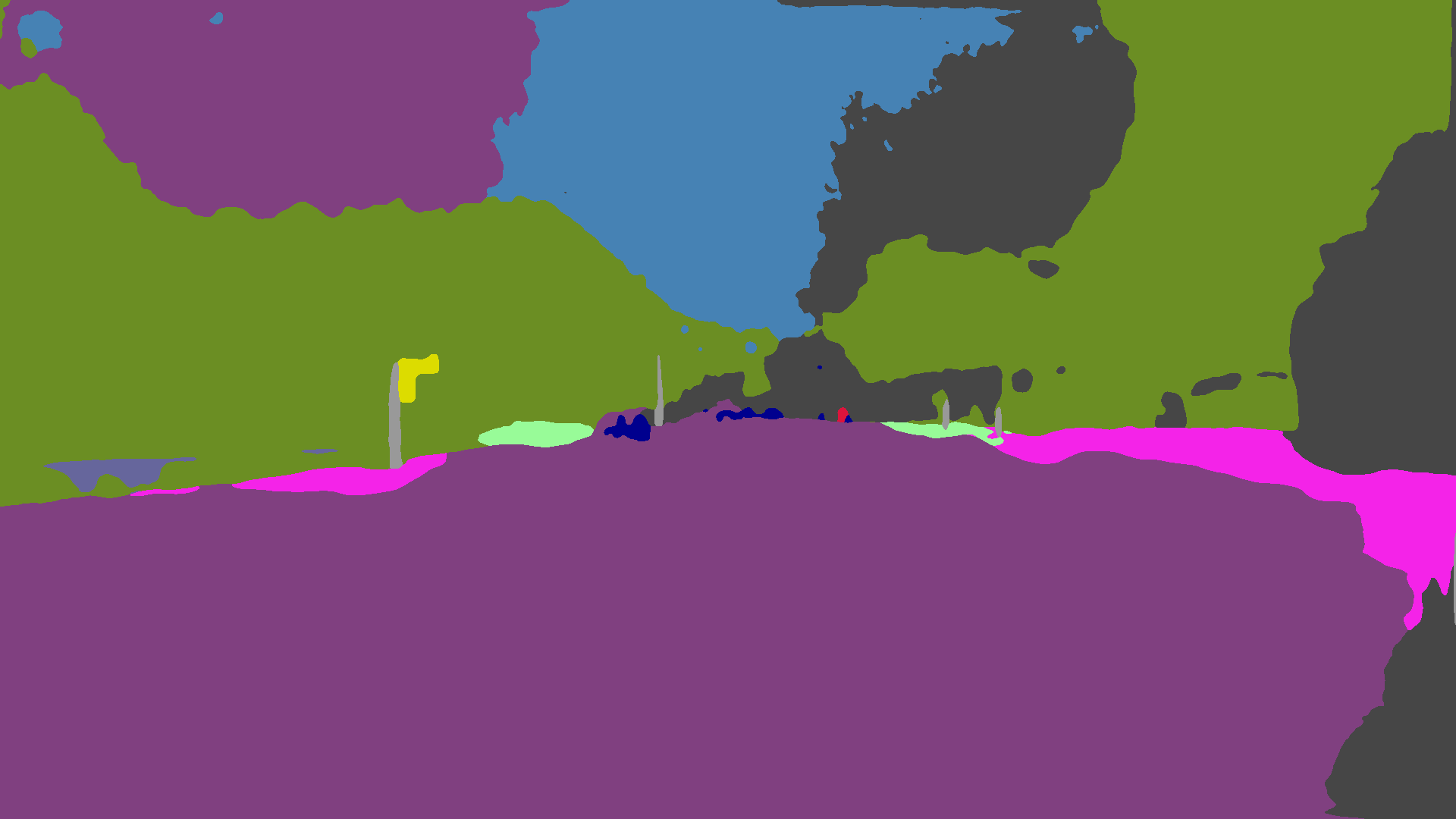}}
    \hfil
    \subfloat{\includegraphics[width=0.195\textwidth]{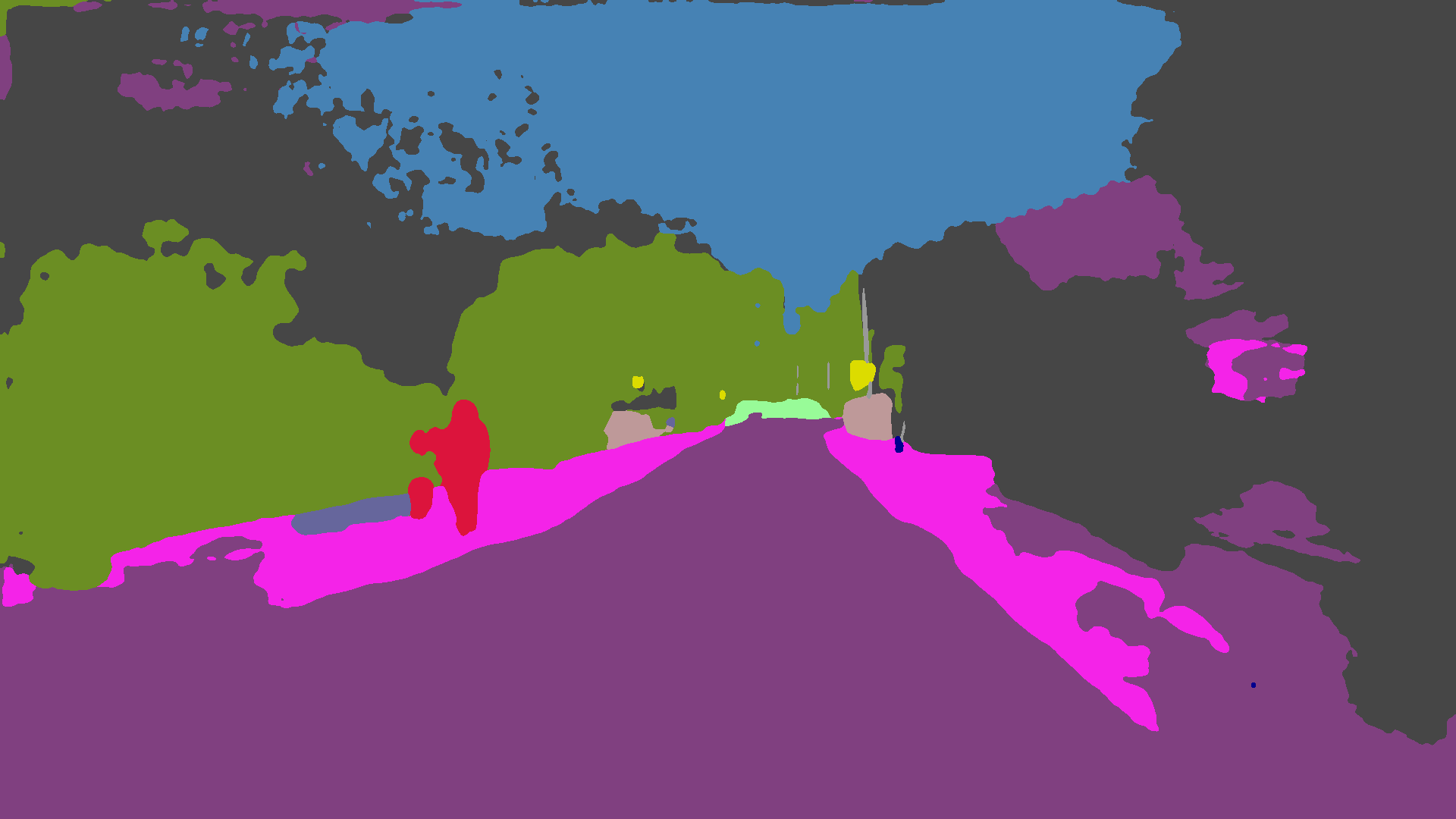}}
    \hfil
    \subfloat{\includegraphics[width=0.195\textwidth]{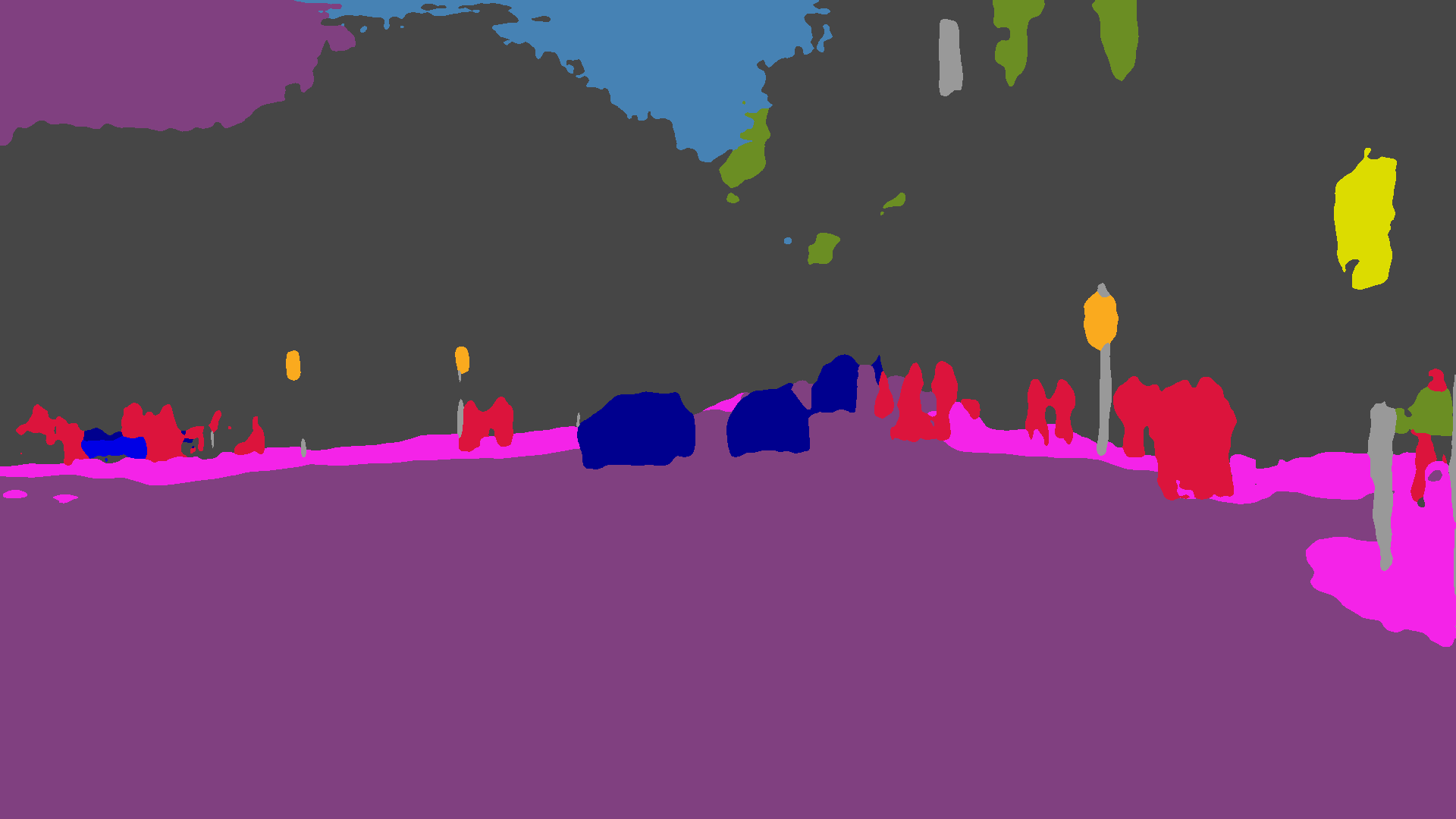}}
    \hfil
    \subfloat{\includegraphics[width=0.195\textwidth]{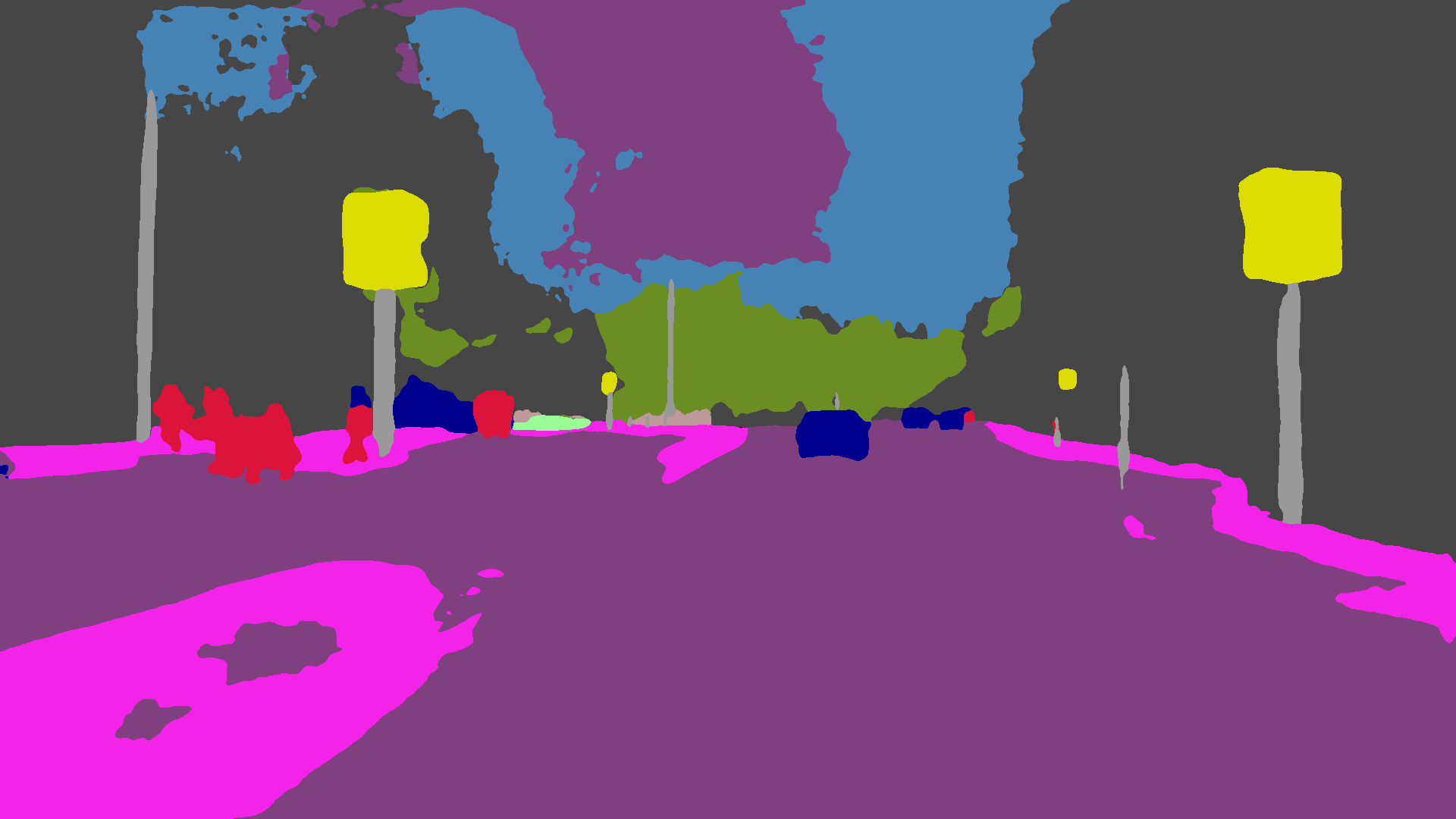}}
    \hfil
    \subfloat{\includegraphics[width=0.195\textwidth]{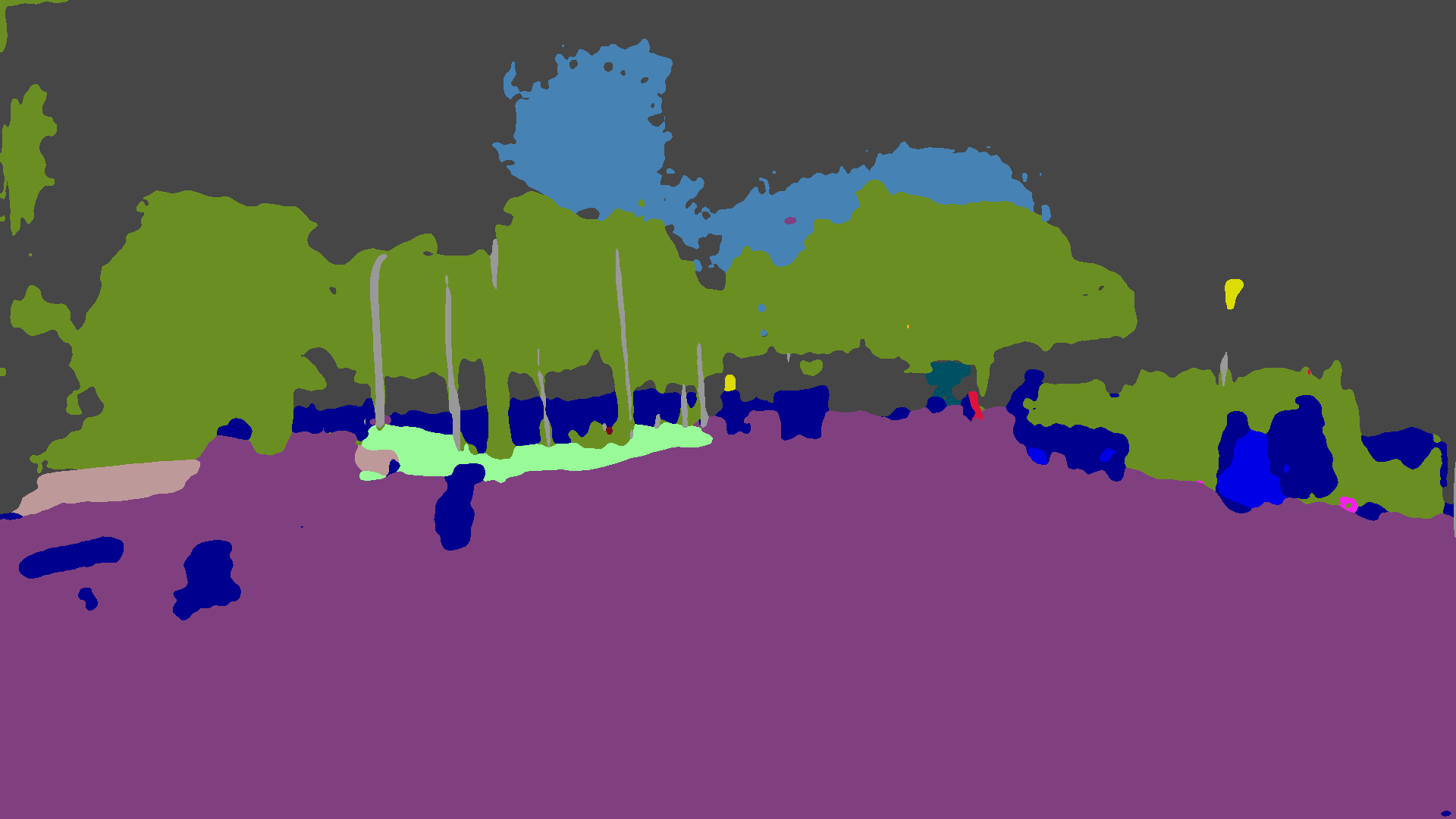}}
    \\
    \vspace{-0.3cm}
    \subfloat{\includegraphics[width=0.195\textwidth]{figures/GP010364_frame_000382_ours_v2.png}}
    \hfil
    \subfloat{\includegraphics[width=0.195\textwidth]{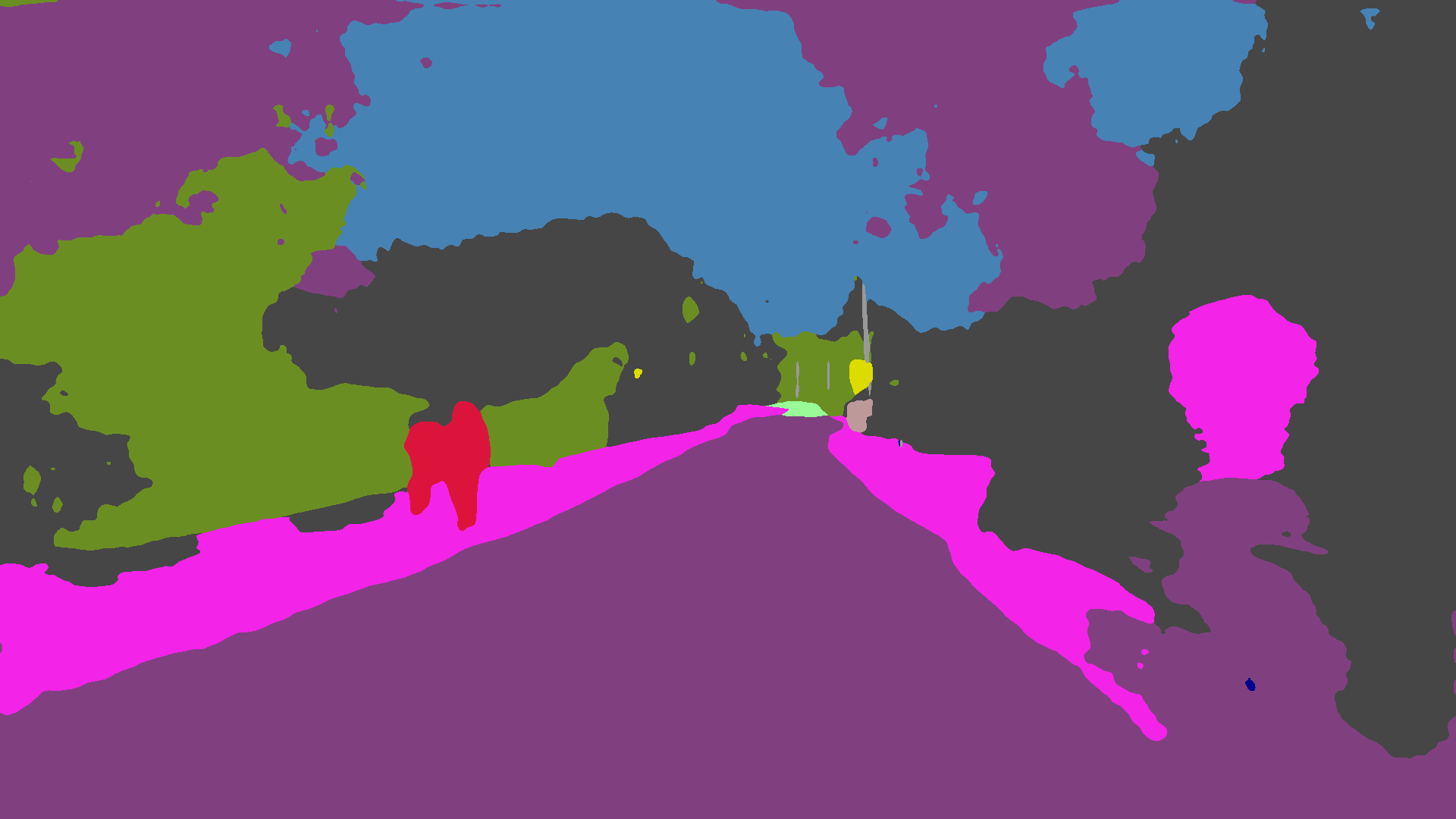}}
    \hfil
    \subfloat{\includegraphics[width=0.195\textwidth]{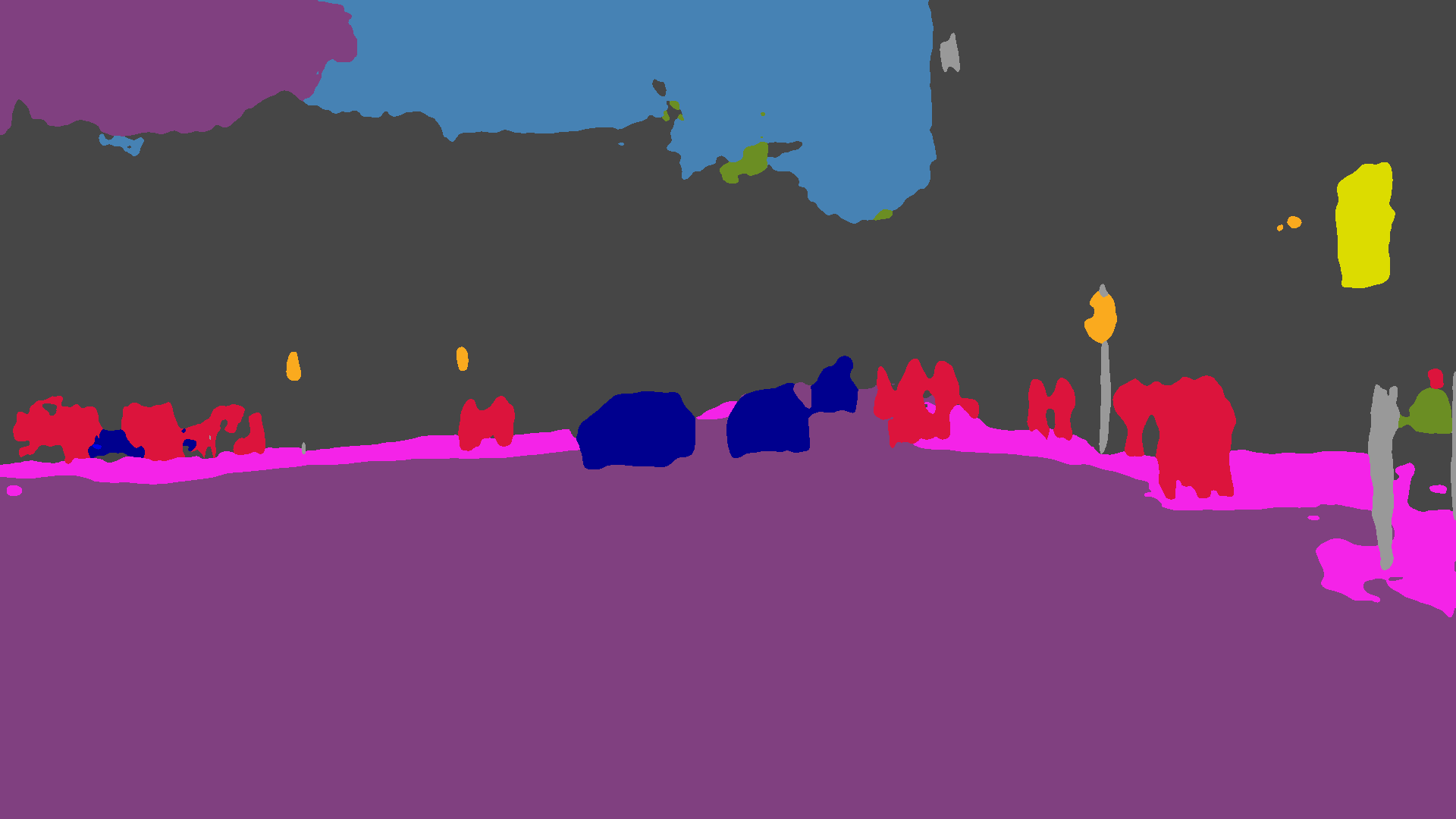}}
    \hfil
    \subfloat{\includegraphics[width=0.195\textwidth]{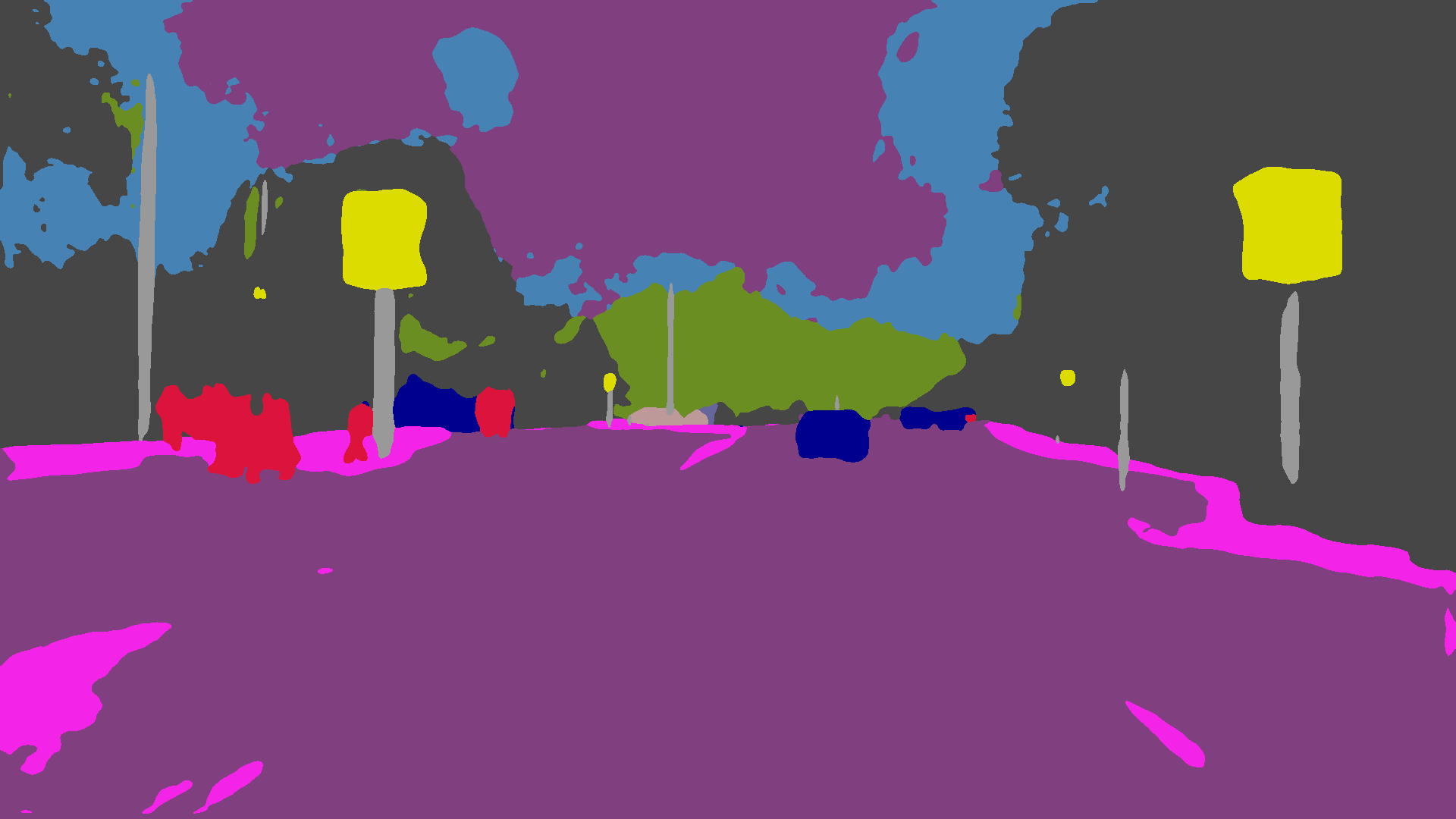}}
    \hfil
    \subfloat{\includegraphics[width=0.195\textwidth]{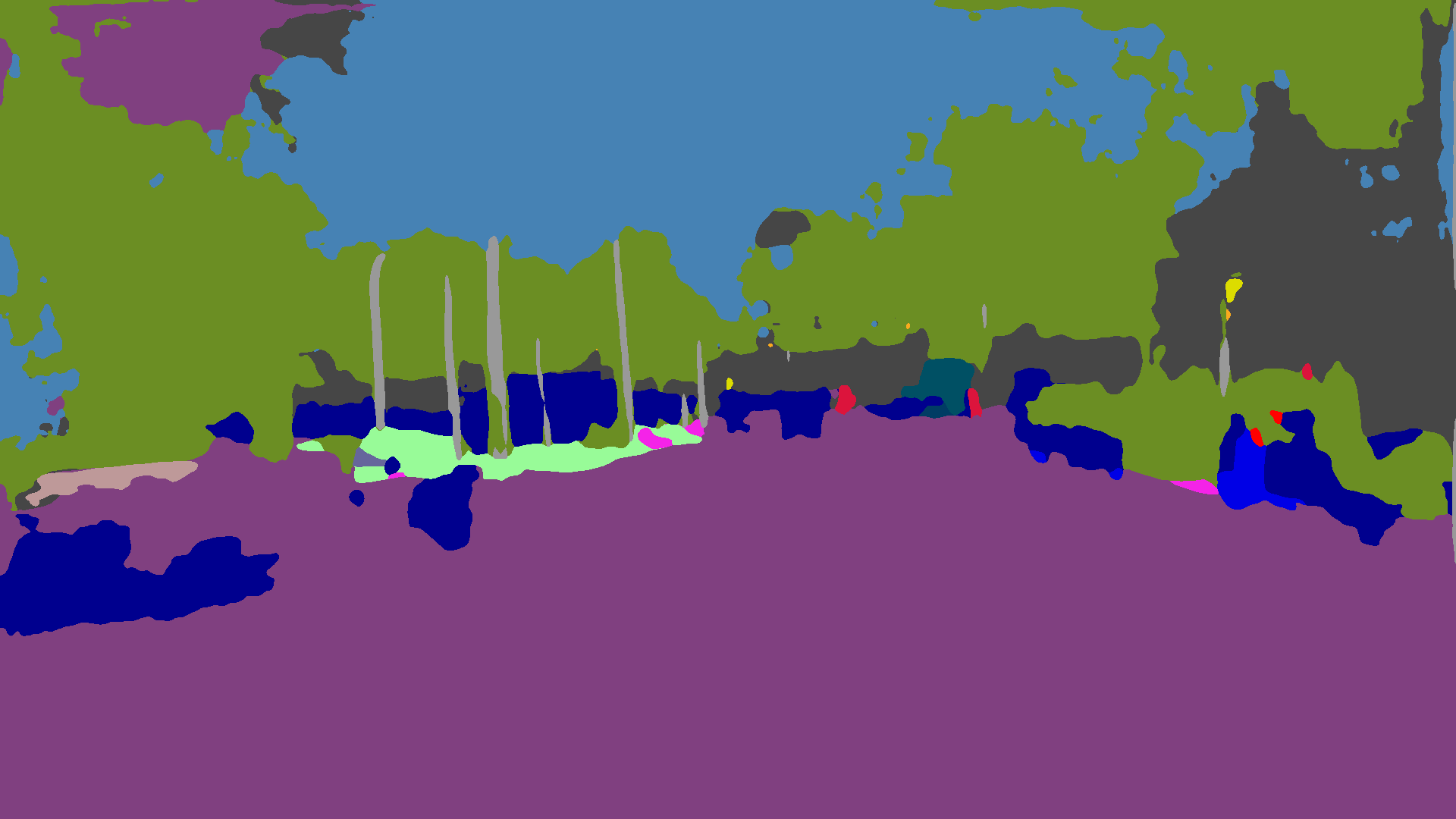}}
    \caption{Examples of our annotations and qualitative semantic segmentation results on \emph{Dark Zurich-test}. From top to bottom row: nighttime image, invalid mask annotation overlaid on the image (valid pixels are colored green), semantic annotation, AdaptSegNet~\cite{adapt:structured:output:cvpr18}, DMAda~\cite{daytime:2:nighttime}, GCMA (ours), and MGCDA (ours).}
    \label{fig:sem:seg:extra}
\end{figure*}

In the first case, the first term in \eqref{eq:proof:nonempty:improved} is strictly positive, so \eqref{eq:proof:true:positive:iou:theta1} implies
\begin{equation} \label{eq:proof:correct:predict:ineq:strict}
|\text{TP}(1/C)| < |\text{TP}(\theta_1)| + |\text{TI}(\theta_1)|.
\end{equation}
We establish the inequality we are after by writing
\begin{align}
&{\text{IoU}} = \nonumber \\
&{=}\;\frac{|\text{TP}(1/C)|}{|\text{TP}(1/C)| + |\text{FN}(1/C)| + |\text{FP}(1/C)|} \nonumber \\
&{=}\;\frac{|\text{TP}(1/C)|}{|\text{TP}(\theta_1)| + |\text{FN}(\theta_1)| + |\text{TI}(\theta_1)| + |\text{FI}(\theta_1)| + |\text{FP}(1/C)|} \nonumber \\
&{\leq}\;\frac{|\text{TP}(1/C)|}{|\text{TP}(\theta_1)| + |\text{TI}(\theta_1)| + |\text{FP}(\theta_1)| + |\text{FN}(\theta_1)| + |\text{FI}(\theta_1)|} \nonumber \\
&{<}\;\frac{|\text{TP}(\theta_1)| + |\text{TI}(\theta_1)|}{|\text{TP}(\theta_1)| + |\text{TI}(\theta_1)| + |\text{FP}(\theta_1)| + |\text{FN}(\theta_1)| + |\text{FI}(\theta_1)|} \nonumber \\
&{=}\;\text{UIoU}(\theta_1),
\end{align}
where we have used the definition of IoU in the second line, \eqref{eq:proof:true:positive:false:negative:iou} in the third line, $\text{FP}(\theta_1) \subseteq \text{FP}(1/C)$ in the fourth line, \eqref{eq:proof:correct:predict:ineq:strict} in the fifth line, and the definition of UIoU that has been introduced in \eqref{eq:uiou} in the last line.

In the second case, the second term in \eqref{eq:proof:nonempty:improved} is strictly positive, which implies that
\begin{equation} \label{eq:proof:fp:ineq}
|\text{FP}(1/C)| > |\text{FP}(\theta_1)|.
\end{equation}
Besides, applying the nonnegativity of the first term in \eqref{eq:proof:nonempty:improved} to \eqref{eq:proof:true:positive:iou:theta1} leads to
\begin{equation} \label{eq:proof:correct:predict:ineq}
|\text{TP}(1/C)| \leq |\text{TP}(\theta_1)| + |\text{TI}(\theta_1)|.
\end{equation}
Similarly to the first case, we establish the inequality we are after by writing
\begin{align}
&{\text{IoU}} = \nonumber \\
&{=}\;\frac{|\text{TP}(1/C)|}{|\text{TP}(\theta_1)| + |\text{TI}(\theta_1)| + |\text{FP}(1/C)| + |\text{FN}(\theta_1)| + |\text{FI}(\theta_1)|} \nonumber \\
&{<}\;\frac{|\text{TP}(1/C)|}{|\text{TP}(\theta_1)| + |\text{TI}(\theta_1)| + |\text{FP}(\theta_1)| + |\text{FN}(\theta_1)| + |\text{FI}(\theta_1)|} \nonumber \\
&{\leq}\;\frac{|\text{TP}(\theta_1)| + |\text{TI}(\theta_1)|}{|\text{TP}(\theta_1)| + |\text{TI}(\theta_1)| + |\text{FP}(\theta_1)| + |\text{FN}(\theta_1)| + |\text{FI}(\theta_1)|} \nonumber \\
&{=}\;\text{UIoU}(\theta_1),
\end{align}
where we have used the definition of IoU as well as \eqref{eq:proof:true:positive:false:negative:iou} in the second line, \eqref{eq:proof:fp:ineq} in the third line, \eqref{eq:proof:correct:predict:ineq} in the fourth line, and the definition of UIoU in the last line.
\end{proof}

\section{Additional Qualitative Results}
\label{supp:sec:results}

In Fig.~\ref{fig:sem:seg:extra}, we compare our MGCDA approach against our original GCMA approach, AdaptSegNet~\cite{adapt:structured:output:cvpr18} and DMAda~\cite{daytime:2:nighttime} on additional images from \emph{Dark Zurich-test}, further demonstrating the superiority of MGCDA. For these images, we also present our annotations for invalid masks and semantic labels, which show that a significant portion of ground-truth invalid regions is indeed assigned a reliable semantic label through our annotation protocol and can thus be included in the evaluation.

\section{Parameter Selection for Prediction Fusion}
\label{supp:sec:parameters}

\begin{figure*}
    \centering
    \subfloat[Dark image $I^z$]{\includegraphics[width=0.24\textwidth]{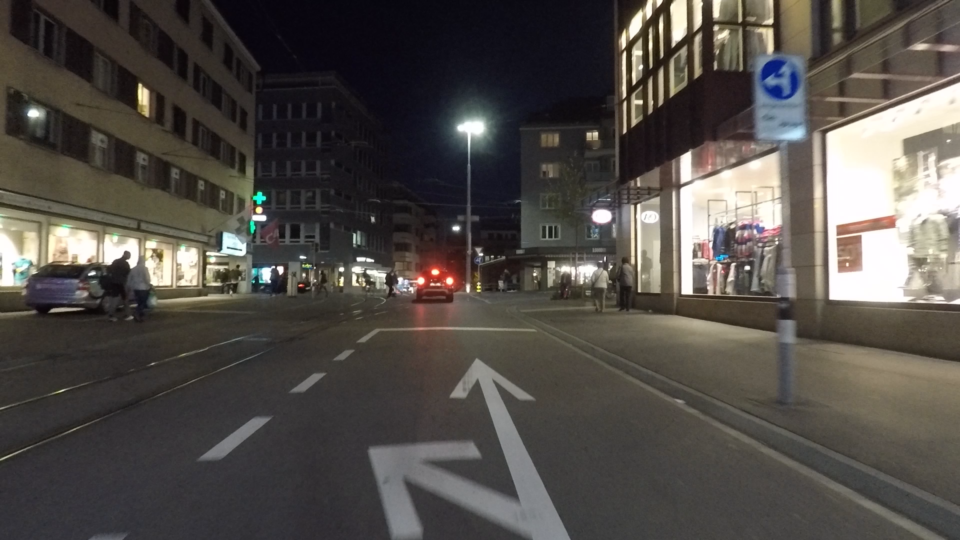}\label{fig:prediction:fusion:parameters:twilight}}
    \hfil
    \subfloat[Initial prediction $\mathbf{S}^z$ for $I^z$]{\includegraphics[width=0.24\textwidth]{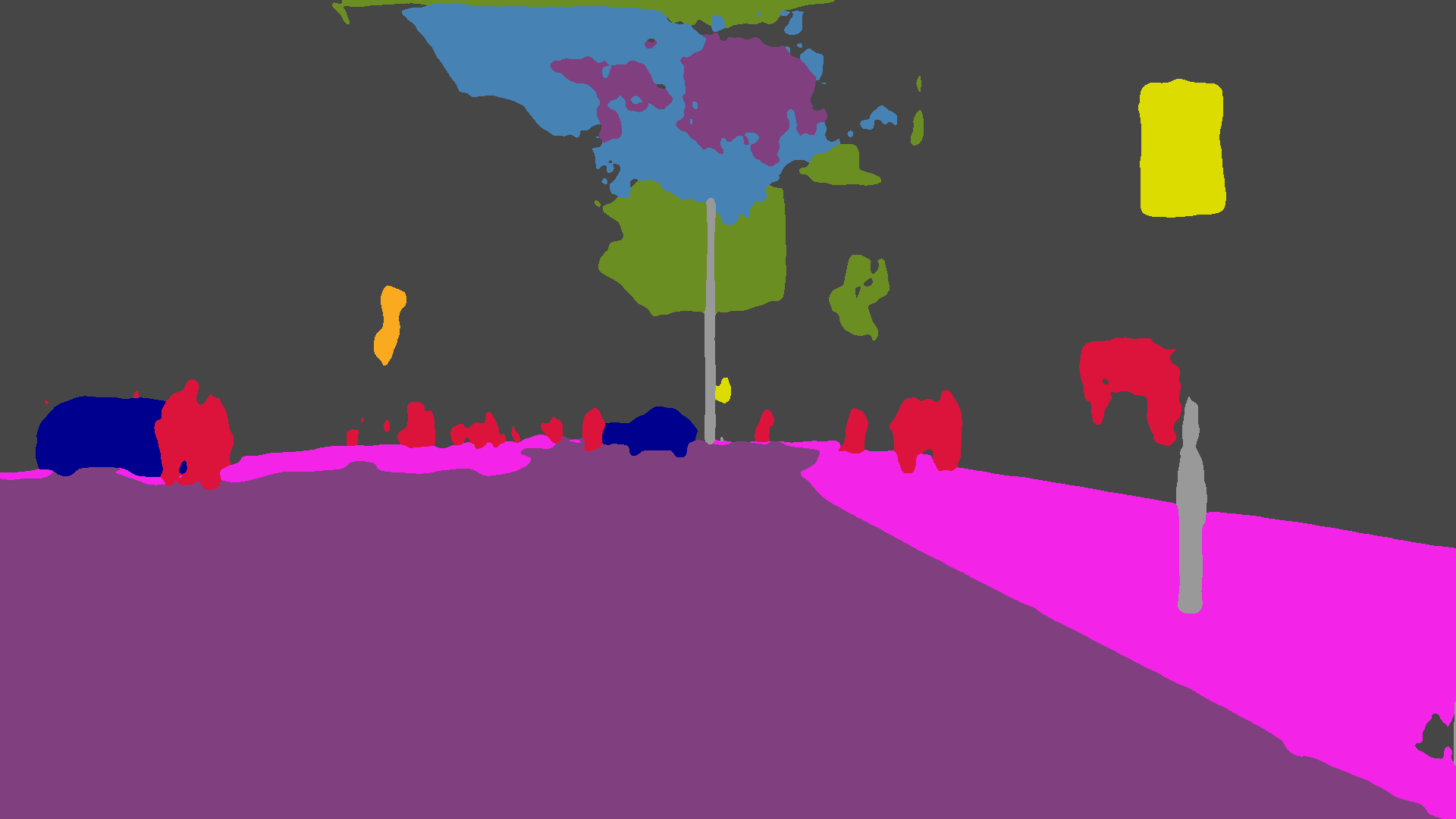}\label{fig:prediction:fusion:parameters:twilight:init}}
    \hfil
    \subfloat[Daytime image $I^1$]{\includegraphics[width=0.24\textwidth]{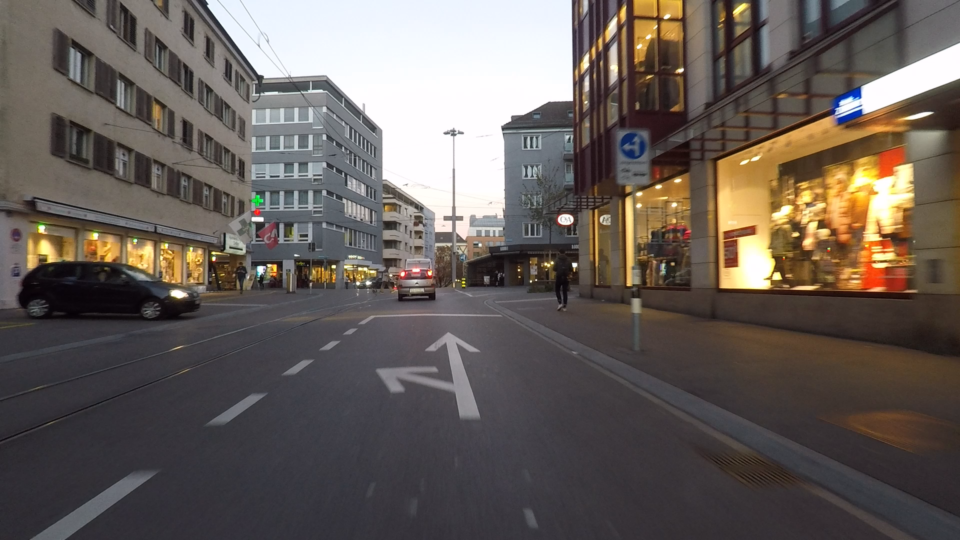}\label{fig:prediction:fusion:parameters:day}}
    \hfil
    \subfloat[Initial prediction $\mathbf{S}^1$ for $I^1$]{\includegraphics[width=0.24\textwidth]{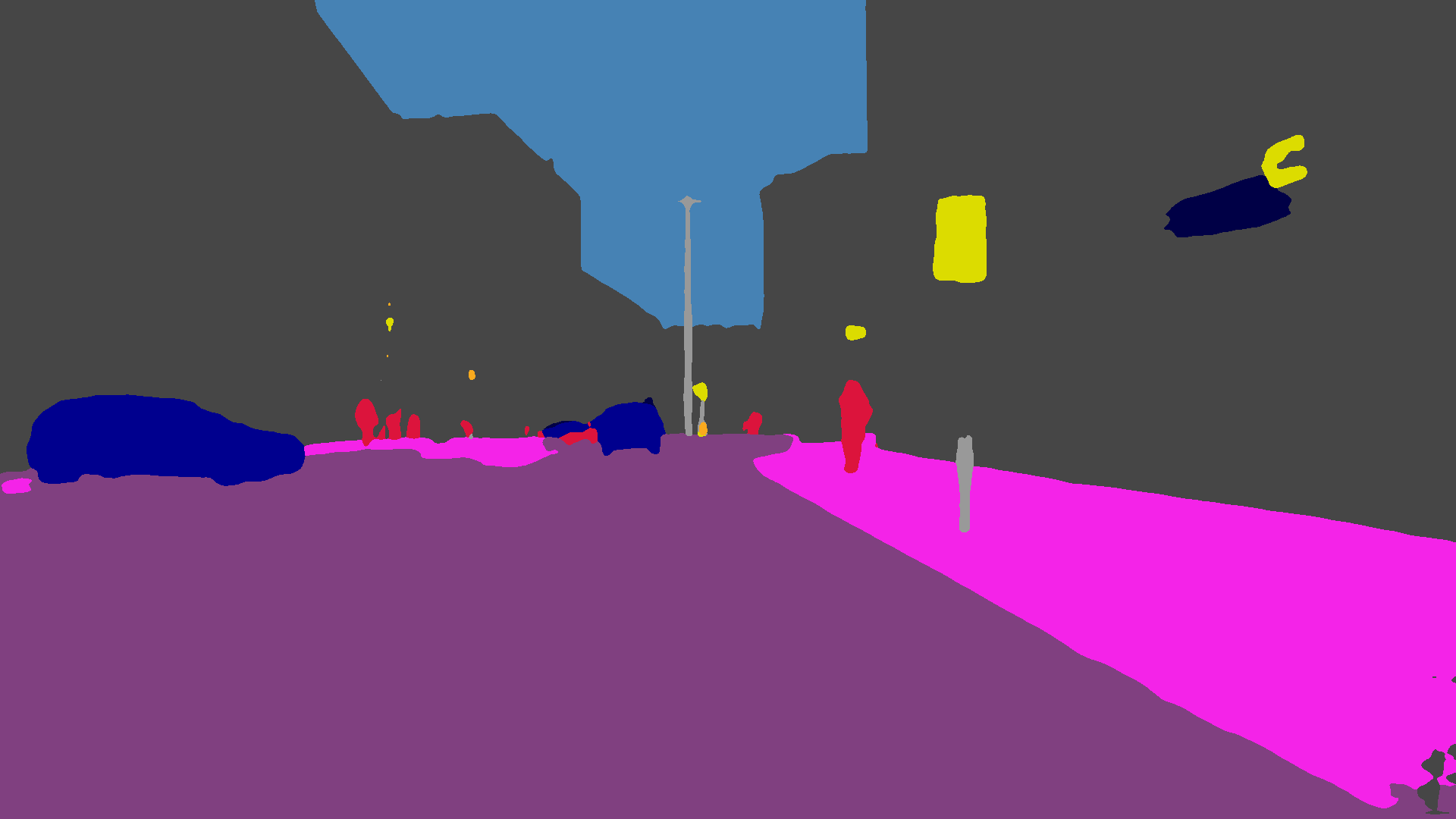}\label{fig:prediction:fusion:parameters:day:init}}
    \\
    \subfloat[Aligned prediction $\tilde{\mathbf{S}}^{1}$ for $I^1$ with depth-based warping]{\includegraphics[width=0.24\textwidth]{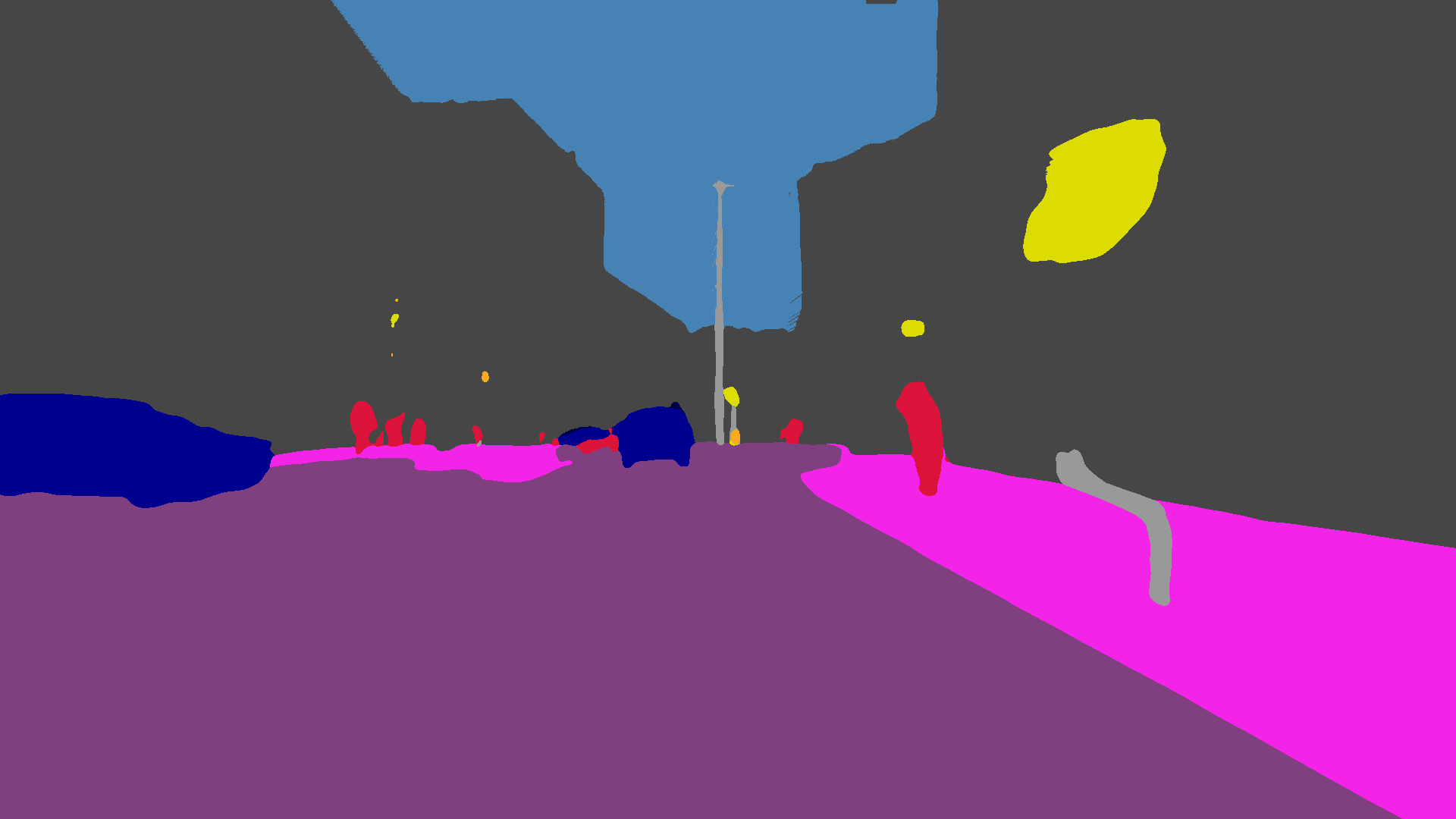}\label{fig:prediction:fusion:parameters:day:warping}}
    \hfil
    \subfloat[Refined prediction $\hat{\mathbf{S}}^{z}$ for $I^z$ with $\alpha_h = 0.3$ ]{\includegraphics[width=0.24\textwidth]{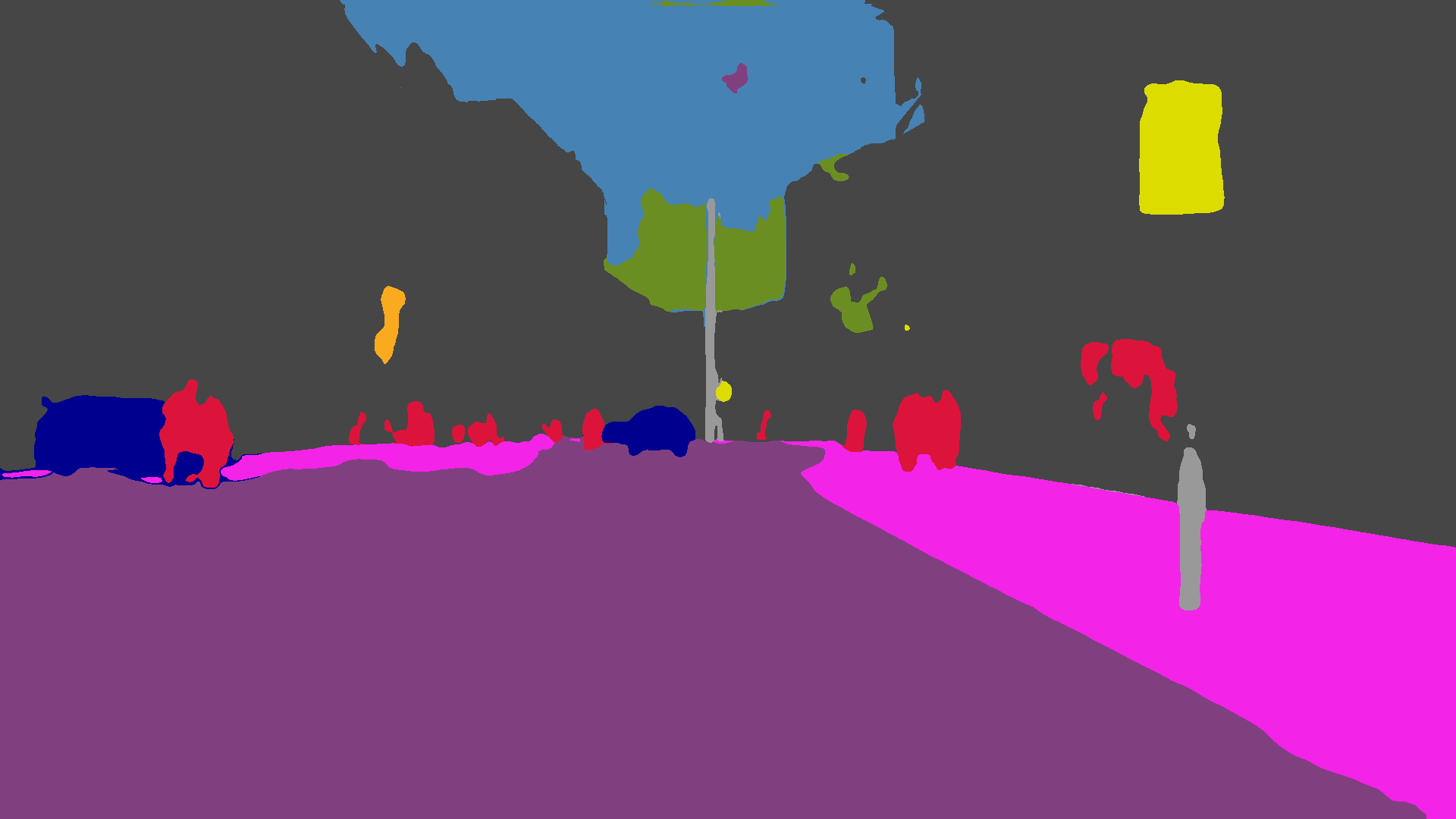}\label{fig:prediction:fusion:parameters:twilight:warping:aH:03}}
    \hfil
    \subfloat[Refined prediction $\hat{\mathbf{S}}^{z}$ for $I^z$ with $\alpha_h = 0.6$]{\includegraphics[width=0.24\textwidth]{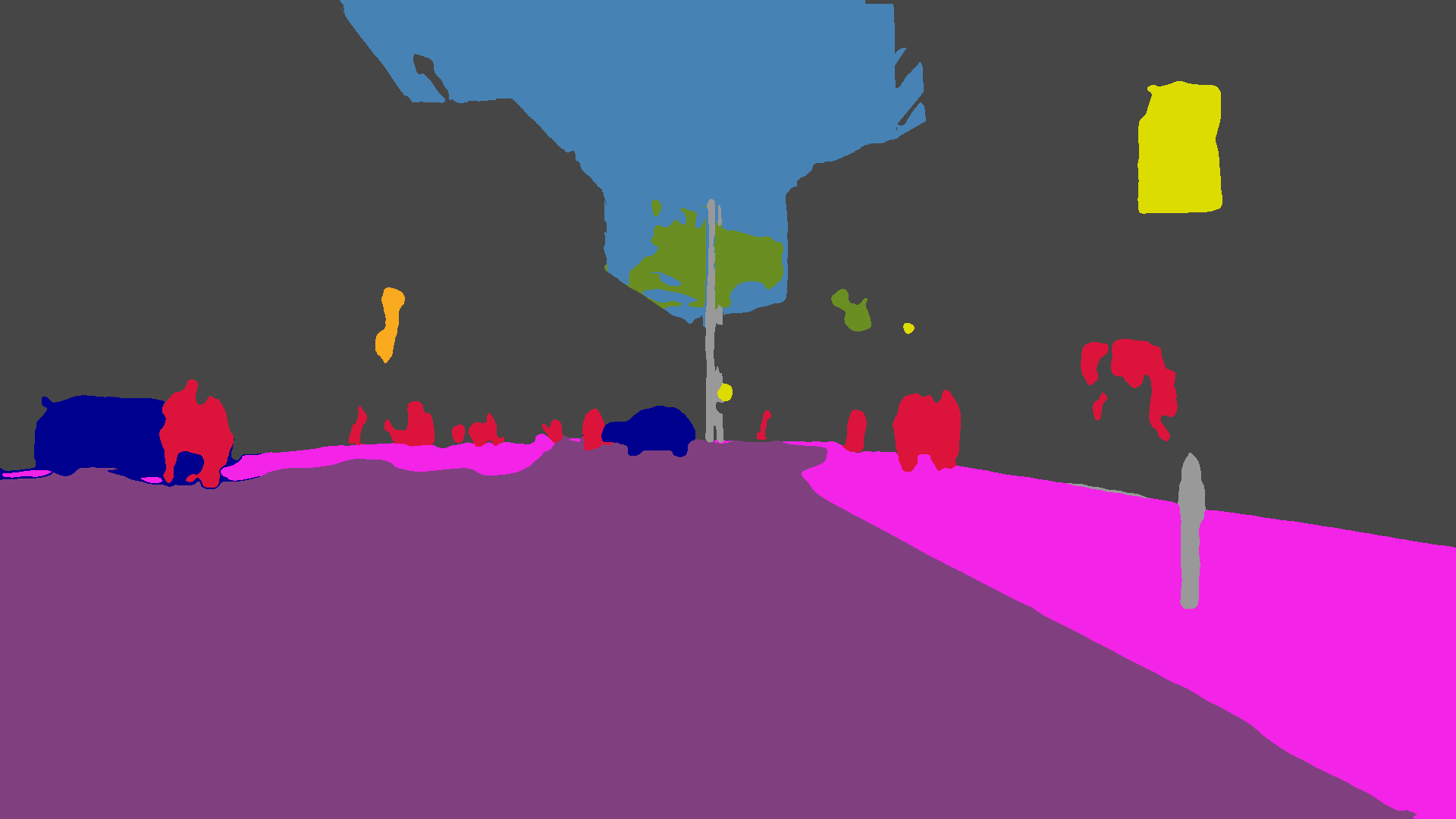}\label{fig:prediction:fusion:parameters:twilight:warping:aH:06}}
    \hfil
    \subfloat[Refined prediction $\hat{\mathbf{S}}^{z}$ for $I^z$ with $\alpha_h = 1$]{\includegraphics[width=0.24\textwidth]{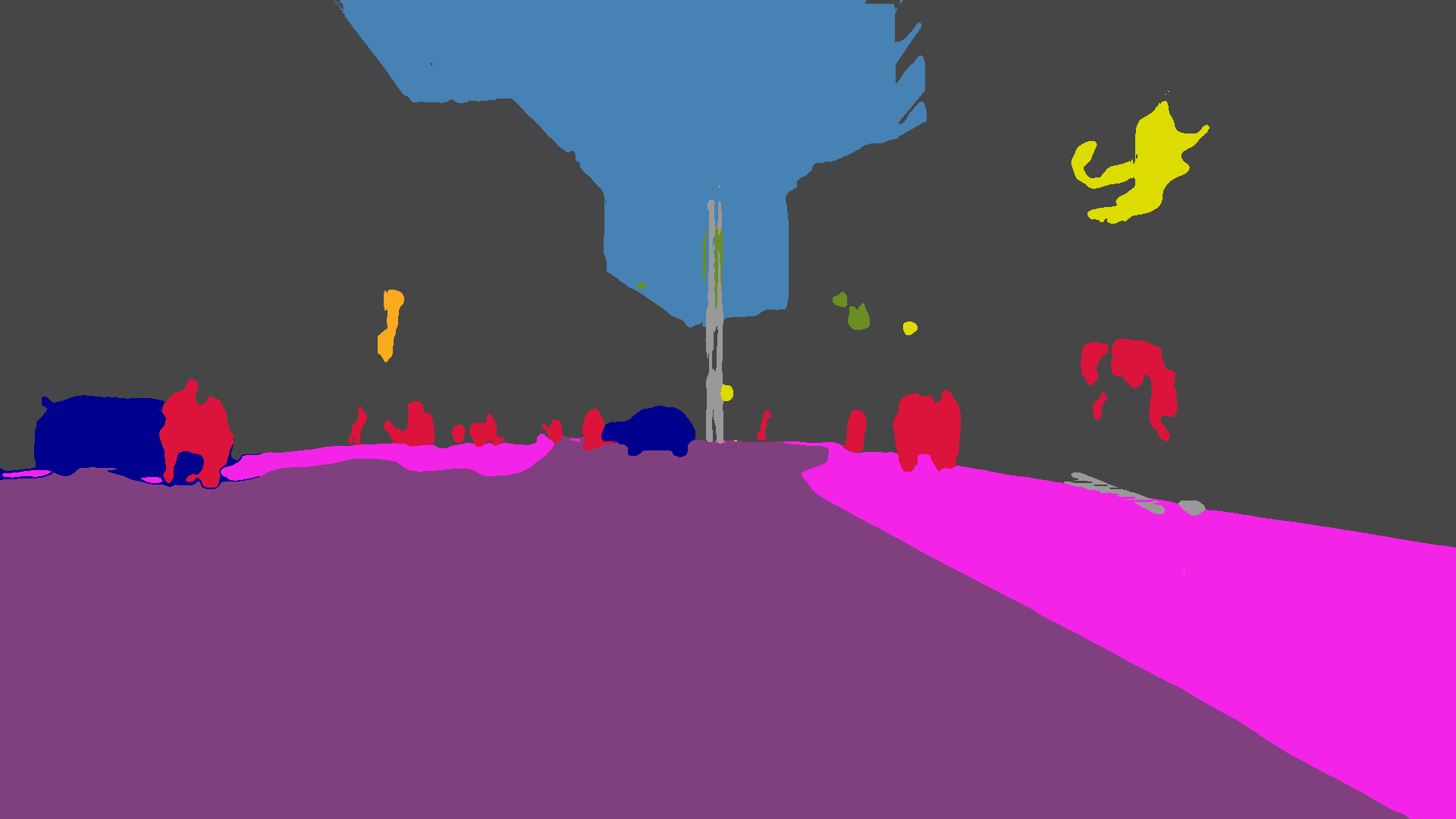}\label{fig:prediction:fusion:parameters:twilight:warping:aH:1}}
    \caption{Comparison of confidence-adaptive prediction fusion results for different values of $\alpha_h$ on an example image pair from \emph{Dark Zurich}. The other two parameters of prediction fusion are fixed to their default values $\alpha_l = 0.3$ and $\eta = 0.2$ across all compared cases.}
    \label{fig:prediction:fusion:parameters}
\end{figure*}

In Fig.~\ref{fig:prediction:fusion:parameters}, we illustrate the effect of parameter $\alpha_h$ of our confidence-adaptive prediction fusion presented in Sec.~\ref{sec:mgcda:guidance:fusion}. Fig.~\ref{fig:prediction:fusion:parameters}\subref{fig:prediction:fusion:parameters:twilight:warping:aH:03}--\subref{fig:prediction:fusion:parameters:twilight:warping:aH:1} show the refined predictions after fusion for increasing values of $\alpha_h$. Setting $\alpha_h$ to a low value in Fig.~\ref{fig:prediction:fusion:parameters}\subref{fig:prediction:fusion:parameters:twilight:warping:aH:03} prevents the refinement of some erroneous parts of the initial prediction for the dark image, such as the region in the central part of the image which is incorrectly labeled as \emph{vegetation}. Increasing $\alpha_h$ to $0.6$ in Fig.~\ref{fig:prediction:fusion:parameters}\subref{fig:prediction:fusion:parameters:twilight:warping:aH:06} corrects a large part of this region to \emph{sky}, as the aligned daytime prediction (Fig.~\ref{fig:prediction:fusion:parameters}\subref{fig:prediction:fusion:parameters:day:warping}), which contains the correct label \emph{sky} in this region, is assigned a higher weight in the prediction fusion. However, when $\alpha_h$ is further increased to $1$ in Fig.~\ref{fig:prediction:fusion:parameters}\subref{fig:prediction:fusion:parameters:twilight:warping:aH:1}, the initially correctly segmented traffic sign and pole in the right part of the image have their labels aggravated due to the overly high weight assigned to the daytime prediction, which is imperfectly aligned to the dark image for these particular objects (cf.\ Fig.~\ref{fig:prediction:fusion:parameters}\subref{fig:prediction:fusion:parameters:day:warping}). This demonstrates the trade-off in the selection of $\alpha_h$, which requires striking a balance between exploiting the daytime prediction, as it is performed in an easier domain, and not relying on it too much, due to its imperfect alignment to the dark image. We expect that improving the prediction alignment, primarily via increasing the accuracy of the depth map for the daytime view, will enable a better exploitation of the daytime prediction in our geometrically guided segmentation refinement by placing a higher weight on it in the prediction fusion step.

\ifCLASSOPTIONcompsoc
  \section*{Acknowledgments}
\else
  \section*{Acknowledgment}
\fi

This work is funded by Toyota Motor Europe via the research project TRACE-Z\"urich. We thank Simon Hecker for his advice on decoding GoPro GPS data.



\bibliographystyle{IEEEtran}
\bibliography{IEEEabrv,refs}

\begin{thebibliography}{10}
\providecommand{\url}[1]{#1}
\csname url@samestyle\endcsname
\providecommand{\newblock}{\relax}
\providecommand{\bibinfo}[2]{#2}
\providecommand{\BIBentrySTDinterwordspacing}{\spaceskip=0pt\relax}
\providecommand{\BIBentryALTinterwordstretchfactor}{4}
\providecommand{\BIBentryALTinterwordspacing}{\spaceskip=\fontdimen2\font plus
\BIBentryALTinterwordstretchfactor\fontdimen3\font minus
  \fontdimen4\font\relax}
\providecommand{\BIBforeignlanguage}[2]{{%
\expandafter\ifx\csname l@#1\endcsname\relax
\typeout{** WARNING: IEEEtran.bst: No hyphenation pattern has been}%
\typeout{** loaded for the language `#1'. Using the pattern for}%
\typeout{** the default language instead.}%
\else
\language=\csname l@#1\endcsname
\fi
#2}}
\providecommand{\BIBdecl}{\relax}
\BIBdecl

\bibitem{vision:atmosphere}
S.~G. Narasimhan and S.~K. Nayar, ``Vision and the atmosphere,''
  \emph{International Journal of Computer Vision}, vol.~48, no.~3, pp.
  233--254, 2002.

\bibitem{dehazing:LAPNet}
Y.~Li, Q.~Miao, W.~Ouyang, Z.~Ma, H.~Fang, C.~Dong, and Y.~Quan, ``{LAP-Net}:
  Level-aware progressive network for image dehazing,'' in \emph{The IEEE
  International Conference on Computer Vision (ICCV)}, 2019.

\bibitem{pspnet}
H.~Zhao, J.~Shi, X.~Qi, X.~Wang, and J.~Jia, ``Pyramid scene parsing network,''
  in \emph{The IEEE Conference on Computer Vision and Pattern Recognition
  (CVPR)}, 2017.

\bibitem{refinenet}
G.~Lin, A.~Milan, C.~Shen, and I.~Reid, ``{RefineNet}: Multi-path refinement
  networks with identity mappings for high-resolution semantic segmentation,''
  in \emph{IEEE Conference on Computer Vision and Pattern Recognition (CVPR)},
  2017.

\bibitem{dilated:convolution}
F.~Yu and V.~Koltun, ``Multi-scale context aggregation by dilated
  convolutions,'' in \emph{International Conference on Learning
  Representations}, 2016.

\bibitem{pascal:2011}
M.~Everingham, L.~Van~Gool, C.~K. Williams, J.~Winn, and A.~Zisserman, ``The
  {PASCAL} visual object classes ({VOC}) challenge,'' \emph{International
  Journal of Computer Vision}, vol.~88, no.~2, pp. 303--338, 2010.

\bibitem{Cityscapes}
M.~Cordts, M.~Omran, S.~Ramos, T.~Rehfeld, M.~Enzweiler, R.~Benenson,
  U.~Franke, S.~Roth, and B.~Schiele, ``The {C}ityscapes dataset for semantic
  urban scene understanding,'' in \emph{The IEEE Conference on Computer Vision
  and Pattern Recognition (CVPR)}, 2016.

\bibitem{Mapillary}
G.~Neuhold, T.~Ollmann, S.~Rota~Bulò, and P.~Kontschieder, ``The {M}apillary
  {V}istas dataset for semantic understanding of street scenes,'' in \emph{The
  IEEE International Conference on Computer Vision (ICCV)}, 2017.

\bibitem{cv:hazop}
O.~Zendel, M.~Murschitz, M.~Humenberger, and W.~Herzner, ``How good is my test
  data? {I}ntroducing safety analysis for computer vision,''
  \emph{International Journal of Computer Vision}, vol. 125, no.~1, pp.
  95--109, 2017.

\bibitem{SynRealDataFogECCV18}
C.~Sakaridis, D.~Dai, S.~Hecker, and L.~Van~Gool, ``Model adaptation with
  synthetic and real data for semantic dense foggy scene understanding,'' in
  \emph{The European Conference on Computer Vision (ECCV)}, 2018.

\bibitem{daytime:2:nighttime}
D.~Dai and L.~Van~Gool, ``Dark model adaptation: Semantic image segmentation
  from daytime to nighttime,'' in \emph{IEEE International Conference on
  Intelligent Transportation Systems}, 2018.

\bibitem{wilddash}
O.~Zendel, K.~Honauer, M.~Murschitz, D.~Steininger, and G.~Fernandez~Dominguez,
  ``{WildDash} - creating hazard-aware benchmarks,'' in \emph{The European
  Conference on Computer Vision (ECCV)}, 2018.

\bibitem{GCMA_UIoU:v1}
C.~Sakaridis, D.~Dai, and L.~Van~Gool, ``Guided curriculum model adaptation and
  uncertainty-aware evaluation for semantic nighttime image segmentation,'' in
  \emph{The IEEE International Conference on Computer Vision (ICCV)}, 2019.

\bibitem{night:vision:pedestrian:05}
F.~Xu, X.~Liu, and K.~Fujimura, ``Pedestrian detection and tracking with night
  vision,'' \emph{IEEE Transactions on Intelligent Transportation Systems},
  vol.~6, no.~1, pp. 63--71, 2005.

\bibitem{pedestrian:detection:tracking:night:09}
J.~Ge, Y.~Luo, and G.~Tei, ``Real-time pedestrian detection and tracking at
  nighttime for driver-assistance systems,'' \emph{IEEE Transactions on
  Intelligent Transportation Systems}, vol.~10, no.~2, pp. 283--298, 2009.

\bibitem{cnn:human:detection:nighttime:17}
J.~H. Kim, H.~G. Hong, and K.~R. Park, ``Convolutional neural network-based
  human detection in nighttime images using visible light camera sensors,''
  \emph{Sensors}, vol.~17, no.~5, 2017.

\bibitem{nighttime:pedestrian:detection:08}
Y.~Chen and C.~Han, ``Night-time pedestrian detection by visual-infrared video
  fusion,'' in \emph{World Congress on Intelligent Control and Automation},
  2008.

\bibitem{kaist:day:night:data:18}
Y.~{Choi}, N.~{Kim}, S.~{Hwang}, K.~{Park}, J.~S. {Yoon}, K.~{An}, and I.~S.
  {Kweon}, ``Kaist multi-spectral day/night data set for autonomous and
  assisted driving,'' \emph{IEEE Transactions on Intelligent Transportation
  Systems}, vol.~19, no.~3, pp. 934--948, 2018.

\bibitem{nighttime:object:proposal:18}
H.~Kuang, K.~Yang, L.~Chen, Y.~Li, L.~L.~H. Chan, and H.~Yan, ``Bayes
  saliency-based object proposal generator for nighttime traffic images,''
  \emph{IEEE Transactions on Intelligent Transportation Systems}, vol.~19,
  no.~3, pp. 814--825, 2018.

\bibitem{night:rear:lights:16}
R.~K. Satzoda and M.~M. Trivedi, ``Looking at vehicles in the night: Detection
  and dynamics of rear lights,'' \emph{IEEE Transactions on Intelligent
  Transportation Systems}, 2016.

\bibitem{road:detection:illumination:invariant}
J.~M.~A. Alvarez and A.~M. Lopez, ``Road detection based on illuminant
  invariance,'' \emph{IEEE Transactions on Intelligent Transportation Systems},
  vol.~12, no.~1, pp. 184--193, 2011.

\bibitem{outdoor:transformation:labeling:iv15}
G.~Ros and J.~M. Alvarez, ``Unsupervised image transformation for outdoor
  semantic labelling,'' in \emph{IEEE Intelligent Vehicles Symposium (IV)},
  2015.

\bibitem{AdapNet:adverse:17}
A.~{Valada}, J.~{Vertens}, A.~{Dhall}, and W.~{Burgard}, ``{AdapNet}: Adaptive
  semantic segmentation in adverse environmental conditions,'' in \emph{IEEE
  International Conference on Robotics and Automation (ICRA)}, 2017.

\bibitem{Oxford}
W.~Maddern, G.~Pascoe, C.~Linegar, and P.~Newman, ``1 year, 1000 km: The
  {Oxford} {RobotCar} dataset,'' \emph{The International Journal of Robotics
  Research}, vol.~36, no.~1, pp. 3--15, 2017.

\bibitem{localization:benchmarking:adverse}
T.~Sattler, W.~Maddern, C.~Toft, A.~Torii, L.~Hammarstrand, E.~Stenborg,
  D.~Safari, M.~Okutomi, M.~Pollefeys, J.~Sivic, F.~Kahl, and T.~Pajdla,
  ``Benchmarking {6DOF} outdoor visual localization in changing conditions,''
  in \emph{The IEEE Conference on Computer Vision and Pattern Recognition
  (CVPR)}, 2018.

\bibitem{astar:3d:dataset:adverse:icra20}
Q.-H. Pham, P.~Sevestre, R.~S. Pahwa, H.~Zhan, C.~H. Pang, Y.~Chen, A.~Mustafa,
  V.~Chandrasekhar, and J.~Lin, ``{A*3D} dataset: Towards autonomous driving in
  challenging environments,'' in \emph{The International Conference in Robotics
  and Automation (ICRA)}, 2020.

\bibitem{MBLLEN}
F.~Lv, F.~Lu, J.~Wu, and C.~Lim, ``{MBLLEN}: Low-light image/video enhancement
  using {CNNs},'' in \emph{Proceedings of the British Machine Vision Conference
  (BMVC)}, 2018.

\bibitem{learning:see:dark:cvpr18}
C.~Chen, Q.~Chen, J.~Xu, and V.~Koltun, ``Learning to see in the dark,'' in
  \emph{The {IEEE} Conference on Computer Vision and Pattern Recognition
  (CVPR)}, 2018.

\bibitem{Wang_2019_CVPR}
R.~Wang, Q.~Zhang, C.-W. Fu, X.~Shen, W.-S. Zheng, and J.~Jia, ``Underexposed
  photo enhancement using deep illumination estimation,'' in \emph{The IEEE
  Conference on Computer Vision and Pattern Recognition (CVPR)}, 2019.

\bibitem{ZeroDCE}
C.~Guo, C.~Li, J.~Guo, C.~C. Loy, J.~Hou, S.~Kwong, and R.~Cong,
  ``Zero-reference deep curve estimation for low-light image enhancement,'' in
  \emph{The IEEE/CVF Conference on Computer Vision and Pattern Recognition
  (CVPR)}, 2020.

\bibitem{benchmark:sensor:adverse:weather:18}
M.~{Bijelic}, T.~{Gruber}, and W.~{Ritter}, ``Benchmarking image sensors under
  adverse weather conditions for autonomous driving,'' in \emph{IEEE
  Intelligent Vehicles Symposium (IV)}, 2018.

\bibitem{wulfmeier2017addressing}
M.~Wulfmeier, A.~Bewley, and I.~Posner, ``Addressing appearance change in
  outdoor robotics with adversarial domain adaptation,'' in \emph{IEEE/RSJ
  International Conference on Intelligent Robots and Systems}, 2017.

\bibitem{continuous:manifold:adaptation}
J.~Hoffman, T.~Darrell, and K.~Saenko, ``Continuous manifold based adaptation
  for evolving visual domains,'' in \emph{The IEEE Conference on Computer
  Vision and Pattern Recognition (CVPR)}, 2014.

\bibitem{SFSU_synthetic}
C.~Sakaridis, D.~Dai, and L.~Van~Gool, ``Semantic foggy scene understanding
  with synthetic data,'' \emph{International Journal of Computer Vision}, vol.
  126, no.~9, pp. 973--992, 2018.

\bibitem{CMAda:IJCV2020}
D.~Dai, C.~Sakaridis, S.~Hecker, and L.~Van~Gool, ``Curriculum model adaptation
  with synthetic and real data for semantic foggy scene understanding,''
  \emph{International Journal of Computer Vision}, vol. 128, no.~5, pp.
  1182--1204, 2020.

\bibitem{FoggySynscapes}
M.~Hahner, D.~Dai, C.~Sakaridis, J.-N. Zaech, and L.~Van~Gool, ``Semantic
  understanding of foggy scenes with purely synthetic data,'' in
  \emph{Proceedings of the 22nd IEEE International Conference on Intelligent
  Transportation Systems}, 2019.

\bibitem{cyCADA}
J.~Hoffman, E.~Tzeng, T.~Park, J.-Y. Zhu, P.~Isola, K.~Saenko, A.~Efros, and
  T.~Darrell, ``{CyCADA}: Cycle-consistent adversarial domain adaptation,'' in
  \emph{International Conference on Machine Learning}, 2018.

\bibitem{learning:synthetic:data:cvpr18}
S.~{Sankaranarayanan}, Y.~{Balaji}, A.~{Jain}, S.~N. {Lim}, and R.~{Chellappa},
  ``Learning from synthetic data: Addressing domain shift for semantic
  segmentation,'' in \emph{The IEEE Conference on Computer Vision and Pattern
  Recognition (CVPR)}, 2018.

\bibitem{chen2018road}
Y.~Chen, W.~Li, and L.~Van~Gool, ``{ROAD}: Reality oriented adaptation for
  semantic segmentation of urban scenes,'' in \emph{The IEEE Conference on
  Computer Vision and Pattern Recognition (CVPR)}, 2018.

\bibitem{adapt:structured:output:cvpr18}
Y.-H. Tsai, W.-C. Hung, S.~Schulter, K.~Sohn, M.-H. Yang, and M.~Chandraker,
  ``Learning to adapt structured output space for semantic segmentation,'' in
  \emph{The IEEE Conference on Computer Vision and Pattern Recognition (CVPR)},
  2018.

\bibitem{incremental:adversarial:DA:18}
M.~Wulfmeier, A.~Bewley, and I.~Posner, ``Incremental adversarial domain
  adaptation for continually changing environments,'' in \emph{IEEE
  International Conference on Robotics and Automation (ICRA)}, 2018.

\bibitem{conditional:GAN:adaptation}
W.~Hong, Z.~Wang, M.~Yang, and J.~Yuan, ``Conditional generative adversarial
  network for structured domain adaptation,'' in \emph{The IEEE Conference on
  Computer Vision and Pattern Recognition (CVPR)}, 2018.

\bibitem{FCNs:adaptation}
Y.~Zhang, Z.~Qiu, T.~Yao, D.~Liu, and T.~Mei, ``Fully convolutional adaptation
  networks for semantic segmentation,'' in \emph{The IEEE Conference on
  Computer Vision and Pattern Recognition (CVPR)}, 2018.

\bibitem{conservative:loss:adaptation}
X.~Zhu, H.~Zhou, C.~Yang, J.~Shi, and D.~Lin, ``Penalizing top performers:
  Conservative loss for semantic segmentation adaptation,'' in \emph{The
  European Conference on Computer Vision (ECCV)}, 2018.

\bibitem{DCAN:adaptation}
Z.~Wu, X.~Han, Y.-L. Lin, M.~G. Uzunbas, T.~Goldstein, S.~Nam~Lim, and L.~S.
  Davis, ``{DCAN}: Dual channel-wise alignment networks for unsupervised scene
  adaptation,'' in \emph{The European Conference on Computer Vision (ECCV)},
  2018.

\bibitem{bidirectional:learning:adaptation}
Y.~Li, L.~Yuan, and N.~Vasconcelos, ``Bidirectional learning for domain
  adaptation of semantic segmentation,'' in \emph{The IEEE Conference on
  Computer Vision and Pattern Recognition (CVPR)}, 2019.

\bibitem{advent:adaptation}
T.-H. Vu, H.~Jain, M.~Bucher, M.~Cord, and P.~Perez, ``{ADVENT}: Adversarial
  entropy minimization for domain adaptation in semantic segmentation,'' in
  \emph{The IEEE Conference on Computer Vision and Pattern Recognition (CVPR)},
  2019.

\bibitem{madan:adaptation}
S.~Zhao, B.~Li, X.~Yue, Y.~Gu, P.~Xu, R.~Hu, H.~Chai, and K.~Keutzer,
  ``Multi-source domain adaptation for semantic segmentation,'' in
  \emph{Advances in Neural Information Processing Systems}, 2019.

\bibitem{ccm:adaptation}
G.~Li, G.~Kang, W.~Liu, Y.~Wei, and Y.~Yang, ``Content-consistent matching for
  domain adaptive semantic segmentation,'' in \emph{The European Conference on
  Computer Vision (ECCV)}, 2020.

\bibitem{plca:adaptation}
G.~Kang, Y.~Wei, Y.~Yang, Y.~Zhuang, and A.~G. Hauptmann, ``Pixel-level cycle
  association: A new perspective for domain adaptive semantic segmentation,''
  \emph{CoRR}, vol. abs/2011.00147, 2020.

\bibitem{curriculum:domain:adaptation:17}
Y.~{Zhang}, P.~{David}, and B.~{Gong}, ``Curriculum domain adaptation for
  semantic segmentation of urban scenes,'' in \emph{The IEEE International
  Conference on Computer Vision (ICCV)}, 2017.

\bibitem{self:training:adaptation}
Y.~Zou, Z.~Yu, B.~Vijaya~Kumar, and J.~Wang, ``Unsupervised domain adaptation
  for semantic segmentation via class-balanced self-training,'' in \emph{The
  European Conference on Computer Vision (ECCV)}, 2018.

\bibitem{compound:domain:adaptation}
Z.~Liu, Z.~Miao, X.~Pan, X.~Zhan, D.~Lin, S.~X. Yu, and B.~Gong, ``Open
  compound domain adaptation,'' in \emph{The IEEE Conference on Computer Vision
  and Pattern Recognition (CVPR)}, 2020.

\bibitem{cross:season:correspondence}
M.~Larsson, E.~Stenborg, L.~Hammarstrand, M.~Pollefeys, T.~Sattler, and
  F.~Kahl, ``A cross-season correspondence dataset for robust semantic
  segmentation,'' in \emph{The IEEE Conference on Computer Vision and Pattern
  Recognition (CVPR)}, 2019.

\bibitem{panoptic:segmentation}
A.~Kirillov, K.~He, R.~Girshick, C.~Rother, and P.~Doll{\'{a}}r, ``Panoptic
  segmentation,'' in \emph{The IEEE Conference on Computer Vision and Pattern
  Recognition (CVPR)}, 2019.

\bibitem{uncertainty:bayesian}
A.~Kendall and Y.~Gal, ``What uncertainties do we need in {B}ayesian deep
  learning for computer vision?'' in \emph{Advances in Neural Information
  Processing Systems}, 2017.

\bibitem{simultaneous:segmentation:outliers}
P.~Bevandi{\'{c}}, I.~Kre{\v{s}}o, M.~Or{\v{s}}i{\'{c}}, and
  S.~{\v{S}}egvi{\'{c}}, ``Simultaneous semantic segmentation and outlier
  detection in presence of domain shift,'' in \emph{German Conference on
  Pattern Recognition}, 2019.

\bibitem{localization:probabilistic:maps}
J.~{Levinson} and S.~{Thrun}, ``Robust vehicle localization in urban
  environments using probabilistic maps,'' in \emph{IEEE International
  Conference on Robotics and Automation (ICRA)}, 2010.

\bibitem{video:localization:iv14}
J.~{Ziegler}, H.~{Lategahn}, M.~{Schreiber}, C.~G. {Keller}, C.~{Knöppel},
  J.~{Hipp}, M.~{Haueis}, and C.~{Stiller}, ``Video based localization for
  bertha,'' in \emph{IEEE Intelligent Vehicles Symposium}, 2014.

\bibitem{map:drivable:space:12}
J.~{Moras}, F.~S.~A. {Rodríguez}, V.~{Drevelle}, G.~{Dherbomez},
  V.~{Cherfaoui}, and P.~{Bonnifait}, ``Drivable space characterization using
  automotive lidar and georeferenced map information,'' in \emph{Intelligent
  Vehicles Symposium (IV)}, 2012.

\bibitem{map:road:recognition:13}
K.~{Irie} and M.~{Tomono}, ``Road recognition from a single image using prior
  information,'' in \emph{International Conference on Intelligent Robots and
  Systems (IROS)}, 2013.

\bibitem{navigation:openstreetmap:10}
M.~Hentschel and B.~Wagner, ``Autonomous robot navigation based on
  openstreetmap geodata,'' in \emph{IEEE Conference on Intelligent
  Transportation Systems}, 2010.

\bibitem{drive:surroundview:route:planner}
S.~Hecker, D.~Dai, and L.~Van~Gool, ``End-to-end learning of driving models
  with surround-view cameras and route planners,'' in \emph{European Conference
  on Computer Vision (ECCV)}, 2018.

\bibitem{map:aided:driving:scene:understanding:15}
M.~{Kurdej}, J.~{Moras}, V.~{Cherfaoui}, and P.~{Bonnifait}, ``Map-aided
  evidential grids for driving scene understanding,'' \emph{IEEE Intelligent
  Transportation Systems Magazine}, vol.~7, no.~1, pp. 30--41, 2015.

\bibitem{HDNET:2018}
B.~Yang, M.~Liang, and R.~Urtasun, ``Hdnet: Exploiting hd maps for 3d object
  detection,'' in \emph{Conference on Robot Learning}, vol.~87, 2018, pp.
  146--155.

\bibitem{argoverse:cvpr19}
M.-F. Chang, J.~Lambert, P.~Sangkloy, J.~Singh, S.~Bak, A.~Hartnett, D.~Wang,
  P.~Carr, S.~Lucey, D.~Ramanan, and J.~Hays, ``Argoverse: 3d tracking and
  forecasting with rich maps,'' in \emph{The IEEE Conference on Computer Vision
  and Pattern Recognition (CVPR)}, June 2019.

\bibitem{action:prediction:map:20}
J.-N. Zaech, D.~Dai, A.~Liniger, and L.~{Van Gool}, ``Action sequence
  predictions of vehicles in urban environments using map and social context,''
  \emph{arXiv:2004.14251}, 2020.

\bibitem{cycleGAN}
J.-Y. Zhu, T.~Park, P.~Isola, and A.~A. Efros, ``Unpaired image-to-image
  translation using cycle-consistent adversarial networks,'' in \emph{The IEEE
  International Conference on Computer Vision (ICCV)}, 2017.

\bibitem{curriculum:learning}
Y.~Bengio, J.~Louradour, R.~Collobert, and J.~Weston, ``Curriculum learning,''
  in \emph{International Conference on Machine Learning}, 2009.

\bibitem{bilateral:grid}
S.~Paris and F.~Durand, ``A fast approximation of the bilateral filter using a
  signal processing approach,'' \emph{International Journal of Computer
  Vision}, vol.~81, no.~1, pp. 24--52, 2009.

\bibitem{monocular:neural:rendering}
X.~Chen, J.~Song, and O.~Hilliges, ``Monocular neural image based rendering
  with continuous view control,'' in \emph{The IEEE International Conference on
  Computer Vision (ICCV)}, 2019.

\bibitem{monodepth2}
C.~Godard, O.~Mac~Aodha, M.~Firman, and G.~J. Brostow, ``Digging into
  self-supervised monocular depth estimation,'' in \emph{The IEEE International
  Conference on Computer Vision (ICCV)}, 2019.

\bibitem{SURF}
H.~Bay, T.~Tuytelaars, and L.~Van~Gool, ``{SURF}: Speeded up robust features,''
  in \emph{The European Conference on Computer Vision (ECCV)}, 2006.

\bibitem{raincouver}
F.~{Tung}, J.~{Chen}, L.~{Meng}, and J.~J. {Little}, ``The {Raincouver} scene
  parsing benchmark for self-driving in adverse weather and at night,''
  \emph{IEEE Robotics and Automation Letters}, vol.~2, no.~4, pp. 2188--2193,
  2017.

\bibitem{BDD100K}
F.~Yu, W.~Xian, Y.~Chen, F.~Liu, M.~Liao, V.~Madhavan, and T.~Darrell,
  ``{BDD100K:} a diverse driving video database with scalable annotation
  tooling,'' \emph{CoRR}, vol. abs/1805.04687, 2018.

\bibitem{DeepLab:v2}
L.~C. Chen, G.~Papandreou, I.~Kokkinos, K.~Murphy, and A.~L. Yuille,
  ``{DeepLab}: Semantic image segmentation with deep convolutional nets, atrous
  convolution, and fully connected {CRFs},'' \emph{IEEE Transactions on Pattern
  Analysis and Machine Intelligence}, vol.~40, no.~4, pp. 834--848, 2018.

\bibitem{resnet}
K.~He, X.~Zhang, S.~Ren, and J.~Sun, ``Deep residual learning for image
  recognition,'' in \emph{The IEEE Conference on Computer Vision and Pattern
  Recognition (CVPR)}, June 2016.

\bibitem{joint:bilateral:upsampling}
J.~Kopf, M.~F. Cohen, D.~Lischinski, and M.~Uyttendaele, ``Joint bilateral
  upsampling,'' in \emph{ACM SIGGRAPH 2007 Papers}, 2007.

\bibitem{adaptive:histogram:equalization}
S.~M. Pizer, E.~P. Amburn, J.~D. Austin, R.~Cromartie, A.~Geselowitz, T.~Greer,
  B.~ter Haar~Romeny, J.~B. Zimmerman, and K.~Zuiderveld, ``Adaptive histogram
  equalization and its variations,'' \emph{Computer Vision, Graphics, and Image
  Processing}, vol.~39, no.~3, pp. 355--368, 1987.

\end{thebibliography}

%

\begin{IEEEbiography}[{\includegraphics[width=1in,clip,keepaspectratio]{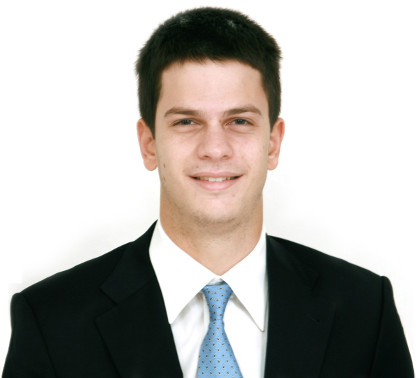}}]{Christos Sakaridis}
is a PhD candidate at Computer Vision Lab, ETH Zurich, since 2016. His broad research fields are Computer Vision and Machine Learning, while his current focus is on visual recognition in adverse conditions, domain adaptation, and segmentation. Prior to joining Computer Vision Lab, he received the MSc degree in Computer Science from ETH Zurich in 2016 and the Diploma degree in Electrical and Computer Engineering from National Technical University of Athens in 2014. 
\end{IEEEbiography}

\begin{IEEEbiography}[{\includegraphics[width=1in,clip,keepaspectratio]{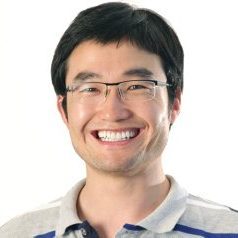}}]{Dengxin Dai}
is a senior scientist working with the Computer Vision Lab at ETH Zurich. In 2016, he obtained his PhD in Computer Vision at ETH Zurich. Since then he is the Team Leader of the TRACE-Zurich team, working on Autonomous Driving in collaboration with Toyota. His research interests lie in autonomous driving, robust perception in adverse weather and lighting conditions, automotive sensors, and learning under limited supervision. He has organized a CVPR Workshop series ('19, '20) on Vision for All Seasons: Bad Weather and Nighttime, and has organized an ICCV'19 workshop on Autonomous Driving. He is the lead guest editor for the IJCV special issue Vision for All Seasons, and an area chair for WACV'20 and CVPR'21.
\end{IEEEbiography}

\vfill

\newpage

\begin{IEEEbiography}[{\includegraphics[width=1in,clip,keepaspectratio]{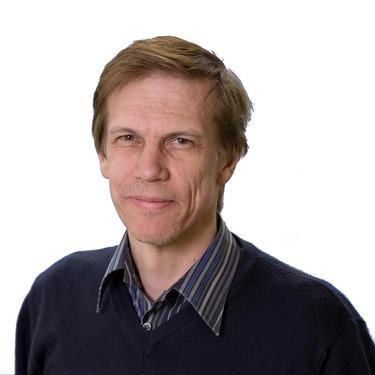}}]{Luc Van Gool}
is a full professor for Computer Vision at ETH Zurich and the KU Leuven. He leads research and teaches at both places. He has authored over 300 papers. He has been a program committee member of several, major computer vision conferences (e.g.\ Program Chair ICCV’05, Beijing, General Chair of ICCV’11, Barcelona, and of ECCV’14, Zurich). His main interests include 3D reconstruction and modeling, object recognition, and autonomous driving. He received several Best Paper awards (e.g.\ David Marr Prize ’98, Best Paper CVPR’07). He received the Koenderink Award in 2016 and the ‘Distinguished Researcher’ nomination by the IEEE Computer Society in 2017. In 2015 he also received the 5-yearly Excellence Prize by the Flemish Fund for Scientific Research. He was the holder of an ERC Advanced Grant (VarCity). Currently, he leads computer vision research for autonomous driving in the context of the Toyota TRACE labs in Leuven and at ETH, and has an extensive collaboration with Huawei on the topic of image and video enhancement.
\end{IEEEbiography}

\vfill


\end{document}